\documentclass{article}

\PassOptionsToPackage{numbers, compress}{natbib}
\usepackage{colm}

\usepackage[utf8]{inputenc} 
\usepackage[T1]{fontenc}    
\usepackage{hyperref}       
\usepackage{url}            
\usepackage{booktabs}       
\usepackage{amsfonts}       
\usepackage{nicefrac}       
\usepackage{microtype}      
\usepackage{graphicx}
\usepackage{listings}
\usepackage{algorithm}
\usepackage{amsfonts} 
\usepackage{amsmath}
\usepackage{algpseudocode}
\usepackage[nameinlink,capitalise]{cleveref}      
\usepackage{enumitem,amssymb}
\usepackage{placeins}
\usepackage{wrapfig}
\usepackage{booktabs}
\usepackage{adjustbox}
\usepackage[normalem]{ulem}
\usepackage[font=small,labelfont=bf]{caption}
\usepackage[nounderscore]{syntax}
\usepackage{xargs}
\usepackage[at]{easylist}
\usepackage{bold-extra}
\usepackage{siunitx}
\usepackage{multirow}
\usepackage{threeparttable}
\usepackage{etoolbox}
\usepackage{bbm}
\usepackage{subcaption}
\usepackage{rotating}
\usepackage{pdflscape}
\usepackage{tabularx}
\usepackage{colortbl}

\AtBeginEnvironment{easylist}{\ListProperties(Start1=1)}
\Crefformat{appendix}{#2Supplementary~information~#1#3}

\usepackage[colorinlistoftodos,prependcaption,textsize=scriptsize]{todonotes}
\newcommandx{\fixme}[2][1=]{\todo[linecolor=red,backgroundcolor=red!25,bordercolor=red,#1]{#2}}
\newcommandx{\changeme}[2][1=]{\todo[linecolor=blue,backgroundcolor=blue!25,bordercolor=blue,#1]{#2}}
\newcommandx{\info}[2][1=]{\todo[linecolor=OliveGreen,backgroundcolor=OliveGreen!25,bordercolor=OliveGreen,#1]{#2}}
\newcommandx{\ablation}[2][1=]{\todo[linecolor=BurntOrange,backgroundcolor=BurntOrange!25,bordercolor=BurntOrange,#1]{#2}}
\newcommandx{\todone}[2][1=]{\todo[linecolor=Gray,backgroundcolor=Gray!25,bordercolor=Gray,#1]{\sout{#2}}}

\newcommand{\dsl}[1]{{\it $\langle$#1$\rangle$}}

\newlist{todolist}{itemize}{2}
\setlist[todolist]{label=$\square$}
\usepackage{pifont}

\newlength{\gamecolumnwidth}
\newlength{\textcolumnwidth}
\lstdefinelanguage{PDDL}
{
  sensitive=false,    
  morecomment=[l]{;}, 
  alsoletter={:,-},   
  morekeywords={
    define,domain,problem,not,and,or,when,forall,exists,either,
    :domain,:requirements,:types,:objects,:constants,
    :predicates,:action,:parameters,:precondition,:effect,
    :fluents,:primary-effect,:side-effect,:init,:goal,
    :strips,:adl,:equality,:typing,:conditional-effects,
    :negative-preconditions,:disjunctive-preconditions,
    :existential-preconditions,:universal-preconditions,:quantified-preconditions,
    :functions,assign,increase,decrease,scale-up,scale-down,
    :metric,minimize,maximize,
    :durative-actions,:duration-inequalities,:continuous-effects,
    :durative-action,:duration,:condition,
    preference,then,once,hold,hold-while,at-end,
    game-optional,game-conserved
  },
}
\definecolor{SyntaxBlue}{HTML}{569CD6}
\definecolor{VariableBlue}{HTML}{9CDCFE}
\definecolor{CommentGreen}{HTML}{008000}
\definecolor{CommentGray}{HTML}{969696}
\definecolor{VariableGreen}{HTML}{009900}
\usepackage{etoolbox}
\makeatletter
\patchcmd{\hyper@makecurrent}{%
    \ifx\Hy@param\Hy@chapterstring
        \let\Hy@param\Hy@chapapp
    \fi
}{%
    \iftoggle{inappendix}{
        \@checkappendixparam{chapter}%
        \@checkappendixparam{section}%
        \@checkappendixparam{subsection}%
        \@checkappendixparam{subsubsection}%
        \@checkappendixparam{paragraph}%
        \@checkappendixparam{subparagraph}%
    }{}%
}{}{\errmessage{failed to patch}}

\newcommand*{\@checkappendixparam}[1]{%
    \def\@checkappendixparamtmp{#1}%
    \ifx\Hy@param\@checkappendixparamtmp
        \let\Hy@param\Hy@appendixstring
    \fi
}
\makeatletter

\newtoggle{inappendix}
\togglefalse{inappendix}

\apptocmd{\appendix}{\toggletrue{inappendix}}{}{\errmessage{failed to patch}}

\algnewcommand{\LineComment}[1]{\State \(\triangleright\) #1}

\title{Goals as Reward-Producing Programs}

\colmfinalcopy

\author{%
    Guy Davidson\textsuperscript{$\alpha$*\textdagger}, Graham Todd\textsuperscript{$\beta$*}, Julian Togelius\textsuperscript{$\beta$}, Todd M. Gureckis\textsuperscript{$\gamma$},  Brenden M. Lake\textsuperscript{$\alpha,\gamma$} \\
    \textsuperscript{$\alpha$}Center for Data Science, \textsuperscript{$\beta$}Game Innovation Lab, \textsuperscript{$\gamma$}Department of Psychology \\
    New York University \\
    *: Co-first authorship \\
    \textdagger: Corresponding author: \texttt{guy.davidson@nyu.edu}  \\
}

\begin{document}

\maketitle

\begin{abstract}
People are remarkably capable of generating their own goals, beginning with child's play and continuing into adulthood.
Despite considerable empirical and computational work on goals and goal-oriented behavior, models are still far from capturing the richness of everyday human goals.
Here, we bridge this gap by collecting a dataset of human-generated playful goals (in the form of scorable, single-player games), modeling them as reward-producing programs, and generating novel human-like goals through program synthesis. 
Reward-producing programs capture the rich semantics of goals through symbolic operations that compose, add temporal constraints, and allow for program execution on behavioral traces to evaluate progress. 
To build a generative model of goals, we learn a fitness function over the infinite set of possible goal programs and sample novel goals with a quality-diversity algorithm.
Human evaluators found that model-generated goals, when sampled from partitions of program space occupied by human examples, were indistinguishable from human-created games. 
We also discovered that our model's internal fitness scores predict games that are evaluated as more fun to play and more human-like.
\end{abstract}



\begin{figure}[!bthp]
    \centering
    \includegraphics[width=\columnwidth]{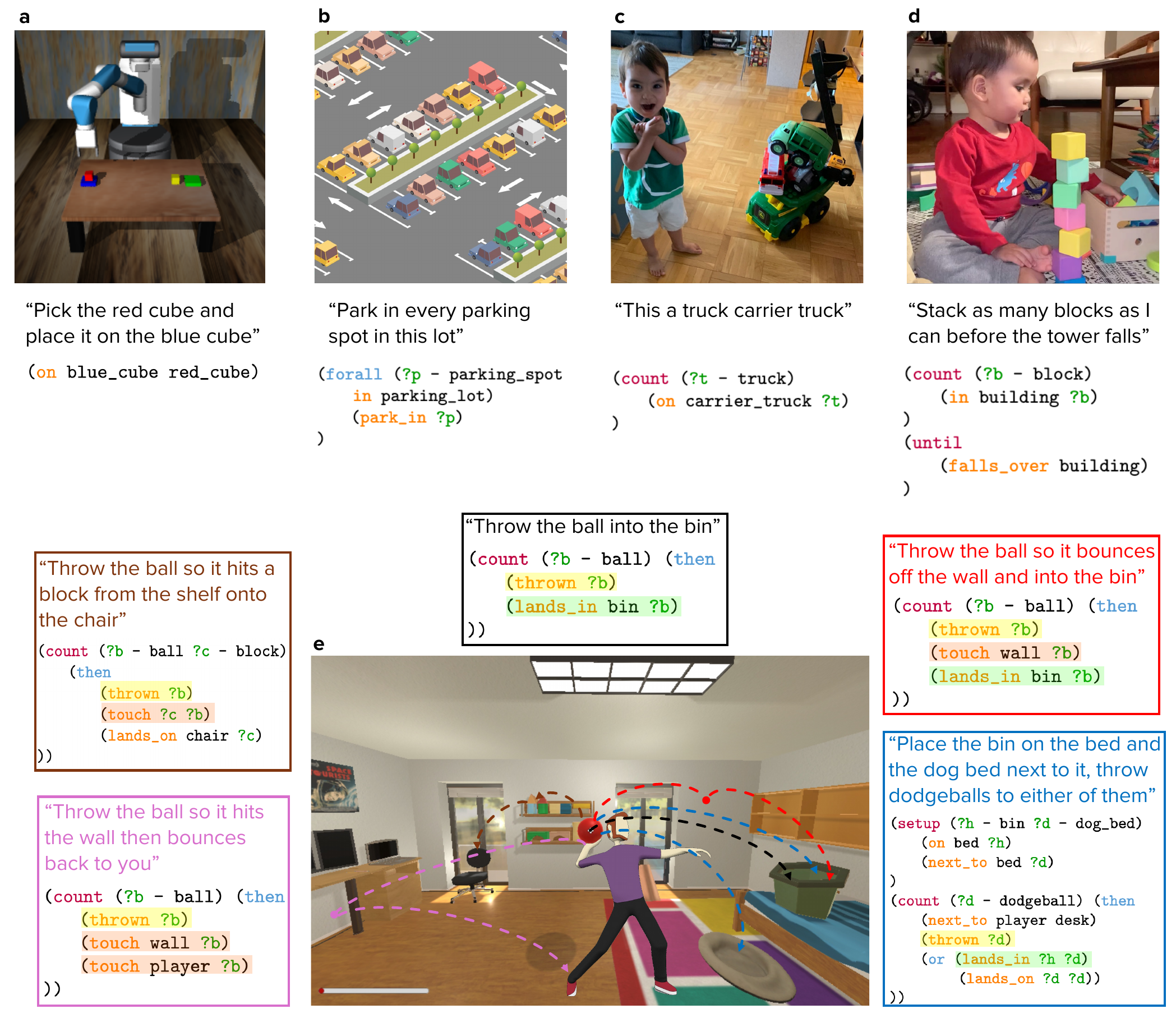}
    \caption{\textbf{Goals as Reward-Producing Programs.}
    Panels \textbf{a-d} show different goals, presented in natural language and mapped to pseudo-code in a program-like representation. 
    Panel \textbf{e} shows a set of varied yet related goals in our experiment environment, of which the blue and pink were created by participants in our experiment. 
    Each goal is represented by a throw trajectory (dashed line in the illustration) matching a description of the goal (whose text is the same color as the line). 
    We highlight shared compositional components between programs in yellow, orange, and green. 
    Our program representations are reward-producing, that is, run on sequences of agent interactions with an environment (state-action pairs) and emit a score with respect to the specified goal. 
    Our pseudo-code and domain-specific language both use a LISP-like syntax, where function calls have the function name as the first token inside the parentheses. 
    Participants in our experiment created some of these goals; see \autoref{fig:appendix-pseudocode-translation} for representations of the blue and pink programs in our domain-specific language.} 
    \label{fig:1}
\end{figure}

Understanding how humans create, represent, and reason about goals is crucial to understanding human behavior.
Goals are pervasive throughout psychology \cite{Dweck1992, Austin1996, Elliot2008}, having been studied from perspectives such as motivation \cite{Hyland1988motivational, Eccles2002, Brown2007motivation}, personality and social psychology \cite{Fishbach2007social, Pervin2015personality}, and learning and decision-making \cite{Moskowitz2009psychology, Molinaro2023}.  
But what is a goal? Elliot \& Fryer offer the workable, albeit simplified definition: a representation of a future object to be approached or avoided (see also \cite{Elliot2008, Molinaro2023}). 
Reinforcement learning offers another formulation, operationalizing goals as maximizing cumulative reward over a series of steps ~\cite{Sutton2018reinforcement}. 
Typical goals in reinforcement learning tasks include reaching a target location, winning in a video or board game \cite{mnih2015human}, or placing an object in a specified position (e.g., \autoref{fig:1}a), such that success can be characterized by reaching a target state.

In contrast, people routinely create novel, idiosyncratic goals with richness beyond these common modeling settings.
Chu et al. \cite{Chu2024} report the example of Gareth Wild, who set an unusual goal for himself to park in every spot in a particular grocery store's parking lot (\autoref{fig:1}b).
Children routinely devise fun and compelling goals without external guidance, such as creating a ``truck carrier truck'' (\autoref{fig:1}c) or stacking as many blocks as possible in a single tower (\autoref{fig:1}d). 
Beyond being fun, these playful goals play a crucial role in learning to structure and solve arbitrary problems \cite{Chu2020a, Lillard2015, Andersen2023}. 
Indeed, it has been argued that autonomously setting and achieving goals is a core component of human intelligence \cite{Chu2024, Oudeyer2007}.

We propose a framework for modeling human goal generation as synthesizing reward-producing programs (\autoref{fig:1}, bottom row).
There are several advantages to representing goals as symbolic programs, which map an agent's behavior to a reward score indicating the degree of success.
First, a structured language facilitates the compositional reuse of motifs across disparate goals. 
Such reuse makes capturing the wide range of human creativity in goal creation substantially more tractable:
In \autoref{fig:1}e, we illustrate a simple ball-throwing game (in black) and four distinct variants (in red, blue, pink, and brown) composed in part from shared components: balls being thrown (highlighted in yellow), the thrown ball hitting something (orange), and the thrown ball landing somewhere (green). 
Second, our choice of representation makes goal semantics explicit. 
The particular grammatical elements of our representation each fulfill particular roles, such as \textit{predicates} (i.e., specific and evaluable relations between objects, colored orange in the programs in \autoref{fig:1}) and \textit{temporal modals} (i.e., relationships in time between goal components, such as `until' and `then' in \autoref{fig:1}).
Finally, goals-as-programs are executable; that is, they can be computationally interpreted to detect when a goal is entirely or partially achieved (\autoref{fig:1}e, each program would be interpreted and provide a score only when the matching throw trajectory is completed).

In this article, we demonstrate that programs can capture real human-created goals in a naturalistic domain and build a model capable of generating new programs representing human-like goals.  
We devised a rich experimental environment for goal generation and asked human participants to generate playful goals in the form of single-player scoreable games (and see \cite{Nguyen2020GamesAgency} on the relationship between games and goals). 
We translated these games into programs in a domain-specific language that explicitly models the core semantics of the participants' creations. 
We also developed a Goal Program Generator (GPG) model to generate new goals in this representation, learning a fitness metric over programs to capture human likeness and sampling diverse goal programs to maximize fitness. 
We found that the model succeeds in generating novel games distinct from examples in the training dataset. 
Human raters evaluated several characteristics of model-generated games, including how human-like they were.  
Model games from sections of program space closer to participant-created games were judged indistinguishably from the real games, but model samples further away were not rated as highly on average. 
Analyses revealed that our learned fitness function predicts several human judgment questions, including how human-like games are rated. 
These results demonstrate that our goal representations and model capture important aspects of how people creatively construct new goals, generating plausible, diverse goals and predicting understandability and fun ratings.
We conclude with a discussion of the scope of our representational hypothesis (capturing goals as programs), the relationship to prior work, some limitations of our model, and avenues for future work.

\section*{Behavioral results}

\begin{figure}[!bthp]
    \centering
    \includegraphics[width=\columnwidth]{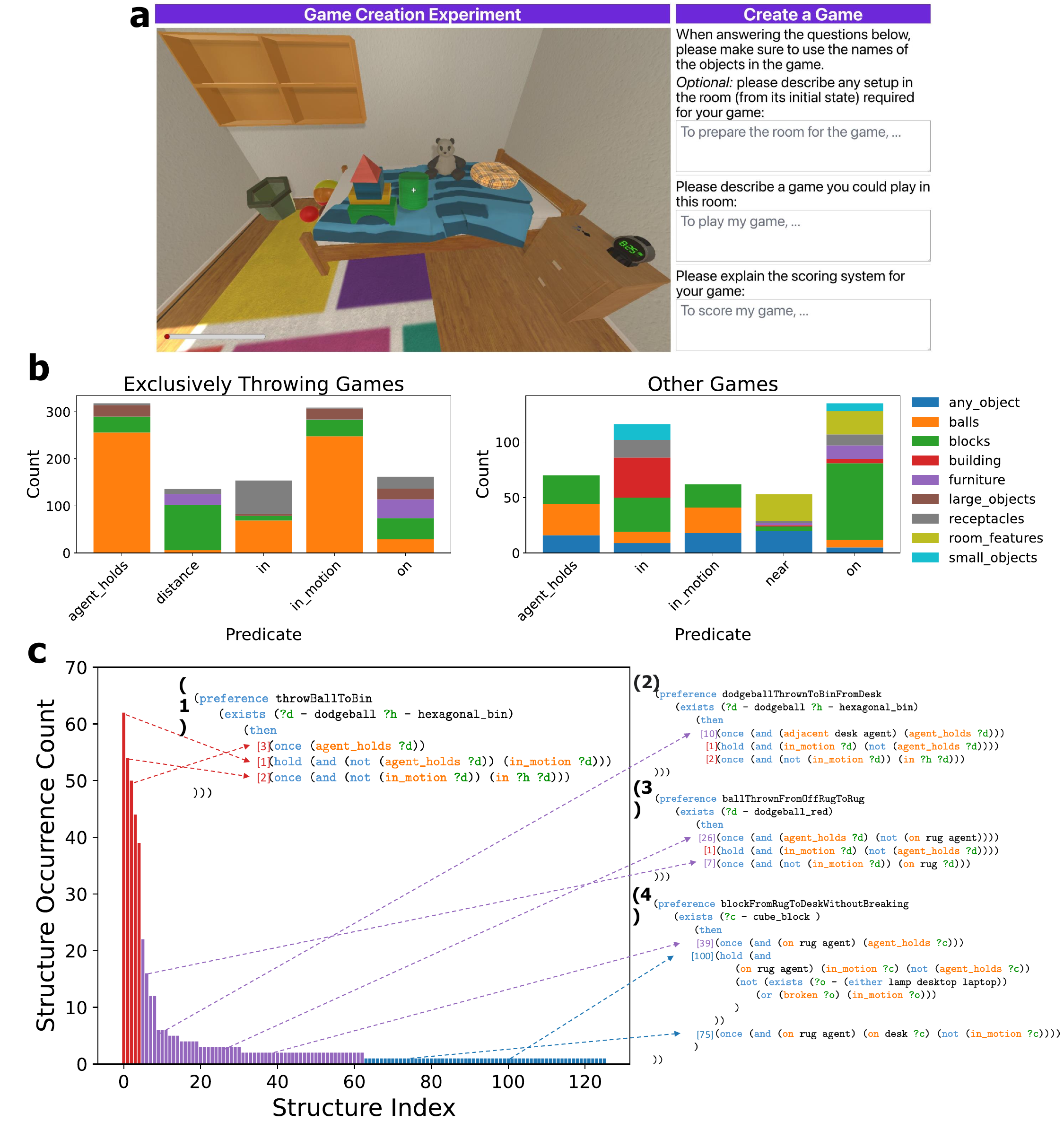}
    \caption{\textbf{Participants in our behavioral experiment create diverse games reflecting common sense and compositionality.}
    \textbf{(a):} Our online game creation experiment (see full interface in \autoref{fig:experiment-interface}.
    \textbf{(b):} Participants showcase intuitive common sense. 
    \textit{Left:} In games involving exclusively throwing, participants use balls (orange) far more often than any other object type. 
    \textit{Right:} In other games, participants refer to blocks or ``any object'' more often, most often checking where objects are placed (using the \lstinline{in} and \lstinline{on} predicates).
    We most often observe balls being thrown and blocks being stacked, and while a few participants specified block-throwing games, no participant created a game involving ball-stacking. 
    Participants also rarely specified throwing large or cumbersome objects (such as the chair or laptop), and only used buildings to specify stacking objectives (as opposed to moving or throwing them).
    See \autoref{fig:behavioral-common-sense} for an extended version of this panel (including additional object categories and predicate).
    \textbf{(c):} We analyze the occurrence of various abstract structures in our programs (see \nameref{methods:dataset-analyses} for details).
    \textbf{\textcolor[RGB]{214,39,40}{Red:}} The five most common structures cover almost half (47.5\%) of total occurrences, showing extensive compositional reuse. 
    The three most common structures combine into simple ball-to-bin throwing preference (\textbf{(1)}, structure indices in square brackets). 
    \textbf{\textcolor[RGB]{148,103,189}{Purple:}} Other structures are reused fewer times, covering most remaining occurrences (another 40.5\%).
    These rarer structures allow for creating more complex throwing elements, constraining where the player throws the ball from \textbf{(2,3)} or to \textbf{(3)}.
    \textbf{\textcolor[RGB]{31,119,180}{Blue:}} Exactly half of the structures (63 / 126) appear only once --- this long tail of expressions offers evidence of creativity.
    The last throwing preference \textbf{(4)}, specifying throwing a block from the rug onto the desk without moving off the rug or breaking any of the objects on the desk, uses two unique structures. 
    } 
    \label{fig:behavioral-composite}
\end{figure}

Although goals play a crucial role in psychological theory, there are few, if any, empirical paradigms for eliciting wide-ranging goals from study participants. 
We created an experimental setting that aims to capture the rich, playful, and creative nature of how children (and adults) create everyday goals. 
We used AI2-THOR \cite{kolve2017ai2} (an embodied, 3D environment simulation) to set up a room resembling a child's bedroom, filled with toys and other common objects (\autoref{fig:behavioral-composite}a, and see \autoref{fig:experiment-interface} for a larger version). 
In our task, we asked participants to propose a single-player game to be played in the room.  
This design allowed participants to imagine and propose a wide range of playful goals, with the aim of game generation helping to make the resulting goals more concrete.
We collected a dataset of 98 games, described by participants in natural language. In addition, we recorded full state-action traces of each participant's interactions with the environment, which we leveraged in later experiments (see \nameref{methods:dataset} for additional details).

We then manually translated each game from natural language to programs in a domain-specific language (DSL), inspired by language of thought models in computational cognitive science \cite{Fodor1979, Goodman2008, Piantadosi2012, Rule2020, WongGrand2023}. The DSL is used to model the semantics of games in our dataset, independent of the exact natural language phrasing.  
Though the translation from natural language to DSL is unlikely to be lossless, we aim to capture the core semantics of the rich and generative structure of human goals with these relatively simple programs.
This DSL was derived from the Planning Domain Definition Language (PDDL \cite{Ghallab1998}), which offers a basic representation for specifying goals (i.e. end states of plans) and preferences (i.e. other costs to optimize while planning).
Each program in the DSL contains two mandatory sections: gameplay \textbf{preferences} describing how a game is played, and \textbf{scoring} rules specifying how to determine a player's score based on the satisfaction of the game's preferences. 
Game programs may also contain optional \textbf{setup} instructions and \textbf{terminal} conditions (see \cref{sec:appendix-dsl-ref} for the full DSL).

Our choice to represent games as programs allows us to quantitatively analyze their structure and fundamental components.
We found that people recruit an intuitive physical common sense when creating games (\autoref{fig:behavioral-composite}b, and see \nameref{methods:dataset-analyses} for details). 
For instance, if an object is thrown, it's likely a ball, and if an object is stacked, it's likely a block --- and while a few participants specified games involving throwing blocks, none attempted to stack balls. 
Similarly, participants did not specify throwing cumbersome objects (such as the laptop or chair), and a participant who specified throwing a large `beach ball' clarified that it should land \textit{on} the bin (as the ball does not fit within the bin). 
We also observed evidence of both compositionality (common structure reuse) and creativity (preponderance of unique structures) across our participants, summarized in \autoref{fig:behavioral-composite}c (see \nameref{methods:dataset-analyses} for details).
Counting occurrences of grammatical structures while abstracting over the identities of individual objects (i.e., treating the modal expressions `the agent holds a block' \lstinline{(once (agent_holds block))} and `the agent holds a ball' \lstinline{(once (agent_holds ball))} the same), we find the five most common structures cover almost half of the total observations, showing how representing goals as programs can reveal shared, compositional substructure. 
At the other end of the distribution, we also observe a long tail emblematic of creativity, as one-half of the unique structures we count appear exactly once.
Despite not being explicitly prompted to generate novel or creative games, many participants proposed entirely unique gameplay ideas, encouraging us that our experimental paradigm elicits rich and creative goal creation.


\section*{Modelling Results}

\begin{figure}[!bthp]
    \vspace{-0.5in}
    \centering
    \includegraphics[width=\columnwidth]{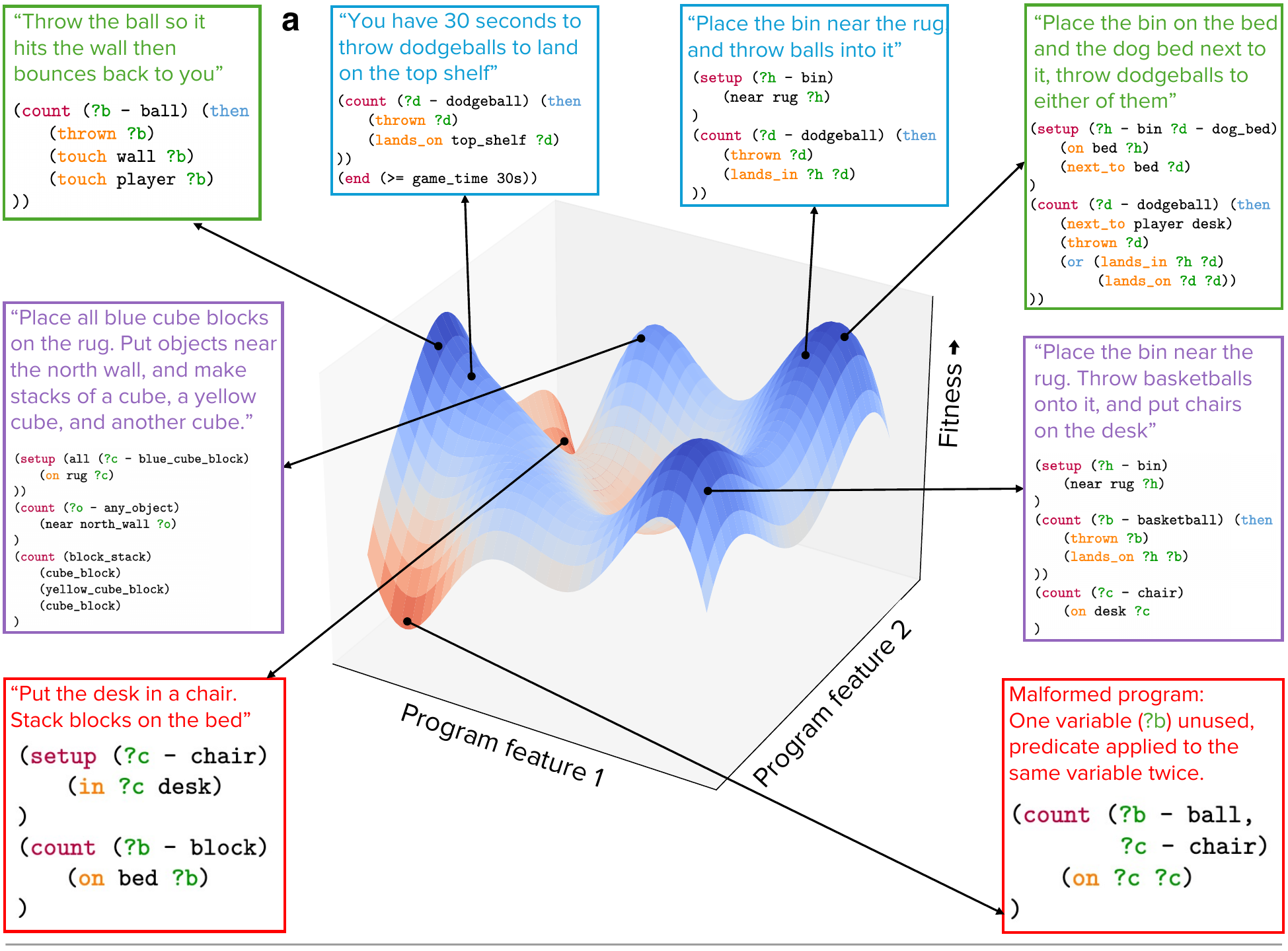}
    \includegraphics[width=\columnwidth]{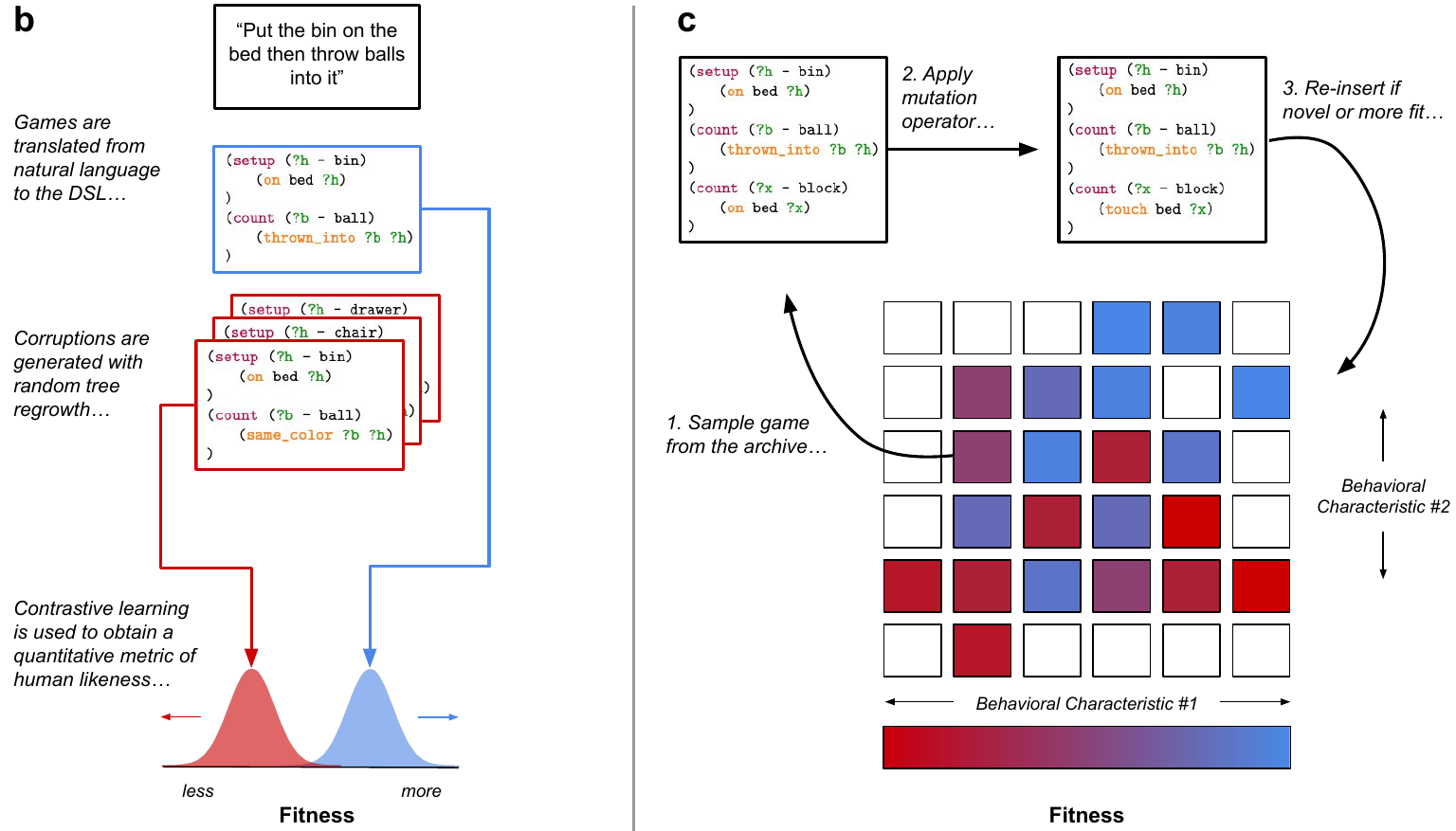}
    \captionsetup{font=small}
    \caption{\textbf{Goal Program Generator model.} 
    \textbf{(a) Overview}: Our model operates on programs in some high-dimensional space (visualized in two dimensions).
    We learn a fitness metric (Z-axis) capturing desirable aspects of programs using a dataset of human-created goals (highlighted in \textcolor[RGB]{78,167,46}{green}).
    Our model then generates diverse new samples maximizing the fitness measure, some ``matched'' to participant-created goal programs on diversity criteria (in \textcolor[RGB]{15,158,213}{blue}) and other ``unmatched'' novel goals (in \textcolor[RGB]{148,103,189}{purple}).
    These programs stand in contrast to potential failure modes, such as generating programs that are malformed or semantically incoherent (in \textcolor{red}{red}).
    All (non-red) goals in this figure were created by participants in our experiment or our model; see \autoref{fig:appendix-pseudocode-translation} for their full representations in our domain-specific language.
    \textbf{(b) Parameter learning:} We contrastively learn a quantitative measure of fitness (the Z axis in \textbf{(a)}) by maximizing the distance between human-generated exemplar games and a set of corruptions obtained through random tree regrowth. 
    \textbf{(c) Search:} This measure is then used as the basis for quality-diversity optimization using MAP-Elites. The algorithm maintains an archive of games that differ across phenotypic ``behavioral characteristics.'' At each step, a game is randomly sampled from the archive (1), randomly mutated (2), and re-evaluated for fitness and its position in the archive. It is added to the archive only if it would occupy a previously empty position or if it is more fit than the current occupant (3).
    }
    \label{fig:model-overview}
\end{figure}

We next develop a computational model to synthesize human-like goals.
Guided by insights from our behavioral analyses, we design our model to explicitly leverage cognitive capacities that people seem to recruit in creating goals.
Our Goal Program Generator model (GPG, illustrated in \autoref{fig:model-overview}) operates over a high-dimensional program space and learns how to generate goals maximizing a fitness measure. 
Upon entering a new environment, people can create goals without extensive data-driven demonstrations; therefore we aim for a model that can similarly generate goals without a large number of examples.
The GPG consists of two main elements: a fitness function and a search procedure. 
The fitness function (learned from data) attempts to quantify human likeness over the space of goal programs (\autoref{fig:model-overview}a), such that a higher score indicates a better generated goal (\autoref{fig:model-overview}b). 
The search procedure generates diverse samples that maximize this fitness function (\autoref{fig:model-overview}c).
As a framework, the GPG model is committed to the idea of evaluating the quality of goals-as-programs with a learned objective function and less so to the specific algorithms used for optimization and search.


The fitness function $f(g) = \theta \cdot \phi(g)$ maps $f: \mathcal{G} \rightarrow \mathbb{R}$ from a game $g \in \mathcal{G}$ to a real-valued score that aims to encode its human-likeness (\autoref{fig:model-overview}b). 
We transform each game into an 89-dimensional vector of features that capture properties relating to structure (e.g., the size and depth of its syntax tree), logic (e.g., whether any expressions are redundant), or goal semantics (e.g., the extent to which different parts of the goal are interrelated). 
We leverage our programmatic representation of goals in order to automate this feature extraction process (see \nameref{methods:fitness-function} for details). 
In this implementation, parameter learning of feature weights $\theta$ proceeds in a \textit{contrastive} fashion \cite{Chopra2005, LeKhac2020} by optimizing for the difference in scores between our set of human-generated games and a substantially larger set of corrupted (i.e. lower quality) games obtained through random tree-regrowth \cite{Goodman2008} on our dataset (see \autoref{fig:model-overview}b and details in \nameref{methods:fitness-function}). 

This learned fitness function then guides an evolutionary search procedure in order to generate novel games (\autoref{fig:model-overview}c). 
Broadly inspired by work in genetic programming, we use a \textit{quality-diversity} algorithm \cite{Pugh2016, Chatzilygeroudis2020} called MAP-Elites \cite{mouret2015illuminating} to generate a set of samples that widely cover the space of programs in addition to optimizing the fitness function. 
The details of our implementation, including the particular \textit{behavioral characteristics} used for maintaining sample diversity and structure our search of program space, are available in \nameref{methods:map-elites}. 

Our model includes several components that explicitly proxy cognitive capacities, such as features representing physical common sense (estimating predicate feasibility from play data) and recombination operators that explicitly leverage compositionality (the crossover operation that recombines programs). 
We describe a few of these components and how we ablated their contribution to our model in \nameref{methods:ablations}.

\subsection*{Generated games}

\begin{figure}[!bthp]
    \centering
    \includegraphics[width=\columnwidth]{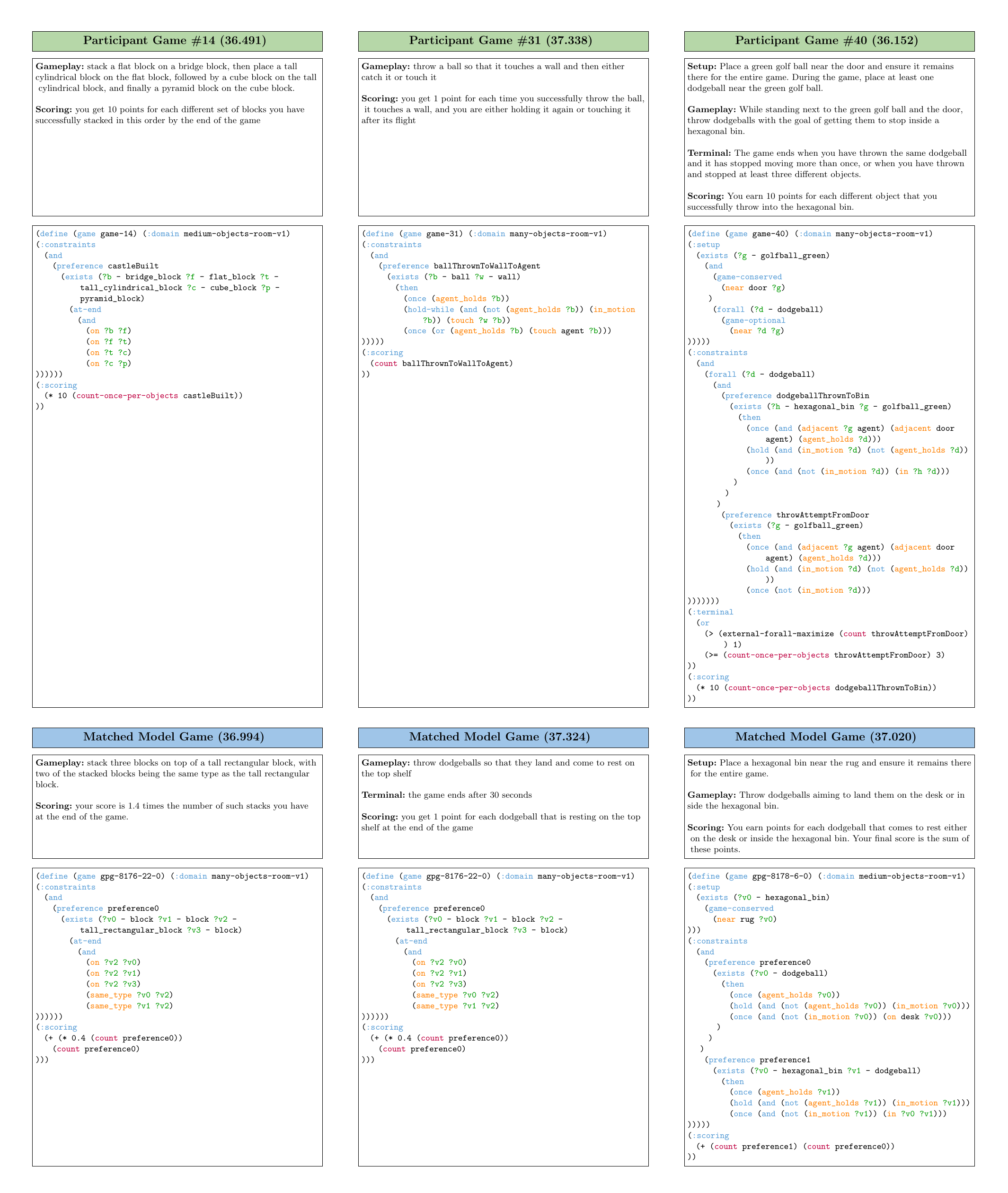}
    \caption{\textbf{Goal Program Generator model produces simple, coherent, human-like games.} Each pair of games in a column has the same set of MAP-Elites behavioral characteristics (a real participant-created game and the corresponding ``matched'' model-generated one). Parentheses: the fitness score assigned by the model to each game. Natural language descriptions are generated through automated back-translation from programs (see \cref{sec:appendix-backtranslation} for details).
    To ascertain that the model-generated programs are distinct from training set examples, we also provide in \autoref{fig:appendix-edit-distance} the most similar real exemplar using an edit distance, and see \cref{sec:appendix-model-sample-edit-distance} for details.
    } 
    \label{fig:model-comparison}
\end{figure}

\begin{figure}[!tbp]
    \centering
    \includegraphics[width=\columnwidth]{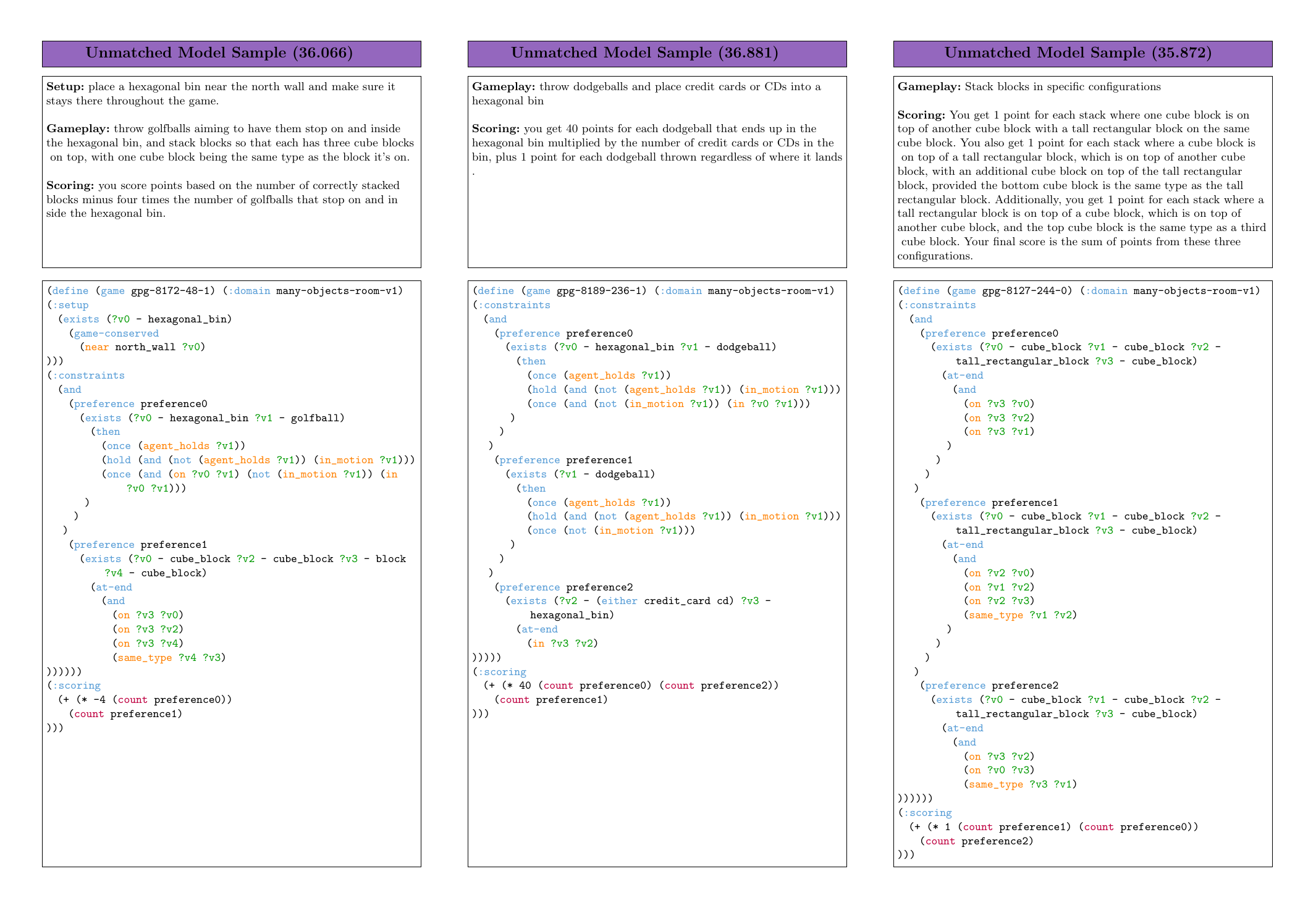}
    \caption{\textbf{Goal Program Generator model produces interesting, novel goals.} Each of the three games below has high fitness and fills an ``unmatched'' cell in the MAP-Elites archive, with no corresponding human game in our dataset. Parentheses: the fitness score assigned by the model to each game.} 
    \label{fig:novel-texts}
\end{figure}

GPG produces a variety of outputs that range from variants of simple games in our reference dataset to games in entirely new regions of program space. In \autoref{fig:model-comparison}, we present examples of model outputs alongside the human-generated games that occupy the same ``niche'' as defined by the MAP-Elites algorithm (see \nameref{methods:map-elites} for details). We call generated games that occupy the same niche as a human game \texttt{matched} and those that don't \texttt{unmatched}.
In the first pair (\autoref{fig:model-comparison}, left), the model proposes an original block-stacking objective: where the human participant created a tower, the model asks to stack three blocks all on the same taller block.
The second and third pairs (\autoref{fig:model-comparison}, middle and right) demonstrate the model's ability to propose throwing games. 
In both cases, the model proposes interesting detailed objectives, some unseen in our training set (e.g., throwing balls onto the top shelf or desk), that match the niche of the participant games by having the same high-level configuration. 
However, the purpose of certain minor elements in generated games tends to be less intuitively obvious (e.g., the scoring condition in the left-most generated game, which arbitrarily multiplies the number of satisfactions by 0.4). 
Our model also produces \texttt{unmatched} games that occupy niches without corresponding human games (\autoref{fig:novel-texts}). 
These include unusual combinations of throwing and block-stacking (\autoref{fig:novel-texts}, left), a game that combines ball throwing and small object placement (\autoref{fig:novel-texts}, middle), and a game that offers a collection of varied block-stacking objectives (all-on-one, a T-shape, and a tower; \autoref{fig:novel-texts}, right).
Though these programs represent creative goals, with preferences that are each individually sensible, their components sometimes fail to combine into a coherent whole (e.g., the golf ball throwing and block placement elements in \autoref{fig:novel-texts}, left, which do not intuitively form a cohesive game). 

Quantitatively, \autoref{fig:results-quantitative} shows that the GPG quickly produces games with fitness scores in the range of human samples and does so across many of the niches defined by our search procedure. 
Of the 2000 programs we report,
1889 programs (94.45\%) exceed the fitness score of the least fit real game, and exactly half (1000) exceed the fitness of the median human game. This demonstrates that our search procedure successfully finds high-fitness samples across much of the range of variation defined by our behavioral characteristics. 
To the extent that our fitness function captures human likeness, our model produces human-like games; we next use human evaluators to extrinsically test our model.  

\subsection*{Human evaluations}
\label{sec:human-evals}
To systematically and extrinsically evaluate our model, we performed human evaluations using a second set of human participants ($n = 100$; see \autoref{fig:methods-human-evals-interface} for the evaluation interface and \nameref{methods:human-evals} for details).
Evaluated games belonged to one of three different categories mentioned above:
\texttt{real} participant-created games from our behavioral experiment, or \texttt{matched} or \texttt{unmatched} model-generated games (see \autoref{fig:model-overview} for category definitions; games in \autoref{fig:model-comparison} and \autoref{fig:novel-texts} were included; see \nameref{methods:human-evals} for details).
Participants evaluated three games in each category above (without knowing their categories) in a randomized order and provided Likert scale ratings on each game for seven measures, including human likeness, fun, and creativity. Our final dataset includes 892 participant-game evaluations, each with a rating for all seven measures.

\begin{table}[!btph]
\centering
\footnotesize
\begin{adjustbox}{center}
\begin{threeparttable}
\caption{\textbf{Mixed model result summary}  }
\label{tab:marginal-means-summary}

\begin{tabular}{lcccccccc}


\toprule
& \multicolumn{8}{c}{\textbf{Pair Comparison}} \\
& \multicolumn{2}{c}{\texttt{Real} $-$ \texttt{Matched}} & & \multicolumn{2}{c}{\texttt{Real} $-$ \texttt{Unmatched} } & &  \multicolumn{2}{c}{\texttt{Matched} $-$ \texttt{Unatched} } \\
\textbf{Measure} & Diff $\pm SE$ & Significance & & Diff $\pm SE$ & Significance & & Diff $\pm SE$ & Significance \\


\midrule

\textit{Understandable} $\uparrow$ & $-0.001 \pm 0.331$ & - & & $1.042 \pm 0.332$ & ** & & $1.042 \pm 0.333$  & ** \\ 
   \textit{Fun to play} $\uparrow$ & $0.143 \pm 0.266$ & - & & $1.020 \pm 0.274$ & *** & & $0.877 \pm 0.273$ & ** \\  
  \textit{Fun to watch} $\uparrow$ & $0.135 \pm 0.250$ & - & & $0.892 \pm 0.259$ & ** & & $0.757 \pm 0.257$ &  **  \\  
  \textit{Helpful}\textsuperscript{\textdagger} $\uparrow$  & $0.016 \pm 0.159$ & - & & $0.251 \pm 0.165$ & - & & $0.236 \pm 0.165$ & - \\ 
  \textit{Difficult} ${\downarrow \atop \uparrow }$ & $-0.200 \pm 0.357$ & - & & $-0.194 \pm 0.165$ & - & & $0.006 \pm 0.361$ & - \\ 
  \textit{Creative} $\uparrow$  & $0.228 \pm 0.310$ & - & & $0.489 \pm 0.316$ & - & & $0.261 \pm 0.314$ & - \\ 
  \textit{Human-like} $\uparrow$ & $0.199 \pm 0.274$ &  - & & $1.396 \pm 0.283$ & ***  & & $1.197 \pm 0.283$ &  ***  \\ 



\bottomrule
\end{tabular}
\begin{tablenotes}
    \item \textbf{Evaluators don't distinguish between participant-created \texttt{real} and \texttt{matched} model games, but do distinguish \texttt{unmatched} games from \texttt{real} (and marginally from  \texttt{matched} ones).} 
    Participants responded to seven Likert questions on a 5-point scale, one for each attribute in the first column (see \nameref{methods:human-evals}). 
    We found fairly low inter-rater agreement (see \cref{sec:appendix-human-evals-analysis}), and so we center our analysis on our fitted mixed-effects models (see \nameref{methods:human-evals}). 
    We use the method of estimated (least-squares) marginal means to compare the three groups of games, accounting for the random effects fitted to particular games and human evaluators. \\
    We report two-sided significance tests adjusted using the Tukey method to control for the multiple difference tests within each attribute, as implemented in the \texttt{emmeans} package.
    See Supplementary \autoref{tab:supplementary-marginal-means} for test statistics and P-values. 
    *: $P < 0.05$, **: $P < 0.01$, ***: $P< 0.001$ \\
    \textdagger: The full measure description is ``Helpful for interacting with the simulated environment.'' \\
    In most measures, higher scores are better, indicated by the $\uparrow$, other than \textit{Difficult}  ${\downarrow \atop \uparrow }$, in which 3 means ``appropriately difficult'', and scores below and above indicate too easy and too hard respectively.
\end{tablenotes}
\end{threeparttable}
\end{adjustbox}
\end{table}

To analyze these results, we performed a mixed-effects regression analysis (we provide the raw score means and non-parametric statistical tests in \autoref{tab:mann-whitney}). 
We fit independent models using each of the seven attributes we asked our human evaluators to judge as the dependent variables. 
We examine two questions: (1) are there any systematic differences between game categories? (2) does our fitness function, learned from corrupting samples in program space, capture any human-evaluated qualities of the games? 
For both questions, we fit mixed-effects models that include a fixed effect for membership in the \texttt{real} and \texttt{matched} groups (treating the \texttt{unmatched} group as a baseline) and random effects for the participants and individual games. 
For the second question, we also include a fixed effect for the fitness score (see \nameref{methods:human-evals} and Supplementary \cref{tab:supplementary-mixed-models} for full details).

To answer the first question, we use the method of estimated marginal means to compare the difference in scores between each pair of categories, averaging out the random effects (\autoref{tab:marginal-means-summary}, and see \cref{sec:appendix-mixed-effect-analyses} for details).
Participants respond similarly to the \texttt{real} and \texttt{matched} games, with no statistically significant differences in the estimated mean scores across all seven attributes.
On the other hand, the \texttt{unmatched} games differ on several attributes.
Participants judge them to be less easily understood and fun to play than \texttt{real} games and less human-like and fun to watch than both \texttt{matched} and \texttt{real} samples. 
We observe similar results using non-parametric statistical tests (\autoref{tab:mann-whitney}).
One potential explanation for the apparent similarity between \texttt{matched} and \texttt{real} games is that the former simply replicate the latter in form and function. 
We examined this question and found that \texttt{matched} and \texttt{real} games have \textit{substantial functional differences} (see summary in \autoref{fig:real-matched-comparison}, details in \cref{sec:appendix-matched-real-similarity}, and methodological details in \nameref{methods:similarity}).

Next, we analyze the mixed-effect models fit with a fixed effect of fitness scores.
First, we replicate the effects of the fitness-less regressions; we continue observing no significant differences between the \texttt{real} and \texttt{matched} groups, and several significant differences between both of those and the \texttt{unmatched} group (see \autoref{tab:supplementary-marginal-means}). 
Next, we examine the fitted coefficients in these regressions (summarized in \autoref{tab:mixed-models}).
We find that our fitness function captures many of the evaluated attributes: higher fitness predicts higher ratings of understandability, fun to play, and human likeness ($\beta_{\text{fit}} > 0$); conversely, higher fitness also predicts lower ratings of helpfulness, difficulty, and creativity ($\beta_{\text{fit}} < 0$). 
Our positive findings are promising: they indicate that our fitness function, learned to maximize human likeness in a symbolic program space, also captures intuitive human notions of understandability and fun. 
Conversely, we view the negative relations as evidence of some degree of mode-seeking: our fitness measure likely assigns the highest scores to the games most representative of the dataset at large.
These modal games are plausibly neither particularly creative nor difficult, which means that participants might also find them less helpful for learning the details of the environment.
To explore this, we highlight the highest fitness games generated both by our model and human participants in \autoref{fig:appendix-highest-fitness}, and observe the type of mode seeking we suggest above (see \cref{sec:appendix-highest-fitness} for details).


We also performed ablations of key model components that explicitly proxy some cognitive capacities we found our participants recruited (see details in \nameref{methods:ablations}).
To ablate \textit{physical common sense}, we remove from our fitness function the two features that estimate the feasibility of a game's preferences by leveraging our database of participant-environment interactions. Analogously, we ablate the intuitive \textit{coherence} we observe in human goals by removing the features that capture the coordination of gameplay elements between different sections. Ablating \textit{compositionality} is more difficult, as our programmatic representation is inherently compositional. We do so by removing the crossover mutation operator used to generate new samples during MAP-Elites, which most explicitly leverages the compositional structure of games. 
In these three ablations, model performance degrades substantially, either in sample fitness scores or in goal plausibility, as estimated using our database of participant-environment interactions.
We also report two other comparisons, one to a model sampling only from the PCFG over our domain-specific language (which performs much worse), and one to a model optimizing a fitness function trained on a subset of our full dataset (which performs comparably and generalizes to the held-out regions of program space).
See \cref{sec:appendix-ablations} for further details.

\section*{Discussion}
Goals are a critical aspect of human cognition and, in fact, the starting place for many models of human behavior.  However, the representation of goals is often impoverished.  In this article, we proposed a new framework for understanding a particular class of human goals as reward-producing programs, as a stepping-stone toward a broader understanding of goal representation and generation.
To evaluate this framework, we developed an interactive experiment in which participants created playful goals, operationalized as games to be played in a virtual environment. 
By analyzing the program-based translation of these games, we highlighted several cognitive capacities recruited by our participants, such as physical common sense and compositionality. These capacities, in turn, informed our modeling efforts.  
We then built a computational model that learns from a small dataset of games and generates coherent, novel goals, where those sampled from partitions of program space occupied by human examples were deemed human-like according to human evaluators.

This work unites various strands of research in cognitive science, artificial intelligence, and game design.
First, we build on substantial literature studying the psychology of goals \cite{Dweck1992, Austin1996, Elliot2008, Molinaro2023, Chu2024} by offering a specific representational hypothesis, in contrast with previous approaches to defining goals. 
We emphasize open-ended goal creation given that generating new exemplars is a core capacity of human conceptual representations \cite{Ward1994} and the utility of games in the study of cognition \cite{Allen2023}.
Our work also relates to goal-conditioned reinforcement learning \cite{Liu2022GCRL}, and we aim to improve on the goal representations used for such agents that tend to lack the variety and richness of human-created goals \cite[ch. 7]{Colas2020review}. 
In this respect, our proposal attempts to abstract from the reward functions and simpler goals used in many reinforcement learning tasks. 
Our goal program interpreter conceptually draws on the notion of reward machines introduced in \cite{ToroIcarte2022}.
Finally, we are inspired by the automatic game design literature, such as synthesizing board game variants \cite{pell1992metagame, hom2007automatic, browne2010evolutionary} or simple video games \cite{togelius2008experiment, smith2010ludocore, zook2014automatic, khalifa2017general}. 
Unlike our approach, these efforts often optimize program synthesis for some heuristic notion of fun \cite{browne2010evolutionary, togelius2008experiment} rather than explicitly modeling human-like game generation.

Our framework is committed to the representation of goals as \textit{reward-producing programs}: computationally executable mappings from behavior to indications of progress towards a goal, which we term ``reward.'' 
We find it crucial that these programs capture the rich, temporally extended nature of goals people create, and that they facilitate the flexible and compositional creation that people seem to engage in \cite{Lake2017, Ward1994}. 
We hope that this proposal is useful to understanding goal representation and generation, not that it losslessly explains every source of variation in human-created and reported goals. 
We note we currently study goal generation through game creation, and while many games have players take on goals \cite{Nguyen2020GamesAgency}, not all goals are fully equivalent or isomorphic to games. 
We believe that our representational hypothesis also has merit for additional kinds of goals, such as the goals created in joint play between multiple children or adults (such as tag or dodgeball), or the objectives a person exploring a new environment might set for themselves (for instance, how to turn on the light at an AirBnB without bumping into anything). 
While we expect the general GPG framework to accommodate such goals in different domains, doing so would certainly entail changes to the specific syntax and semantics of the programmatic representations.
There are, however, types and aspects of goals that might complicate the general procedure of translation into programs. For instance, subjectivity, which may be modeled as listeners forming different representations of the same utterance, might require breaking the assumption that each natural language goal corresponds to a single program.
Similarly, it's not obvious how to represent truly abstract goals like ``I want to do well in school'' as a well-formed program.
In both cases, an avenue forward might be grounding not to a single program but to distributions over programs or programs with stochastic elements (e.g., as suggested in the Rational Meaning Construction framework \cite{WongGrand2023}).
We are excited for future work to continue studying open-ended goal generation in other domains and explore how readily other types of goals map onto programs. 

Our model strongly relies on its approach to sample diversity, which arises from the choice of ``behavioral characteristics'' that define the axes along which the MAP-Elites algorithm maintains diversity. 
In this work, we select behavioral characteristics based on notable gameplay components observed in our human dataset; future work could explore other techniques for maintaining diversity, including the automated selection of behavioral characteristics \cite{Cully2019, Grillotti2021}. 
Our current features approximating intuitive physical common sense are indirect, using participant interactions with the environment to estimate feasibility.
Future approaches could integrate planning or physical simulation to improve our model's understanding of physics \cite{Ullman2017, Chen2023}.
Our model is currently limited to a single kind of common sense, the intuitive physical one; other environments may require leveraging similar knowledge from other domains, such as intuitive social models of agency and theory of mind.
Finally, our model is inherently coupled to the environment and dataset we collected --- particularly given the engineering effort to instantiate various types of knowledge. 
This approach has some distinct advantages: we can isolate various cognitive capacities, interpret their contribution to our fitness measure (\cref{sec:appendix-predictive-features}), and ablate their roles (\cref{sec:appendix-ablations}). 
Simultaneously, some of the challenges our model faces (such as coherence between program components) might be alleviated by incorporating natural language or by leveraging the capabilities of large language models to write code and adapt to in-context instructions. 
Language models could also alleviate our current reliance on manual translations from participant game descriptions to the proposed mental language of goal programs (see \cite{WongGrand2023} for a discussion on using language to construct meaning through programs, and \cite{Tang2024} building programs to act as world models).

We see two particularly promising ways in which our representational framework could be used going forward. First, there is increasing interest in building artificial agents that can flexibly explore and generalize across environments \cite{Reed2022GATO, Gallouedec2024JAT}. 
The \textit{autotelic} perspective argues that empowering agents to propose and pursue self-generated goals is a fruitful way to improve their ability to generalization \cite{Colas2020review}.
However, goals in such systems are often derived from agent or object positions \cite{florensa2018automatic, team2021open}, short natural language descriptions \cite{Du2023, Colas2023}, or limited temporally-aware mechanisms \cite{Littman2017, Leon2022} --- all impoverished when compared with the diverse goals humans flexibly create.
Closely related to our notion of representing goals by programs, recent work proposes to directly synthesize reward functions \cite{Ma2023Eureka} or environment descriptions \cite{Faldor2024omni-epic} using code-generation models. 
We are excited for future work to empower artificial agents with richer goals that reflect human-like novelty and difficulty, for two specific reasons.
First, we believe access to complex and varied goals would enable agents to learn flexible representations of their environments that support higher behavioral adaptability \cite{Chu2024}.
Second, we view compositional goal production as facilitating effective exploration of unseen goals \cite{Colas2020} (and see \cite{Wu2018} for a discussion of generalization and exploration).
We also note our current approach estimates goal fitness without considering additional higher-level objectives that might guide goal generation. 
Prior literature offers curiosity \cite{Ten2022,Berlyne1950}, empowerment \cite{Gopnik2024,Addyman2013,Du2023human}, information gain \cite{Ruggeri2024, Liquin2021}, novelty \cite{Taffoni2014,Berlyne1950}, and learning progress \cite{Ten2021,Baldassarre2014} as compelling potential objectives.
Future work could instantiate goal generators that consider these objectives as auxiliary terms to the fitness function and compare the behaviors that arise in artificial agents through pursuing them.  


If we are to understand goals as programs, our proposed framework may also help advance our understanding of intuitive psychology and goal inference \cite{Spelke2007, Jara-Ettinger2016, Liu2019}.
Previous work proposed that our ability to understand other people's goals, as part of our Theory of Mind, operates through inverse reinforcement learning: inferring an agent's reward from observing their behavior \cite{Jara-Ettinger2019IRL}.
Many prior approaches eschew goals entirely, using some function approximator (e.g., a neural network) to estimate reward, resulting in an uninterpretable estimator that can struggle to generalize \cite{Arora2021IRLSurvey}.
We envision leveraging our goal programs as a prior distribution for a Bayesian Theory of Mind \cite{Baker2011BToM} approach, scaling up previous approaches that relied on a small number of predefined goals \cite{Velez-Ginorio2017}, to create models that would parse an agent's behavior and provide an interpretable, semantically explicit estimate of their goal \cite{HoGriffiths2022}. 
Applying our framework to either of these proposed problems would offer a substantial long-term challenge building on the work we present in this article. 
Nevertheless, we see an exciting prospect to leverage this approach to improve the understanding of human goals and endow machines with human-like goal concepts and capabilities.

\section*{Acknowledgements}
Author GD thanks members of the Human and Machine Learning Lab and the Computation \& Cognition Lab at NYU for their feedback at various stages of this project. 
Authors GD and BML are supported by the National Science Foundation under NSF Award 1922658.
Author GT is supported by the NSF GRFP.
Author TMG's work on this project is supported by NSF BCS 2121102.

\clearpage
\section*{Methods}

\setcounter{figure}{0}
\renewcommand\thefigure{M-\arabic{figure}}
\setcounter{table}{0}
\renewcommand{\thetable}{M-\arabic{table}}

\subsection*{Dataset collection methods}
\label{methods:dataset}
\textbf{Experimental design:} 
This study was performed under NYU IRB ``Active Learning'' under principle investigator Todd M. Gureckis. 
After an informed consent form and instructions quiz, participants completed a tutorial designed to familiarize them with the controls for our environment. 
After successfully completing the tutorial, participants were randomly assigned to one of three variations of the main experiment room, with the same structure but different amounts of available toys and objects. 
Participants were then free to explore this new room until they had a game ready, and could freely reset it to its initial state in the meantime. 
Participants were asked to create games with the following restrictions: single-player, require no additional space or objects that they do not see in the room, and include a scoring system.
While the latter constraint may seem limiting, we note that any arbitrary goal can be scored by rewarding the achievement of the goal. 

\textbf{Dataset collection:} Participants then reported their game in natural language in three text boxes, one of which was optional (see \autoref{fig:experiment-interface}). The optional first one allowed specifying whether there was any setup or preparation required to get the room from its default initial state to one that would allow playing the game (e.g., placing the bin on the bed). 
The second text box allowed participants to describe the game's gameplay, and the third offered space to describe the scoring rules. 
To encourage participants to imagine playing their game, they were also asked to report their perceived difficulty level and how many points they thought they might score if they played it. 
Participants then had a chance to play their game and revise it should they want to; if participants opted to revise their games, we analyzed the revised ones.
We contacted 192 participants via Prolific \cite{Palan2018prolific} of whom 114 finished the experiment and another 12 were paid due to technical difficulties.
Participants were paid a base rate of \$10 and received a \$2 bonus if their game satisfied the required constraints. Successful participants took 44.4 minutes on average, with a standard deviation of 23.3 minutes. 
We then excluded 8 games that did not satisfy the constraints we posed on participants, 6 duplicates (including some due to technical difficulties from participants who restarted the experiment), and 6 other games that were unclear or under-specified.
After accounting for two other games we opted to avoid modeling due to their complexity (one referring directly to the game interface and controls, and another describing several games or levels in the single description we collected), we arrived at our final dataset of 98 games.
We acknowledge the potential arbitrariness of manually translating from natural language to our program representations; we attempted to be maximally faithful to the descriptions and excluded participants whose games required too much subjectivity or interpretation. 

\textbf{Interaction traces:} In addition to the game descriptions in natural language, we record traces of participants' interactions with the environment.
We record state-action traces to allow us to replay and examine how participants interact with our environment. 
We record separate traces for each different segment of the experiment (before creating the game; while reporting their game; playing their game; after editing their game), and for each time the participant resets the environment within each segment.
We end up with 382 total such traces. 
Our primary use for them is in implementing a ``reward machine,'' an interpreter for our goal programs, which parses a goal program into a state machine, and iterates through a trace to emit the score of that trace under the goal. 
We use a limited version of this in our fitness features (see \nameref{methods:fitness-function} for additional details) and in some of our model evaluations and ablations (see \cref{sec:appendix-ablations} for additional details).

\textbf{Natural language to domain-specific language translation:}
We manually translated the games we collected from participants to programs in a domain-specific language (DSL) we created. 
We examined the natural language descriptions our participants provided to identify recurring semantic components, which we then mapped onto elements in our DSL, iterating between translating more programs and updating the DSL grammar.
We began by attempting to translate directly into PDDL (the Planning Domain Definition Language, \cite{Ghallab1998}), which offers a basic representation for specifying planning problems, but deviated from it as we encountered game elements our participants specified with no clear PDDL analogs. 
We assume the translation process is not lossless, as there are likely multiple natural language descriptions for each underlying set of game semantics and multiple programmatic encodings of vague natural language descriptions; however, we aimed to develop representations that capture the core semantics of the rich, generative, and creative structure in goals. 
We also perform some analyses to validate the extent to which these translated programs capture semantic concepts that were intended by participants, which we report in \cref{sec:appendix-nl-to-dsl}.

\textbf{Goal program interpreter:}
Inspired by the ``reward machine'' proposed by \cite{ToroIcarte2018RewardMachines}, we similarly implement an interpreter for the goal programs in our domain-specific language.
The interpreter parses a program in our DSL into a state machine.
This state machine enumerates over environment states and participant actions emitted as a participant plays in our experiment (see \textbf{Interaction traces} above), tracks the participant's progress with respect to each program component (setup conditions, gameplay preferences, and terminal conditions), and emits a reward according to the scoring conditions defined in the goal program. 
This allows us to ground programs to participant interactions and evaluate partial or complete fulfillment of the specified goal.
We use this reward machine as part of our feature set (see \nameref{methods:fitness-function}), to analyze functional similarity between programs our model generates and participant-created games (see \autoref{fig:real-matched-comparison}), and to assess our manual translations of participant-provided descriptions (see \cref{sec:appendix-nl-to-dsl}).
Our current implementation of the interpreter covers the vast majority of predicates and grammar elements; we omitted grounding a few rarely-used predicates due to their complexity and lack of frequency. 
In these cases, we attempted to ensure our implementation would be biased towards false negatives rather than false positives --- we would rather fail to count an interaction that occurred than count interactions that did not occur. 

\subsection*{Game dataset analyses methods}
\label{methods:dataset-analyses}
\textbf{Common sense through predicate role-filler analysis:} 
We analyze predicate role-filler occurrences, coarsening individual objects to higher-level categories (see the legend on the right of \autoref{fig:behavioral-composite}b). 
To split between the two panels of \autoref{fig:behavioral-composite}b, we categorize each game by whether it includes the following motifs: throwing (e.g., balls into a bin), stacking (e.g., blocks in a building), organizing (e.g., placing objects in specified places), or other. 
We split the figure into games involving only throwing motifs (left panel) and games involving any other motifs, potentially in addition to throwing (right panel). 
In games involving only throwing (left panel), participants most often refer to balls, primarily checking whether or not the agent holds a ball or a ball is in motion (as part of quantifying the act of throwing). 
Other predicates are often used to specify some additional conditions on throwing (such as specifying the bin being on the bed or the agent being next to the desk) and are used with a variety of object categories.
Conversely, in games involving other elements (right panel), we see blocks and the generic ``any\_object'' being used far more often, mostly in various placement and stacking constraints. 

\textbf{Compositionality and creativity through abstract structure occurrence:}
We analyze how often participant games make use of various grammatical structures to showcase both compositional reuse and long-tail creativity.
Each structure involves a temporal modal (such as \lstinline{once} or \lstinline{hold}) and the predicate expression nested under it, such as \lstinline{(once (agent_holds ?b) )}, where \lstinline{?b} is a variable quantified earlier. 
We count structures, abstracting away specific variables and their types -- so the expression above would be coarsened as \lstinline{(once (agent_holds <obj>) )}, and would be counted together with any other expression coarsened to this pattern. 
We encounter a total of 126 unique expressions in our dataset, the most common one with 62 occurrences being \lstinline{(hold (and (not (agent_holds <obj>) ) (in_motion <obj>) ) )}, which maps loosely to ``find a sequence of states where an object is not held and is in motion'' --- that is, is currently moving with the agent touching it, for instance while being thrown or rolled. 
Of the 126 expressions, exactly half (63) occur only once. 

\subsection*{Fitness function methods}
\label{methods:fitness-function}

\textbf{Fitness function form:} We use the most direct mapping from feature values to a real-valued score as our fitness function: a learned, weighted linear combination of a set of features extracted programmatically from each game that is optimized to assign high scores to ``human-like'' games and low scores to everything else. It is a function $f : \mathcal{G} \rightarrow \mathbb{R}$ that maps individual games $g \in \mathcal{G}$ to real-valued scores: $f(g) = \theta \cdot \phi(g)$, where $\theta$ is a learned vector of weights and $\phi: \mathcal{G} \rightarrow [0, 1]^{F}$ is a feature extractor. 

\textbf{Feature extractor and feature set:} The feature extractor $\phi$ represents each game as an 89-dimensional vector (i.e. $F = 89$). Each entry in the vector corresponds to a particular structural or semantic property of the game, from the size and depth of the syntax tree to the apparent feasibility of the game's preferences. We normalize the values of each property to fall within the unit interval by using the observed range of values in our dataset. Many features used in the fitness function are directly computable from the DSL representation of a game (for instance, properties of its syntax tree or the misuse of particular grammatical structures). While these features represent the majority of the 89 features used, we also implement two important sets of features that require additional computation.  

The first of these are $n$-gram features that capture the mean log score of the game under a simple $n$-gram language model trained over the set of human-generated syntax trees.
We fit n-gram models using stupid backoff \cite{Brants2007} to account for missing n-grams, using the default discount factor of 0.4 reported in \cite{Brants2007}.
We compute these scores separately for each game section (i.e. setup, preferences, terminal, and scoring) and also for the game overall, resulting in 5 features.

The second set consists of two features that make use of an interpreter that parses game programs into ``reward machines'' \cite{ToroIcarte2018RewardMachines}: finite-state machines that process a trace of player inputs and emit a reward whenever the particular scoring conditions of the game are met. The interpreter programmatically implements each of the predicates in the DSL, which allows us to construct a dataset of which objects were used to satisfy which predicates across our dataset of 382 human \textit{play traces}. The two features query this database in order to get an approximate common sense measure of a game's ``feasibility,'' computing the proportion of a game's predicate-argument combinations that have been satisfied by human players in our dataset (one feature does this for individual predicates, while the other does this for boolean logical expressions over predicates). While these feasibility measures give a sense of whether the objectives of a game can be completed in the physical reality of the simulation, the limited nature of our play trace dataset means they are far from perfect proxies.

We developed our feature set starting from features used in similar prior work (e.g., features representing the length and depth of the syntax tree \cite{Rothe2017}). 
We then fit a fitness function using the procedure described below and inspected the fittest games from our set of negative examples. 
We iteratively added features to account for mistakes our model made (flawed negatives with high fitness) and removed features that our fitness function seemed to ignore (by learning a weight with a low magnitude). 
The complete set of features used (and accompanying descriptions) is available in \cref{sec:appendix-features}, with the most important features (by their learned weights) highlighted in \cref{sec:appendix-predictive-features}.

\textbf{Fitness function learning algorithm:} To learn the weight vector $\theta$, we take inspiration from the contrastive learning of energy-based models \cite{Chopra2005} with the objective of separating a set of \textit{positive} examples (our set of human-generated games) from a set of \textit{negative} examples (and see a summary in \autoref{fig:model-overview}b). To learn an effective fitness function, these negatives must be qualitatively worse than our set of human games without being trivially distinguishable from them. We generate a set of plausible negatives by \textit{corrupting} games from our positive set. To corrupt a game, we select a random node in its syntax tree, delete the node and its children, and randomly re-sample a sub-tree according to the DSL grammar (illustrated in red in \autoref{fig:model-overview}b). This ``tree-regrowth'' approach \cite{Goodman2008} generally produces sub-trees that are syntactically valid but semantically ``out-of-place,'' with the severity of the corruption tending to correspond to the height of the re-sampled node in the syntax tree. To account for the variance in the difficulty of distinguishing between a given positive and negative example, we generate a large set of negatives: 1024 for each of the 98 positives, for a total of 100,352 negatives. 

We train the fitness function (i.e. optimize $\theta$) using a softmax loss, not unlike the MEE loss used to train energy-based models \cite{LeCun2006} or the InfoNCE loss \cite{VanDenOord2018}.
For a positive example $g^+$ and a set of negative examples $\{ g^-_k \}, k \in \{1, 2, \cdots, K \}$, we assign the loss:

\begin{equation}
    \label{eq:loss}
    \mathcal{L}(g^+, \{ g^-_k \}_1^K ; \theta) = - \log \frac{\exp (f_\theta(g^+))}{\exp (f_\theta(g^+)) + \sum_{k=1}^K \exp ( f_\theta(g^-_k )) }
\end{equation}

This loss encourages the model to assign higher fitness scores to the real games than the negative examples. Simultaneously, this loss provides a diminishing incentive to push negative fitness scores down as the distance between the positives and negatives increases, intuitively assigning higher loss to negative examples with fitness closer to the positive example's fitness. See \cref{sec:appendix-objective-function-algorimth} for full details of our training and cross-validation setups. 

\textbf{Final fitness function:} Note that while we perform cross-validation for hyperparameter selection, once we fixed a set of fitness features and hyperparameters, we fit a final fitness function using our entire dataset (98 participant-created examples and their corresponding negatives). 
Given the minuscule human dataset we collected, we opted against holding out data from the final objective function to best guide our model's search process (though see \autoref{sec:appendix-held-out} for a comparison to fitness function trained on a subset of our dataset). 

\subsection*{MAP-Elites methods}
\label{methods:map-elites}
\textbf{MAP-Elites overview:} MAP-Elites is a population-based, evolutionary algorithm that works by defining a set of \textit{behavioral characteristics}: discrete-valued functions over genotypes (in our case, game programs in the DSL) that form the axes of a multi-dimensional \textit{archive} of cells (and see an overview in \autoref{fig:model-overview}c).
At each step, a game $g$ is selected uniformly from among the individuals in the archive (\autoref{fig:model-overview}c, step 1) and mutated to form a new game $g'$ (\autoref{fig:model-overview}c, step 2) . The mutated $g'$ is evaluated both under the fitness function $f$ and each of the $n$ behavioral characteristics $b_i:\mathcal{G} \to \{0, \dots, k_i\}$ in order to determine which cell $c = [b_1(g), \dots, b_n(g)]$ it occupies. If the cell is unoccupied, then $g'$ enters the archive. Otherwise, it enters the archive (and replaces the previous occupant) only if its fitness is greater than the current occupant of the cell (\autoref{fig:model-overview}c, step 3). In this way, the algorithm maintains an ``elite'' for each possible combination of values under the behavioral characteristics. 

\textbf{Behavioral characteristics:} Inspired by prior work on using MAP-Elites for procedural content generation \cite{charity2020mech}, we define a set of integer-valued behavioral characteristics that each indicate how many preferences in each archive game match one of nine archetypal exemplar gameplay preferences (illustrated as the axes of the grid in \autoref{fig:model-overview}c). 
These include several types of ball-throwing preferences, as well as ones capturing block-stacking, object-sorting, and other miscellaneous activities.
We also include two other features: one that captures whether or not the game includes a setup component and one that captures the total number of preferences. 
For additional details and descriptions of the exemplar preferences, see \cref{sec:appendix-map-elites}. 
We use nine exemplar preferences, in addition to these two other features, as a trade-off between covering many behaviors that participants demonstrate and avoiding exploding the size of the archive: as it is, the 11 total behavioral characteristics result in a total archive size of 2000 games. 
The 98 participant-created games in our dataset map onto 47 different archive cells; conversely, most archive cells (1953, or $97.65\%$) have no corresponding participant-created exemplar.

\textbf{Auxiliary coherence check:} 
We include an auxiliary ``pseudo behavioral characteristic'' that explicitly captures a few general coherence properties of games, which we use to help our model search the space of programs. 
This characteristic computes a conjunction of the values of 21 features, ones that we either expect \textit{all} plausibly human-generated games to either exhibit or \textit{none} of them to exhibit (indeed, all participant-created programs in our dataset pass this check).
These include features such as checking that all quantified variables are referenced at least once, that all game preferences defined are mentioned in the terminal or scoring conditions, 
that no logical expressions are tautological or redundant.
This check does not use any information beyond the fitness features and serves as a mild additional inductive bias and structure for our search process.

We keep two copies of the 2000-sample archive from the behavioral characteristics using the exemplar preferences above, one with samples passing this auxiliary check and the other with samples failing it. 
During the search process, we sample uniformly from both archives. 
Intuitively, this accomplishes two desiderata: (1) it forces the model to generate a sample in each archive cell that passes this check, and simultaneously, it (2) allows the model to better search the space of programs by also exploring high-fitness samples that fail this check.
We consider as outputs of our final model only goals from the archive copy that pass this check, and those are the only ones we report in fitness-based and human evaluations.
See \cref{sec:appendix-map-elites} for additional details. 

\textbf{Mutation operators:} To mutate a game, we randomly select an operator from among the following: \textbf{regrowing} a random node and its children in its syntax tree, \textbf{inserting \& deleting} the child of a node with multiple potential children, \textbf{crossing over} with the syntax tree of another randomly-selected game, \textbf{resampling the variables, initial conditions, or final conditions} used by a preference, and \textbf{resampling the optional game sections} (i.e. setup and terminal conditions).  We seed the initial archive by naively sampling candidates from the PCFG---\textit{not} with real, human-participant-created games or corruptions thereof that were used to train the fitness function. 
Further details of the algorithm are available in \cref{sec:appendix-map-elites}.

\textbf{Archive initialization:} Our search process is not seeded from any real participant-created examples. Instead, we initialize the MAP-Elites archive with examples generated by sampling from the PCFG defining our domain-specific language. We generate 1024 initial samples, sort them by their fitness scores, and add at most 128 of them to the archive. See \cref{sec:appendix-map-elites} for additional details.

\subsection*{Ablation methods}
\label{methods:ablations}
We ablate several components of our model that leverage cognitive capacities people appear to use when creating goals. 
We describe the components and briefly elaborate on their respective cognitive capacities below, and  report the full ablations in \cref{sec:appendix-ablations}:

\textbf{Common sense:} We offer evidence in \autoref{fig:behavioral-composite} and our discussion of the Behavioral results that participants seem to leverage (physical) common sense reasoning in their goal creation.
The domain-specific language we use to represent goals is underconstrained with respect to this type of common sense and allows to generate expressions that are physically improbable or entirely impossible. 
To aid our model in generating physically plausible expressions, we include two fitness features that query a dataset of participant interactions with our environment (see \autoref{methods:dataset}) and score predicate expressions on whether or not any participants ever satisfied them in their play behavior. 
We report the results of this ablation in \cref{sec:appendix-ablations-common-sense}, where we find that these features are crucial for our model. 

\textbf{Compositionality:} We offer evidence of the way participants appear to recombine simple elements to create diverse games in \autoref{fig:behavioral-composite}. 
Compositionality is core to our domain-specific language, as programs naturally offer the ability to compose expressions of the same type.
We ablate this ability by removing the mutation operators that implement compositions.
We first remove some of our custom resampling operations and then remove the crossover operation, which explicitly composes two programs in our archive to create two new candidates (see \autoref{methods:map-elites} for additional details).
We report the results of this ablation in \cref{sec:appendix-ablations-compositionality}, where we observe that the crossover operation is crucial for our model and that our custom operators offer a smaller though measurable effect.

\textbf{Coherence:} We observe that most participants create coherent goals that ``fit together'' without any explicit prompting to do so: different components of a goal tend to refer to one another and avoid disjointedness. 
After earlier versions of our model struggled with this type of higher-level coherence, we included several fitness features that attempt to measure it at different degrees of abstraction (see \cref{sec:appendix-ablations-coherence} for additional details and the full results).
We find that including these features substantially improves the model-generated games. 

\textbf{PCFG-sampling only baseline:}
To illustrate the necessity of a complex search process over the space of programs in our DSL, we created a baseline that repeatedly samples from the PCFG representing our grammar, with rule and terminal counts fitted to our human datasets.
We match the total number of samples to the total number of candidates our full model generates in its search.
We find both low occupancy rates (sampling from this prior fails to explore the space) and low fitness scores.
See \cref{sec:appendix-prior-only-baseline} for additional details and the full results.

\textbf{Held-out data ablation:}
We perform a held-out evaluation of our model to evaluate how robust our procedure is to unobserved data.
We split our dataset of 98 games into 20 test games and 78 training games and fit the fitness function only using those games (with the same set of fitness features as our full model). 
We then run our search to optimize the fitness function fitted to the partial data. 
We find the results comparable to our full model, both in overall fitness scores and when particularly examining the model-generated games corresponding to the held-out samples.
See \cref{sec:appendix-held-out} for additional d details and the full results. 

\subsection*{Human evaluation methods}
\label{methods:human-evals}

\textbf{Evaluation dataset:} We select games to be evaluated using the following procedure:
\begin{easylist}
@ \texttt{real}: We include 30 participant-created games, each with a different set of behavioral characteristics --- that is, each being considered `different' according to how our model searches through the space of games (see \nameref{methods:map-elites}) for additional details).

@ \texttt{matched}: For each of the \texttt{real} games included above, we include the model-generated game from our final model from the corresponding MAP-Elites archive cell. 
Each of these games includes the same number of gameplay preferences as the corresponding \texttt{real} participant-created games, matching the same exemplar preferences.

@ \texttt{unmatched}: We also include 30 additional games from our model's archive. 
We sample these in a fashion that aims to be balanced across the different preference counts and usage of the different exemplar preferences.
That said, given that human games cover only 47 out of the 2000 archive cells, that leaves 1953 potential \texttt{unmatched} games to sample; it is difficult to know how representative our set of 30 (which is about 1.5\%) is. 
We initially sampled 40 \texttt{unmatched} games and had participants evaluate 4. We then discovered that some of these model samples have drastically lower fitness scores from the \texttt{real} and \texttt{matched} samples.
We therefore excluded evaluations of the 10 lowest-fitness \texttt{unmatched} samples from our analyses to reduce the degree to which fitness scores confound our analyses.  

\end{easylist}

We collected evaluations from $n = 100$ human participants, and our final dataset includes 892 participant-game evaluations, of which 300 are in the \texttt{real} category, 300 in the \texttt{matched} category, and 292 in the \texttt{unmatched} category (due to the exclusions mentioned above). 

\textbf{GPT-4-based back-translation:} 
Rather than ask participants to interpret our domain-specific language, we use the GPT-4 \cite{OpenAI2023GPT-4} language model to perform a multi-step back-translation from programs in our domain-specific language to structured natural language. 
For fairness and consistency, we use this procedure on the \texttt{real} games in addition to the model-generated \texttt{matched} and \texttt{unmatched} games.
We first apply a rule-based system to apply templates, translating expressions in the DSL to natural language sentence fragments. 
We then use GPT-4 to first map the templated fragments to a more natural language, and then to combine the description of each game component (setup, gameplay preferences, terminal conditions, and scoring rules) to a short coherent description.
See \cref{sec:appendix-backtranslation} for full details and prompts used. 

\textbf{Human evaluations structure:}
\autoref{fig:methods-human-evals-interface} presents our human evaluation interface. 
Following instructions and an understanding quiz, participants evaluated nine total games: 3 \texttt{real} ones, the corresponding 3 \texttt{matched} ones, and 3 \texttt{unmatched} ones. 
Participants were presented one game at a time and provided two short textual responses, one explaining the game in their own words, and one providing a short overall impression of the game.
Participants also answered seven Likert-type questions on 5-point scales, answering the following questions about the italicized attributes:
\begin{easylist}
@ \textit{Understandable:} ``How confident are you that you understand the game described above?'', where 1: not at all confident, 3: moderately confident, and 5: very confident
@ \textit{Fun to play:} ``How fun would it be to play the game yourself?'', where 1: not at all fun, 3: moderately fun, and 5: extremely fun.
@ \textit{Fun to watch:} ``How fun would it be to watch someone else play this game?'', where 1: not at all fun, 3: moderately fun, and 5: extremely fun.
@ \textit{Helpful:} ``Imagine that you played this game for several minutes. How helpful would it be for learning to interact with the virtual environment?'', where 1: not at all helpful, 3: moderately helpful, and 5: extremely helpful.
@ \textit{Difficult:} ``Imagine that you played this game for several minutes. Do you think it would be too easy, appropriately difficult, or too hard for you?'', where 1: far too easy, 3: appropriately difficult, and 5: far too hard.
@ \textit{Creative:} ``How creatively designed is this game?'', where 1: not at all creative, 3: moderately creative, and 5: extremely creative.
@ \textit{Human-like:} ``How human-like do you think this game is?'', where 1: not at all human-like, 3: moderately human-like, and 5: extremely human-like. 
\end{easylist}

\textbf{Evaluation statistical analyses:} 
For each attribute and each game category (\texttt{real}, \texttt{matched}, and \texttt{unmatched}, we report the mean score assigned by all participants to games in that category for that attribute.
We then also aggregate these attribute scores by category and report a nonparametric Mann-Whitney $U$-test \cite{MannWhitney1947} for differences in outcomes, as appropriate for ordinal data.
See Supplementary \cref{tab:supplementary-mann-whitney} for the full table including test statistics and P-values. 
Significance results were highly similar when computing two-sample $t$-tests as an alternative statistical test. 
We do not perform any adjustment for multiple comparisons but note that most effects discussed would remain significant at the $\alpha = 0.05$ level under a standard Bonferroni correction. 

\textbf{Mixed effect models:} 
We are interested in modeling the relationship between the scores predicted by our fitness function and the attributes human evaluators predicted. 
To that end, we set up mixed effect regression models \cite{RaudenbushBryk2002,Hox2018}.
We fit separate models for each measure as the dependent variable, regressing a continuous latent score (e.g., $s^i_{\text{fp}}$ for the fun-to-play measure, equation (2) below).
We include fixed effects  for membership in the \texttt{real} ($\mathbbm{1}^i_{\text{real}}$) and \texttt{matched} ($\mathbbm{1}^i_{\text{matched}}$) groups, treating the \texttt{unmatched} group as a baseline.
In our second analysis, we also include a fixed for the fitness score ($x^i$) (which is the full form reported in equation (2) below). 
We include random effects for the individual participants ($\epsilon^{p_i}_{p} \sim \mathcal{N}(0, \sigma_p^2)$) and evaluated games ($\epsilon^{g_i}_{g} \sim \mathcal{N}(0, \sigma_g^2)$). 
We also fit a sequence of cut-points (equation (3)) that transform the latent score to the observed ordinal rating $y^i_{\text{fp}}$ (equation (4)). 
We suppress the subscript for each measure below:
\begin{align}
    s^i &= \beta_{\text{fit}} x_i + \beta_{\text{real}} \mathbbm{1}^i_{\text{real}} + \beta_{\text{matched}} \mathbbm{1}^i_{\text{matched}} + \epsilon^{p_i}_{p} + \epsilon^{g_i}_{g} + \epsilon^i, &  \epsilon^i &\sim \mathcal{N}(0, \sigma^2) \\
    -\infty &\equiv c_0 < c_1 < c_2 < c_3 < c_4 < c_5 \equiv \infty \\
    c_{k - 1} &< s^i \leq c_k \Rightarrow \ \text{observe} \ y^i = k
\end{align}
Models without either random effect performed worse than the full model, so we report results including both random effects. 
We fit cumulative link models for ordinal regression \cite{Agresti2002,GreeneHensher2010} using the \texttt{ordinal} package \cite{OrdinalPackage} in \textbf{R} \cite{RSoftware}. 
We report coefficient significance estimates using the two-sided Wald test, as implemented in the \texttt{ordinal} package. 
The results of these mixed-effect models are summarized in \autoref{tab:mixed-models} (and see \cref{sec:appendix-mixed-effect-analyses}, \autoref{tab:supplementary-mixed-models}, and \autoref{tab:supplementary-marginal-means} for additional details).

\textbf{Marginal means comparisons:}
To compare between the three categories we evaluate (\texttt{real}, \texttt{matched}, and \texttt{unmatched} games), we use the method of estimated (least-square) marginal means.
This allows us to account for variations in the random effects fitted to individual evaluation participants and evaluated games. 
In the models fitted with fitness scores, these similarly allow accounting for variations in observed fitness scores between game categories and their predicted effect on the ratings.
Intuitively, the method simulates the marginal means of the dependent variable as though we had observed each combination of fixed effect (fitness score) and random effects (for individual raters and games) for all values of the group of interest (game type), allowing us to compare its effect most directly. 
We use the \texttt{emmeans} package \cite{EmmeansPackage} to estimate the mean score for each attribute in each category. 
We also report standard errors (of the differences in estimated means) using the \texttt{emmeans} package, and two-sided significance tests adjusted using the Tukey method (to control for the multiple difference tests within each attribute).

\subsection*{Sample similarity comparison methods}
\label{methods:similarity}
For each \texttt{real} game and its corresponding \texttt{matched} game from those included in the human evaluations, we examine which of our recorded participant interactions (see \nameref{methods:dataset} above) fulfills one or more gameplay elements. 
We treat the setup (if specified) and each gameplay preference as a gameplay element --- our aim here is to quantify which participant interaction traces `play' a part of the game. 
We do this using our ``reward machine'' --- our implementation of an interpreter for goal programs in this domain-specific language (see \nameref{methods:dataset}). 
For each pair of games, we then check which particular interactions either (a) `play' parts of both games, (b) only fulfill components in the \texttt{real} game, or (c) only fulfill components in the \texttt{matched} game.
We color these proportions in purple, green, and blue, respectively in \autoref{fig:real-matched-comparison}


\FloatBarrier
\clearpage

\section*{Data availability}
All data for our study, including raw participant responses in the behavioral experiment, their translations to programs in our domain-specific language, and the specification for the domain-specific language, are available on GitHub at \url{https://github.com/guydav/goals-as-reward-producing-programs/}.

\section*{Code availability}
All code for our study, including code used to analyze and generate figures for our behavioral experiment, and the full implementation of our Goal Program Generator model, are available on GitHub at \url{https://github.com/guydav/goals-as-reward-producing-programs/}.

\clearpage


{
\small

\bibliographystyle{unsrtnat}
\bibliography{game_generation}

}

\clearpage
\section*{Extended Data}
\setcounter{figure}{0}
\renewcommand\thefigure{ED-\arabic{figure}}
\setcounter{table}{0}
\renewcommand{\thetable}{ED-\arabic{table}}

\begin{figure}[!bthp]
    \centering
    \includegraphics[width=\columnwidth]{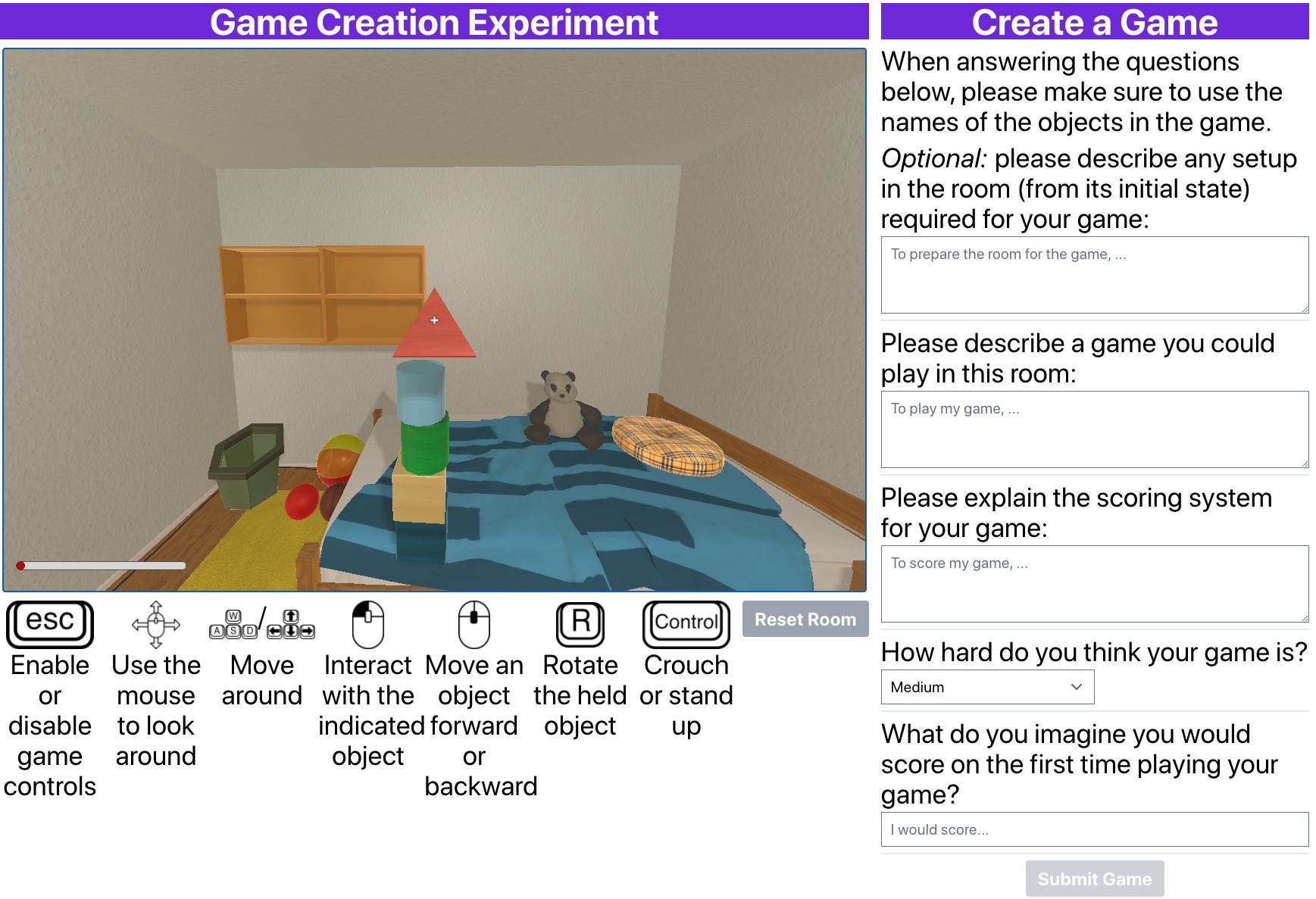}
    \caption{\textbf{Online experiment interface.} The main part of the screen presents the AI2-THOR-based experiment room. 
    Below it, we depict the controls. To the right, we show the text prompts for creating a new game (fonts enlarged for visualization).
    Our experiment is accessible online \href{https://game-generation-public.web.app/}{here}.} 
    \label{fig:experiment-interface}
\end{figure}

\begin{figure}[!bthp]
    \centering
    \includegraphics[width=\columnwidth]{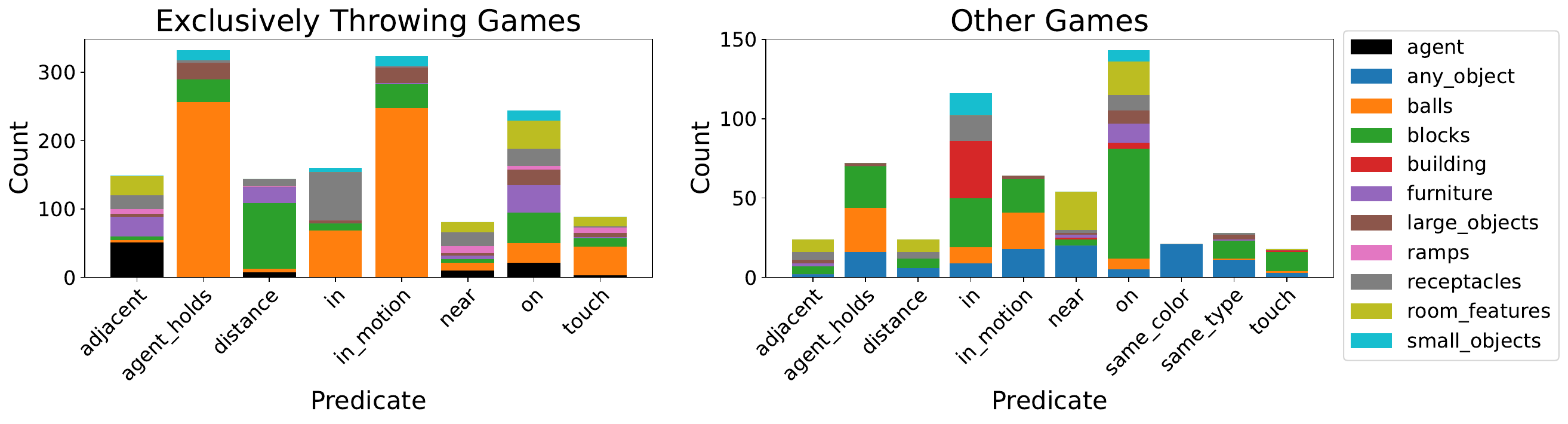}
    \caption{\textbf{Common-sense behavioral analyses .} 
    We plot similar information to \autoref{fig:behavioral-composite}b, but including additional object categories and predicates.
    }
    \label{fig:behavioral-common-sense}
\end{figure}

\begin{figure}[!btp]
    \centering
    \includegraphics[width= \textwidth]{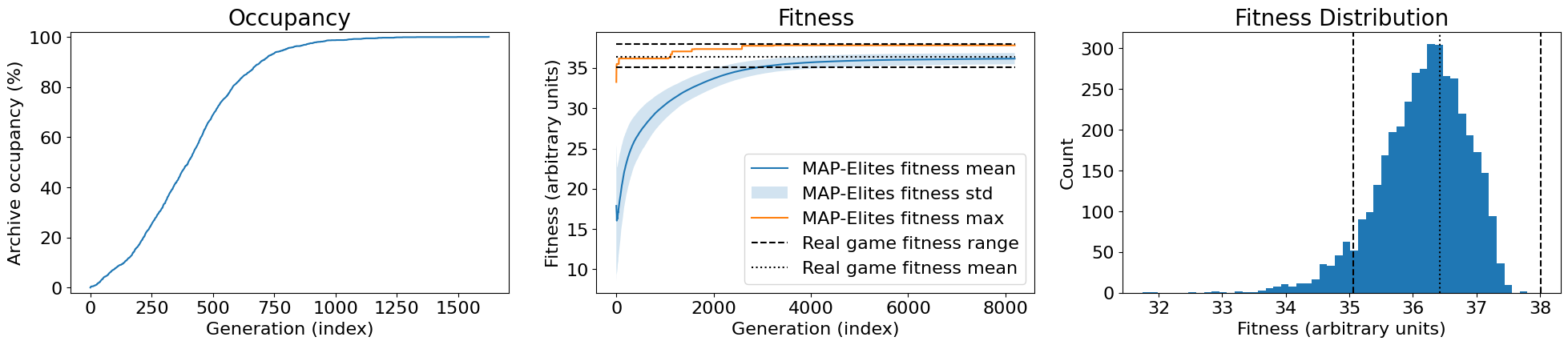}
    \caption{
        \textbf{Our implementation of the Goal Program Generator model fills the archive quickly and finds examples with human-like fitness scores.}
        \textbf{Left:} Our model rapidly finds exemplars for all archive cells (i.e. niches induced by our behavioral characteristics), reaching 50\% occupancy after 400 generations (out of a total of 8192)  and 95\% occupancy after 794 generations---the archive is almost full 1/10th of the way through the search process. 
        \textbf{Middle:} Our model reaches human-like fitness scores. After only three generations, the fittest sample in the archive has a higher fitness score than at least one participant-created game. By the end of the search, the mean fitness in the archive is close to the mean fitness of human games.
        \textbf{Right:} Our model generates the vast majority of its samples within the range of fitness scores occupied by participant-created games, though few samples approach the top of the range.
    }
    \label{fig:results-quantitative}
\end{figure}

\begin{figure}[!bthp]
    \centering
    \includegraphics[width=\columnwidth]{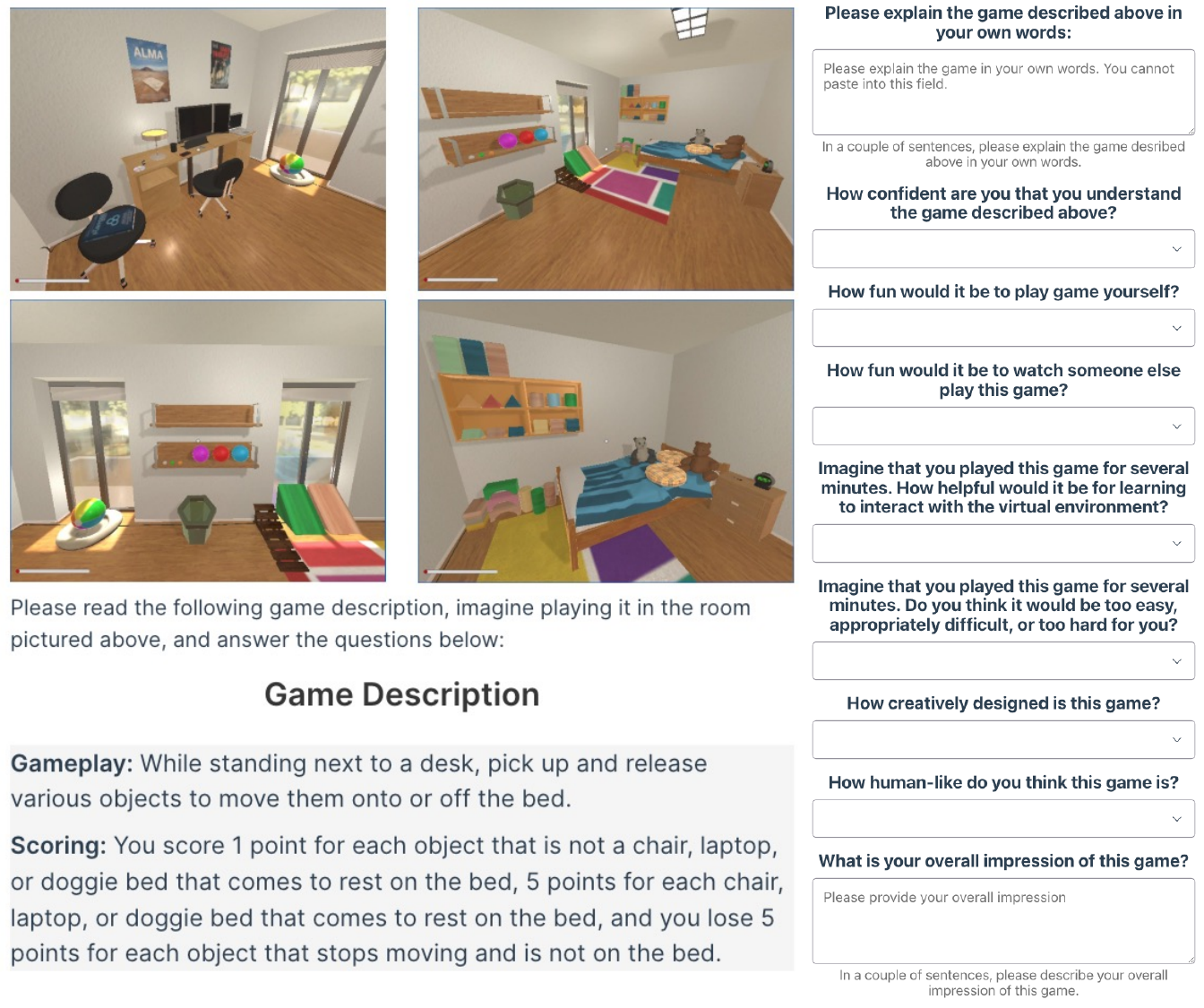}
    \caption{\textbf{Human evaluations interface.} 
    For each game, participants viewed the same four images of the environment, followed by the GPT-4 back-translated description of the game (see \nameref{methods:human-evals} for details.
    They then answered the two free-response and seven multiple-choice questions on the right. 
    In the web-page based version, the questions appeared below the game description; they are presented side-by-side to save space. 
    } 
    \label{fig:methods-human-evals-interface}
\end{figure}

\begin{table}[htb]
\centering
\footnotesize
\begin{adjustbox}{center}
\begin{threeparttable}
\caption{ \textbf{Human evaluation result summary}  }
\label{tab:mann-whitney}
\begin{tabular}{lcccccc}
\toprule

& \multicolumn{3}{c}{\textbf{Mean score by category}} & \multicolumn{3}{c}{\textbf{Significance of difference}} \\

\textbf{Measure} & \texttt{Real (R)} & \texttt{Matched (M)} & \texttt{Unmatched (U)} & \texttt{R} vs \texttt{M} & \texttt{R} vs \texttt{U} & \texttt{M} vs \texttt{U} \\

\midrule

\textit{Understandable} $\uparrow$ & 3.943 & 3.923 & 3.331 & - & *** & *** \\
\textit{Fun to play} $\uparrow$           & 2.522 & 2.430 & 2.068 & - & *** & *** \\
\textit{Fun to watch} $\uparrow$          & 2.385 & 2.313 & 2.024 & - & *** & **  \\
\textit{Helpful}\textsuperscript{\textdagger} $\uparrow$ & 2.997 & 2.987 & 2.840 & - &  -  &  -  \\
\textit{Difficult} ${\downarrow \atop \uparrow }$               & 2.582 & 2.660 & 2.676 & - & - & - \\
\textit{Creative} $\uparrow$               & 2.318 & 2.213 & 2.143 & - & * & - \\
\textit{Human-like} $\uparrow$            & 2.813 & 2.670 & 2.119 & - & *** & *** \\

\bottomrule
\end{tabular}

\begin{tablenotes}
    \item \textbf{Non-parametric significance test results mostly corroborate mixed-model results}.
    Participants responded to seven Likert questions on a 5-point scale, one for each attribute in the first column (see \nameref{methods:human-evals}). 
    We report the two-sided nonparametric Mann-Whitney $U$ test \cite{MannWhitney1947} for differences in outcomes. 
    We find that under this test, (1) evaluators don't distinguish between participant-created \texttt{real} and \texttt{matched} model games, but (2) do distinguish \texttt{unmatched} games from both.
    The first effect is consistent with the summary of our mixed models based on the method of marginal means (\autoref{tab:marginal-means-summary}), and the second effect is in a similar direction, with more statistically significant efects, to the one found above.
    See Supplementary \autoref{tab:supplementary-mann-whitney} for test statistics and P-values. 
    *: $P < 0.05$, **: $P < 0.01$, ***: $P< 0.001$. \\
    \textdagger: The full measure description is ``Helpful for interacting with the simulated environment.'' \\
    In most measures, higher scores are better, indicated by the $\uparrow$, other than \textit{Difficult}  ${\downarrow \atop \uparrow }$, in which 3 means ``appropriately difficult'', and scores below and above indicate too easy and too hard respectively.
\end{tablenotes}
\end{threeparttable}
\end{adjustbox}
\end{table}

\begin{table}[!htbp]
\centering
\footnotesize
\begin{adjustbox}{center}
\begin{threeparttable}
\caption{\textbf{Mixed model result summary}  }
\label{tab:mixed-models}

\begin{tabular}{lcccccccc}


\toprule
& \multicolumn{8}{c}{\textbf{Variable}} \\
& \multicolumn{2}{c}{\texttt{Fitness}} & & \multicolumn{2}{c}{$\mathbbm{1}[\texttt{Matched}]$} & &  \multicolumn{2}{c}{$\mathbbm{1}[\texttt{Real}]$} \\
\textbf{Measure} & $\beta_{\texttt{fit}}$ & Significance & & $\beta_{\texttt{matched}}$ & Significance & & $\beta_{\texttt{real}}$ & Significance \\


\midrule

\textit{Understandable} $\uparrow$ & 0.846  & *** & & 0.525 & -  & & 1.151  & *** \\
\textit{Fun to play} $\uparrow$   & 0.396  & **  & & 0.629 & *  & & 1.059  & *** \\
\textit{Fun to watch} $\uparrow$  & 0.191  & -   & & 0.641 & *  & & 0.912  & *** \\
\textit{Helpful}\textsuperscript{\textdagger} $\uparrow$       & -0.189 & *   & & 0.349 & *  & & 0.232  & -   \\
\textit{Difficult}  ${\downarrow \atop \uparrow }$     & -0.588 & *** & & 0.363 & -  & & -0.250 & -   \\
\textit{Creative} $\uparrow$      & -0.486 & **  & & 0.551 & -  & & 0.438  & -   \\
\textit{Human-like} $\uparrow$    & 0.570  & *** & & 0.837 & ** & & 1.446  & *** \\



\bottomrule
\end{tabular}
\begin{tablenotes}
    \item \textbf{Fitness scores significantly predict several attributes, including understandability and human-likeness.}
    Fitness scores show (statistically) significant positive effects on the understandability, fun to play, and human-likeness attributes, and significant negative effects on the helpfulness, difficulty and creativity questions. Accounting for the role of fitness, the \texttt{matched} group membership shows significant effects only the fun to play and watch, helpfulness, and human likeness questions. 
    The \texttt{real} group shows significant effects on understandability, fun to play and watch, and human likeness. 
    We report coefficient significance estimates using the two-sided Wald test.
    See Supplementary \autoref{tab:supplementary-mixed-models} for test statistics and P-values. 
    *: $P < 0.05$, **: $P < 0.01$, ***: $P< 0.001$ \\
    \textdagger: The full measure description is ``Helpful for interacting with the simulated environment.'' \\
    In most measures, higher scores are better, indicated by the $\uparrow$, other than \textit{Difficult}  ${\downarrow \atop \uparrow }$, in which 3 means ``appropriately difficult'', and scores below and above indicate too easy and too hard respectively.
\end{tablenotes}
\end{threeparttable}
\end{adjustbox}
\end{table}

\begin{figure}[!hbtp]
    \centering
    \includegraphics[width=\columnwidth]{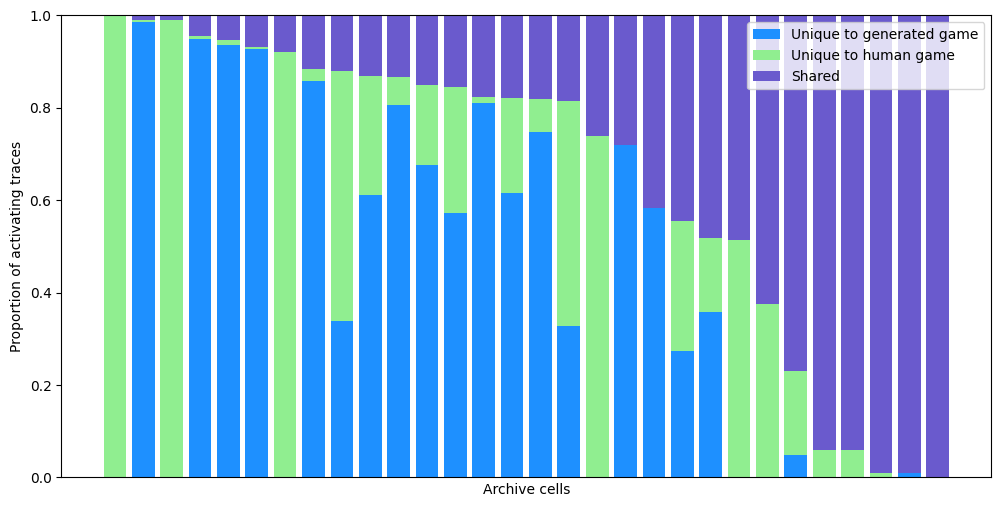}
    \caption{\textbf{Proportion of human interactions activating only \texttt{matched} and \texttt{real} games in the same cell.}
    Each bar corresponds to a pair of corresponding \texttt{matched} and \texttt{real} games. 
    In each bar, we plot the proportion of relevant interactions (state-action traces) that are unique to the \texttt{matched} game (blue), unique to the \texttt{real} game (green), or shared across both (purple). 
    A few games (with the bar mostly or entirely in purple) show high similarity between the corresponding games --- under 25\% (7/30) share more than half of their relevant interactions.
    Most games, however, show substantial differences between the sets of relevant interactions, with some showing a higher fraction unique to human games and others to matched model games.
    The average Jaccard similarity between the sets of relevant interactions for the \texttt{matched} and \texttt{real} game is only 0.347 and the median similarity is 0.180 (identical games would score 1.0, entirely dissimilar games 0).
    } 
    \label{fig:real-matched-comparison}
\end{figure}

\clearpage
\appendix
\setcounter{figure}{0}
\renewcommand\thefigure{SI-\arabic{figure}}
\setcounter{table}{0}
\renewcommand{\thetable}{SI-\arabic{table}}

\section{Pseudocode and program summary translation}

\rotatebox{90}{\begin{minipage}{0.95\textheight}
    \includegraphics[width=\columnwidth]{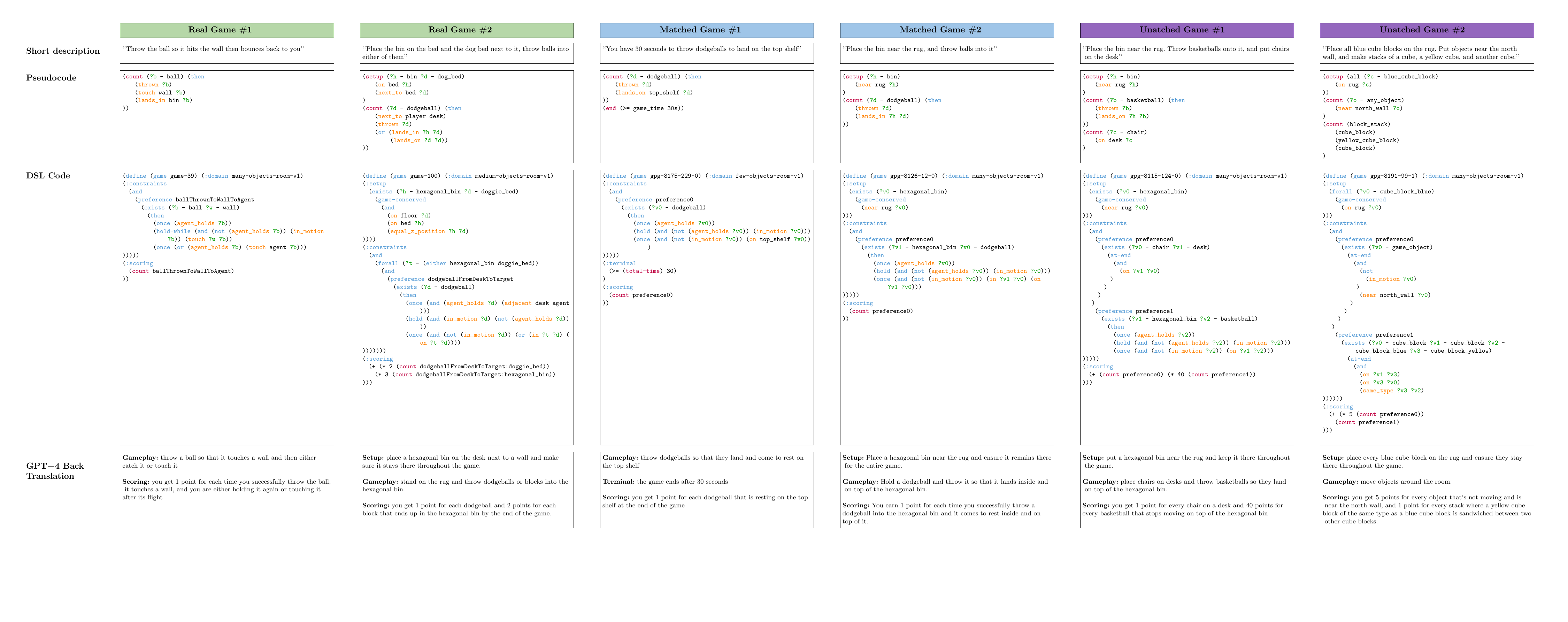}
    \captionof{figure}{\textbf{Translation of program pseudocode to our DSL.}
    Each column depicts a single goal, starting from its summary description in natural language, the pseudocode program we used to represent it, its program representation in our domain-specific language (see \cref{sec:appendix-dsl-ref}), and finally a GPT-4 automatic back-translation to natural language (see \cref{sec:appendix-backtranslation}).
    The first two columns(in \textcolor[RGB]{78,167,46}{green}) highlight real participant-created games in our experiment mentioned in both \autoref{fig:1} and \autoref{fig:model-overview}.
    The two middle columns (in \textcolor[RGB]{15,158,213}{blue}) describe model-generated goals ``matching'' the participant-created ones, and the last two columns (in \textcolor[RGB]{148,103,189}{purple}) outline two ``unmatched'' novel model-generated goals.
    } 
    \label{fig:appendix-pseudocode-translation}
\end{minipage}}

\FloatBarrier
\clearpage

\section{Natural language to domain-specific language translation analyses}
\label{sec:appendix-nl-to-dsl}

We use our goal program interpreter to perform a measure of validation of our manual translation from participant-specified games in natural language to programs in our domain-specific language.
We attempt to answer the following questions:
\begin{enumerate}
    \item Do the translated programs denote semantics that are at all feasible to pursue?
    \item Is there evidence that the translated programs capture the \textit{specific} semantics that the participants attempted to convey?
    \item Are there any clear patterns in which games appear to match (or not match) participants' interactions in our environment?
\end{enumerate}
To do so, we leverage our dataset of participant interactions with the environment (see \textbf{Interaction traces} under \nameref{methods:dataset}) and our implementation of a goal program interpreter (see \textbf{Goal program interpreter} under \nameref{methods:dataset}). 
We ground each participant's interactions to all 98 participant-described games in our final dataset, and count, for each program section (setup conditions or gameplay preference), which participants' interactions satisfied it at least once. 
Because we asked each participant to attempt to accomplish their reported goal, we can empirically examine whether a participant’s actions (which are presumably goal-oriented) cause our translated programs to be satisfied. 
We offer three brief findings:
\begin{enumerate}
    \item \textit{Most goal components are satisfied by at least one participant:} The 98 goals in our dataset map onto programs with a total of 249 components. The vast majority of these (226/249) correspond to a sequence of actions taken by at least one participant. We manually identify that the remaining 23 components are either very challenging but achievable conditions (11/23) or rare conditions that we did not implement in our reward machine (12/23)

    \item \textit{Most participants satisfied the components of their own goals:} We find that the vast majority of participants (86/98) performed a sequence of actions that satisfied at least one component of their translated goal program, and most (56/98) interacted in the environment in a manner that satisfied every aspect of their program inferred from their goal.

    \item \textit{We find no clear patterns in which games are not fulfilled by their creators:} Games of various types (throwing, building, organizing) are fully satisfied, as are games of varying program complexities (measured by the number of program sections). We offer this as minimal evidence that there is no systematic failure to represent the semantics of a particular game archetype.  
\end{enumerate}
Taken together, we view this as evidence that the program representations meaningfully capture some, if not all, of the core semantics of the goals our participants created. We can conceive of factors why these statistics might underestimate the viability of our goal programs
(e.g., the fact that we did not extrinsically incentivize participants to play their games, the possibility that participants may have tried and failed to accomplish their games, or limitations in the program interpreter)
or overestimates the viability (i.e. we capture some of the semantics, but we don’t have clear evidence of which we failed to capture).

\section{Full feature set}
\label{sec:appendix-features}
To simplify training fitness models, we ensure that all feature values are on the unit interval using the following feature types:
\begin{itemize}
    \item A binary value (marked with \texttt{[b]})
    \item A proportion between zero and one (\texttt{[p]})
    \item A real value discretized to two or more levels and treated as an indicator variable (\texttt{[d]}, with the levels listed at the end of the description)
    \item A float value normalized to the unit interval over the full dataset of positive and negative games (\texttt{[f]})
\end{itemize}
For our n-gram features, we extract n-gram tokens from an in-order traversal of the syntax tree. We use 5-gram models with stupid backoff \cite{Brants2007} with a discount factor of 0.4, and report the mean log score as the feature value, both jointly over the entire game program and separately over the different sections (setup, preferences, terminal conditions, and scoring).

For our predicate play trace features, we use a simplified version of the predicate satisfaction computation aspect of our reward machine (DSL program interpreter).
We record, for every human play trace we have, and each predicate listed below, for every object assignment that satisfies it in that trace, all indices of states at which the predicate is satisfied.
Recording specific states allows to us compute conjunctions, disjunctions, and negations in addition to individual predicate satisfactions. 
We limit ourselves to a subset of our predicates, which covers over 95\% of predicate references in our dataset: 
\texttt{above}, \texttt{adjacent}, \texttt{agent\_crouches}, \texttt{agent\_holds}, \texttt{broken}, \texttt{game\_start}, \texttt{game\_over},\texttt{in}, \texttt{in\_motion}, \texttt{object\_orientation}, \texttt{on}, \texttt{open}, \texttt{toggled\_on}, and \texttt{touch}.
Any predicate that is not implemented is assumed to be feasible to have been satisfied. 

Our full feature set is:

{\small
\textbf{ngram}: Features using our n-gran model. 
\begin{easylist}
@ \texttt{ast\_ngram\_full\_n\_5\_score [f]}: What is the mean 5-gram model score under an n-gram model trained on the real games?
@ \texttt{ast\_ngram\_setup\_n\_5\_score [f]}: What is the mean 5-gram model score of the setup section under an n-gram model trained on the real game setup sections?
@ \texttt{ast\_ngram\_constraints\_n\_5\_score [f]}: What is the mean 5-gram model score of the gameplay preferences section under an n-gram model trained on the real game preferences sections?
@ \texttt{ast\_ngram\_terminal\_n\_5\_score [f]}: What is the mean 5-gram model score of the terminal conditions section under an n-gram model trained on the real game terminal sections?
@ \texttt{ast\_ngram\_scoring\_n\_5\_score [f]}: What is the mean 5-gram model score of the scoring section under an n-gram model trained on the real game scoring sections?
\end{easylist}

\textbf{play\_trace\_database}: Features using our play trace database.
\begin{easylist}
@ \texttt{predicate\_found\_in\_data\_prop [p]}: What proportion of predicates are satisfied at least once in our human play trace data?
@ \texttt{predicate\_found\_in\_data\_small\_logicals\_prop [p]}: What proportion of logical expressions over predicates (with four or fewer children, limited for computational reasons) are satisfied at least once in our human play trace data?
\end{easylist}

\textbf{defined\_and\_used}: Features reflecting whether particular game components are defined, and features capturing whether defined components (such as variables, gameplay preferences, or objects in the setup) are then also used elsewhere.
\begin{easylist}
@ \texttt{variables\_used\_all [b]}: Are all variables defined used at least once?
@ \texttt{variables\_used\_prop [p]}: What proportion of variables defined are used at least once?
@ \texttt{preferences\_used\_all [b]}: Are all preferences defined referenced at least once in terminal or scoring expressions?
@ \texttt{preferences\_used\_prop [p]}: What proportion of preferences defined are referenced at least once in terminal or scoring expressions?
@ \texttt{setup\_quantified\_objects\_used [p]}: What proportion of object or types quantified as variables in the setup are also referenced in the gameplay preferences?
@ \texttt{any\_setup\_objects\_used [b]}: Are any objects referenced in the setup also referenced in the gameplay preferences?
@ \texttt{section\_doesnt\_exist\_setup [b]}: Does a game not have an (optional) setup section? (to allow counteracting feature values for the setup for games that do not have a setup component)
@ \texttt{section\_doesnt\_exist\_terminal [b]}: Does a game not have an (optional) terminal conditions section? (to allow counteracting feature values for the terminal conditions for games that do not have a terminal conditions component)
\end{easylist}

\textbf{grammar\_misuse}: Features capturing various modes of grammar misuse---expressions that are grammatical under the DSL but ill-formed, poorly structured, or whose values cannot vary over gameplay. 
\begin{easylist}
@ \texttt{adjacent\_once\_found [b]}: Are there any cases where the \texttt{once} modal, which captures a single state, is used twice in a row?
@ \texttt{adjacent\_same\_modal\_found [b]}: Are there any cases where the same modal is used twice in a row?
@ \texttt{once\_in\_middle\_of\_pref\_found [b]}: Are there any cases where the \texttt{once} modal, which captures a single state, is in the middle of a sequence of modals?
@ \texttt{pref\_without\_hold\_found [b]}: Are there any cases where a sequence of modals is specified with no temporally extended modal (\texttt{hold} or \texttt{hold-while})?
@ \texttt{identical\_consecutive\_seq\_func\_predicates\_found [b]}: Are there any cases where the same exact predicates (and their arguments) are applied in consecutive modals (making them redundant)?
@ \texttt{predicate\_without\_variables\_or\_agent [b]}: Are there any predicates that do not reference any variables or the agent?
@ \texttt{nested\_logicals\_found [b]}: Are there any cases where a logical operator is nested inside the same logical operator (e.g., a negation of a negation, or a conjunction of a conjunction)?
@ \texttt{identical\_logical\_children\_found [b]}: Are there any cases where a logical operator has two or more identical children?
@ \texttt{redundant\_expression\_found [b]}: Are there any cases where a logical expression over predicates is redundant (can be trivially simplified)?
@ \texttt{unnecessary\_expression\_found [b]}: Are there any cases where a logical expression over predicates is unnecessary (contradicts itself, or is trivially true)?
@ \texttt{repeated\_variables\_found [b]}: Are there any cases where the same variable is used twice in the same predicate?
@ \texttt{repeated\_variable\_type\_in\_either [b]}: Are there any cases where the same variable types is used twice in an \texttt{either} quantification?
\end{easylist}

\textbf{scoring\_grammar\_misuse}: Features capturing similar failure modes to the above category, but localized to the scoring and terminal sections of the DSL.
\begin{easylist}
@ \texttt{identical\_scoring\_children\_found [b]}: Are there any cases where a scoring arithmetic or logical expression has two or more identical children?
@ \texttt{redundant\_scoring\_terminal\_expression\_found [b]}: Are there any cases where a scoring or terminal expression is redundant (can be trivially simplified)?
@ \texttt{unnecessary\_scoring\_terminal\_expression\_found [b]}: Are there any cases where a scoring or terminal expression is unnecessary (contradicts itself, or is trivially true)?
@ \texttt{total\_score\_non\_positive [b]}: Do the scoring rules of the game result in a non-positive score regardless of gameplay?
@ \texttt{scoring\_preferences\_used\_identically [b]}: Do the scoring rules of the game treat all gameplay preferences identically?
@ \texttt{two\_number\_operation\_found [b]}: Are there any cases where an arithmetic operation is applied to two numbers? (e.g. \texttt{(+ 5 5)} instead of simplifying it)
\end{easylist}

\textbf{game\_element\_disjointness}: Features capturing whether particular game elements are disjoint---for example, gameplay preferences using disjoint sets of objects, or temporal modals using disjoint sets of variables.
\begin{easylist}
@ \texttt{disjoint\_preferences\_found [b]}: Are there any preferences that quantify over disjoint sets of objects?
@ \texttt{disjoint\_preferences\_scoring\_terminal\_types [p]}: Do the preferences referenced in the scoring and terminal section quantify over disjoint sets of object types?
@ \texttt{disjoint\_preferences\_scoring\_terminal\_predicates [p]}: Do the preferences referenced in the scoring and terminal section use disjoint sets of predicates?
@ \texttt{disjoint\_seq\_funcs\_found [b]}: Are there any cases where modals in a preference refer to disjoint sets of variables or objects?
@ \texttt{disjoint\_at\_end\_found [b]}: Are there any cases where predicate expressions under an \texttt{at\_end} refer to disjoint sets of variables or objects?
@ \texttt{disjoint\_modal\_predicates\_found [b]}: Are there any cases where modals in a preference refer to disjoint sets of predicates?
@ \texttt{disjoint\_modal\_predicates\_prop [p]}: What proportion of modals in a preference refer to disjoint sets of predicates?
\end{easylist}

\textbf{counting}: Features tracking node count or maximal depth in the four different DSL program sections.
\begin{easylist}
@ \texttt{node\_count\_section [d]}: How many nodes are in the \textit{section}, discretized to five bins with different thresholds for each section.
@ \texttt{max\_depth\_section [d]}: What is the maximal depth of the syntax tree in the \textit{section}, discretized to five bins with different thresholds for each section.
\end{easylist}

\textbf{pref\_forall}: Features capturing whether or not and how well the games use the \texttt{forall} over preferences syntax.
\begin{easylist}
@ \texttt{pref\_forall\_used\_correct [b]}: For the \texttt{forall} over preferences syntax, if it is used, is it used correctly in this game?
@ \texttt{pref\_forall\_used\_incorrect [b]}: For the \texttt{forall} over preferences syntax, if it is used, is it used incorrectly in this game? (to allow learning differential values between correct and incorrect usage)
@ \texttt{pref\_forall\_external\_forall\_used\_correct [b]}: If the \texttt{count-once-per-external-objects} count operator is used, is it used correctly in this game?
@ \texttt{pref\_forall\_external\_forall\_used\_incorrect [b]}: If the \texttt{count-once-per-external-objects} count operator is used, is it used incorrectly in this game?
@ \texttt{pref\_forall\_pref\_forall\_correct\_arity\_correct [b]}: If optional object names and types are provided to a count operation, are they provided with an arity consistent with the \texttt{forall} variable quantifications?
@ \texttt{pref\_forall\_pref\_forall\_correct\_arity\_incorrect [b]}: If optional object names and types are provided to a count operation, are they provided with an arity inconsistent with the \texttt{forall} variable quantifications?
@ \texttt{pref\_forall\_pref\_forall\_correct\_types\_correct [b]}: If optional object names and types are provided to a count operation, are they provided with types consistent with the \texttt{forall} variable quantifications?
@ \texttt{pref\_forall\_pref\_forall\_correct\_types\_incorrect [b]}: If optional object names and types are provided to a count operation, are they provided with types inconsistent with the \texttt{forall} variable quantifications?
\end{easylist}
}

\subsection{Features Most Predictive of Real or Regrown Games}
\label{sec:appendix-predictive-features}
The following features (in order) had the largest weight, indicating they were most predictive of positive (real, human-generated) examples in our dataset. The last three features all capture the same concept, whether or not a setup section exists. We surmise the diffused weights over them are a result of using weight decay (an L2 penalty) on the model weights: 
{\small
\begin{easylist}[enumerate]
    @ \texttt{ast\_ngram\_full\_n\_5\_score}
    @ \texttt{ast\_ngram\_constraints\_n\_5\_score}
    @ \texttt{predicate\_found\_in\_data\_prop}
    @ \texttt{ast\_ngram\_setup\_n\_5\_score}	
    @ \texttt{variables\_used\_all}
    @ \texttt{preferences\_used\_all}
    @ \texttt{ast\_ngram\_scoring\_n\_5\_score}
    @ \texttt{max\_depth\_setup\_0} (which indicates a setup section does not exist or is very minimal)
    @ \texttt{node\_count\_setup\_0} (which indicates a setup section does not exist or is very minimal)
    @ \texttt{section\_doesnt\_exist\_setup}
\end{easylist}
}
The following features (in order) had the smallest weights, indicating they were most predictive of negative (regrown) examples in our dataset:
{\small
\begin{easylist}[enumerate]
    @ \texttt{pref\_forall\_used\_incorrect}
    @ \texttt{pref\_forall\_pref\_forall\_correct\_types\_incorrect}
    @ \texttt{disjoint\_seq\_funcs\_found}
    @ \texttt{repeated\_variables\_found}
    @ \texttt{redundant\_expression\_found}
    @ \texttt{pref\_forall\_pref\_forall\_correct\_arity\_incorrect}
    @ \texttt{predicate\_without\_variables\_or\_agent}
    @ \texttt{two\_number\_operation\_found}
    @ \texttt{nested\_logicals\_found}
    @ \texttt{redundant\_scoring\_terminal\_expression\_found}
\end{easylist}
}

\section{Objective function algorithm descriptions}
\label{sec:appendix-objective-function-algorimth}
\cref{alg:fitness-model} below outlines how we train our fitness model. 
The number $N$ of of positive examples is fixed (98 in our full dataset), and fewer during cross-validation.
We generate $M=1024$ negatives for each of the positive examples, and the number of features $F$ is fixed as well.
We perform cross-validation to select hyperparameter values $B \in \{1, 2, 4\}$, and $K \in \{256, 512, 1025\}$, selecting the set that minimizes the cross-validated loss.
We optimize the model with SGD, with a learning rate $\eta \in \{1e-3, 4e-3\}$ also selected via cross-validation.
We use weight decay with $\lambda = 0.003$ to regularize the model.
We train the model for up to 25000 epochs, or until the model plateaus for $P=500$ epochs. 
After cross-validation, we train our final objective function on the entire dataset. 
The final model we report uses $B=1$ positive games per batch, $K=1024$ negatives samples from our entire dataset for that positive, a learning rate $\eta = 4e-3$, and $F = 50$ features.

\label{sec:appendix-algorithms}
\begin{algorithm}
\caption{Fitness model training loop}
\label{alg:fitness-model}
\begin{algorithmic}
\Require Real games $\mathcal{D}^+ \in \mathbb{R}^{N \times \times 1 F}$, regrown games $\mathcal{D}^- \in \mathbb{R}^{N \times M \times F}$
\Require Fitness model $f_\theta: \mathbb{R}^F \to \mathbb{R}$, optimizer \\
$N$ positive examples, $M$ negatives generated per positive, $B$ batch size, $F$ features, $K$ negatives sampled per positive in each epoch, $P$ plateau epochs 
\Ensure Converged fitness model $W_\theta$
\State best model $\gets$ None
\State best loss $\gets \infty$ 
\State last improvement epoch $\gets -1$
\For{epoch $i$}
    \LineComment{Assign negatives randomly to each positive}
    \State Shuffle the first two dimensions of $\mathcal{D}^-$ 
    \LineComment{Reorder the positives in each epoch}
    \State Shuffle the first dimension of $\mathcal{D}^+$ 
    \For{each batch}
        \State $X^+ \gets$ the next $B$ positives \Comment $X^+$: $B \times 1 \times F$
        \State $X^- \gets$ $K$ sampled negatives for each positive \Comment $X^-$: $B \times K \times F$
        \State $X \gets \text{concat}(X^+, X^-)$  \Comment $X$: $B \times (1 + K) \times F$
        \State $Y \gets f_\theta(X)$ \Comment $Y$: $B \times (1 + K)$
        \State $L \gets \text{softmax loss}(Y)$   \Comment $L$: scalar
        \State Take backward step on loss and optimizer step
    \EndFor

    \State epoch validation losses $\gets []$
    \For{each batch in validation}
        \State \textless the above procedure without the optimizer steps\textgreater
        \State \textless append each batch's loss to epoch validation losses\textgreater
    \EndFor
    \State epoch loss $\gets$ mean(epoch validation losses)
    \If{epoch loss $<$ best loss} 
        \State best model $\gets$ copy of $f_\theta(X)$
        \State best loss $\gets$ epoch loss
        \State last improvement epoch $\gets$ i
    \ElsIf{$i \ -$ last improvement epoch $> P$ }
        \State break
    \EndIf
\EndFor
\State return best model
\end{algorithmic}
\end{algorithm}


\FloatBarrier
\section{MAP-Elites algorithm details}
\label{sec:appendix-map-elites}

\begin{table}[!htbp]
\caption{Exemplar preferences used as MAP-Elites behavioral characteristics.}
\label{tab:exemplar-preferences}
\begin{adjustbox}{center}
\begin{tabular}{|p{\gamecolumnwidth}|p{\gamecolumnwidth}|p{\gamecolumnwidth}|}
\toprule
\textbf{Exemplar Preference} & \textbf{Description (GPT-4 back-translated)} & \textbf{Exemplar Features } \\
\midrule
\begin{lstlisting}[aboveskip=-0.4 \baselineskip,belowskip=-0.8 \baselineskip]
(preference throwAttempt
  (exists (?b - dodgeball)
    (then
      (once (agent_holds ?b))
      (hold (and (not (agent_holds ?b)) (in_motion ?b)))
      (once (not (in_motion ?b)))
))) \end{lstlisting} & { \tiny This preference is satisfied when:

-first, the agent holds a dodgeball

-next, the agent throws the dodgeball

-finally, the dodgeball stops moving  } & { \tiny Uses predicate \texttt{agent\_holds} or \texttt{in\_motion}

Uses object category \texttt{balls} } \\
\begin{lstlisting}[aboveskip=-0.4 \baselineskip,belowskip=-0.8 \baselineskip]
(preference throwInBin
  (exists (?b - ball ?h - hexagonal_bin)
    (then
      (once (and (on rug agent) (agent_holds ?b)))
      (hold (and (not (agent_holds ?b)) (in_motion ?b)))
      (once (and (not (in_motion ?b)) (in ?h ?b)))
))) \end{lstlisting} & { \tiny This preference is satisfied when:

-first, the agent is standing on the rug and holding a ball

-next, the agent throws the ball

-finally, the ball stops moving and is inside a hexagonal bin  } & { \tiny Uses predicate \texttt{agent\_holds} or \texttt{in\_motion}

Uses predicate \texttt{in}

Uses predicate \texttt{on}

Uses object category \texttt{balls}

Uses object category \texttt{receptacles}

Uses object category \texttt{furniture} or \texttt{room\_features} } \\
\begin{lstlisting}[aboveskip=-0.4 \baselineskip,belowskip=-0.8 \baselineskip]
(preference ballThrownToBed
  (exists (?d - dodgeball)
    (then
      (once (and (agent_holds ?d) (adjacent desk agent)))
      (hold (and (not (agent_holds ?d)) (in_motion ?d)))
      (once (and (not (in_motion ?d)) (on bed ?d)))
))) \end{lstlisting} & { \tiny This preference is satisfied when:

-first, the agent holds a dodgeball while standing next to a desk

-next, the agent throws the dodgeball

-finally, the dodgeball stops moving and is on the bed  } & { \tiny Uses predicate \texttt{agent\_holds} or \texttt{in\_motion}

Uses predicate \texttt{on}

Uses predicate \texttt{adjacent} or \texttt{near} or \texttt{touch}

Uses object category \texttt{balls}

Uses object category \texttt{furniture} or \texttt{room\_features} } \\
\begin{lstlisting}[aboveskip=-0.4 \baselineskip,belowskip=-0.8 \baselineskip]
(preference itemInClosedDrawerAtEnd
  (exists (?g - game_object)
    (at-end
      (and
        (in top_drawer ?g)
        (not
          (open top_drawer)
))))) \end{lstlisting} & { \tiny This preference is satisfied when:

-at the end of the game, a game object is inside the top drawer and the top drawer is closed  } & { \tiny Uses predicate \texttt{in}

Uses object category \texttt{receptacles}

Uses object category \texttt{small\_objects} or \texttt{large\_objects} or \texttt{any\_object}

Uses \texttt{at\_end} } \\
\begin{lstlisting}[aboveskip=-0.4 \baselineskip,belowskip=-0.8 \baselineskip]
(preference watchOnShelf
  (exists (?w - watch ?s - shelf)
    (at-end
      (on ?s ?w)
))) \end{lstlisting} & { \tiny This preference is satisfied when:

-at the end of the game, a watch is on a shelf  } & { \tiny Uses predicate \texttt{on}

Uses object category \texttt{furniture} or \texttt{room\_features}

Uses object category \texttt{small\_objects} or \texttt{large\_objects} or \texttt{any\_object}

Uses \texttt{at\_end} } \\
\begin{lstlisting}[aboveskip=-0.4 \baselineskip,belowskip=-0.8 \baselineskip]
(preference gameBlockFound
  (exists (?l - block)
    (then
      (once (game_start))
      (hold (not (exists (?b - building) (and (in ?b ?l) (is_setup_object ?b)))))
      (once (agent_holds ?l))
))) \end{lstlisting} & { \tiny This preference is satisfied when:

-first, the game begins

-next, throughout the game, the block is not part of a building that is used in the setup

-finally, the agent picks up the block  } & { \tiny Uses predicate \texttt{agent\_holds} or \texttt{in\_motion}

Uses predicate \texttt{in}

Uses object category \texttt{blocks} or \texttt{building} } \\
\begin{lstlisting}[aboveskip=-0.4 \baselineskip,belowskip=-0.8 \baselineskip]
(preference matchingBuildingBuilt
  (exists (?b1 ?b2 - building)
    (at-end (and
        (is_setup_object ?b1)
        (not (is_setup_object ?b2))
        (forall (?l1 ?l2 - block) (or
            (not (in ?b1 ?l1))
            (not (in ?b1 ?l2))
            (not (on ?l1 ?l2))
            (exists (?l3 ?l4 - block) (and
                (in ?b2 ?l3)
                (in ?b2 ?l4)
                (on ?l3 ?l4)
                (same_type ?l1 ?l3)
                (same_type ?l2 ?l4)
)))))))) \end{lstlisting} & { \tiny This preference is satisfied when:

-at the end of the game, one building is part of the setup while the other is not

-and for any two blocks, neither is inside the building that is part of the setup

-if one block is not on top of the other, then there must be two other blocks of the same types inside the building that is not part of the setup, with one of these blocks on top of the other  } & { \tiny Uses predicate \texttt{in}

Uses predicate \texttt{on}

Uses object category \texttt{blocks} or \texttt{building}

Uses \texttt{at\_end} } \\
\begin{lstlisting}[aboveskip=-0.4 \baselineskip,belowskip=-0.8 \baselineskip]
(preference ballDroppedInBin
  (exists (?b - ball ?h - hexagonal_bin)
    (then
      (once (and (adjacent ?h agent) (agent_holds ?b)))
      (hold (and (in_motion ?b) (not (agent_holds ?b))))
      (once (and (not (in_motion ?b)) (in ?h ?b)))
))) \end{lstlisting} & { \tiny This preference is satisfied when:

-first, the agent is next to a hexagonal bin and is holding a ball

-next, the agent throws the ball

-finally, the ball stops moving and is inside the hexagonal bin  } & { \tiny Uses predicate \texttt{agent\_holds} or \texttt{in\_motion}

Uses predicate \texttt{in}

Uses predicate \texttt{adjacent} or \texttt{near} or \texttt{touch}

Uses object category \texttt{balls}

Uses object category \texttt{receptacles} } \\
\begin{lstlisting}[aboveskip=-0.4 \baselineskip,belowskip=-0.8 \baselineskip]
(preference pillowMovedToRoomCenter
  (exists (?o - pillow) (then
      (once (and (agent_holds ?o)))
      (hold (and (in_motion ?o) (not (agent_holds ?o))))
      (once (and (not (in_motion ?o)) (near room_center ?o) (exists (?o1 ?o2 ?o3 - game_object) (and (same_color ?o1 pink) (near room_center ?o1) (same_color ?o2 blue) (near room_center ?o2) (same_color ?o3 brown) (near room_center ?o3)))))
))) \end{lstlisting} & { \tiny This preference is satisfied when:

-first, the agent picks up a pillow

-next, the agent throws the pillow and it is no longer being held by the agent

-finally, the pillow stops moving near the center of the room, and there are three other objects near the center of the room as well: one that is pink, one that is blue, and one that is brown  } & { \tiny Uses predicate \texttt{agent\_holds} or \texttt{in\_motion}

Uses predicate \texttt{adjacent} or \texttt{near} or \texttt{touch}

Uses object category \texttt{furniture} or \texttt{room\_features}

Uses object category \texttt{small\_objects} or \texttt{large\_objects} or \texttt{any\_object} } \\
\bottomrule
\end{tabular}
\end{adjustbox}
\end{table}

We use a set of 9 exemplar preferences as the basis for our MAP-Elites behavioral characteristics, detailed in \cref{tab:exemplar-preferences}. 
To score each game with respect to each exemplar preference, we count how many of the game's preferences are a close match to the exemplar. 
We explored matching preferences by edit distance (in string or syntax tree space) but discovered the edit distance is rather easily game-able by the model, producing semantically similar preferences with high edit distance from each other. 
Instead, we represent each exemplar preference as a binary feature vector, with features for which groups of predicates the preference uses (4 features: \texttt{agent\_holds} or \texttt{in\_motion}, \texttt{in}, \texttt{on}, and \texttt{adjacent} or \texttt{near} or \texttt{touch}), and for which object categories the preference uses (5 features: \texttt{balls}, \texttt{receptacles}, \texttt{blocks} or \texttt{buildings}, \texttt{furniture} or \texttt{room\_features}, and \texttt{small\_items} or \texttt{large\_items} or the generic \texttt{game\_object}). 
Preferences in each archive game are also represented using this feature space.
A preference in an archive game is considered a match for an exemplar if it has an L1 distance of 0 or 1 in this space, and if a preference matches more than one exemplar, a match is randomly chosen. 
Exemplar preferences were iteratively chosen, starting from a seed preference (the first in \autoref{tab:exemplar-preferences}), and then greedily adding additional exemplars from the preferences defined in participant-created games.
At each step, the preference added was chosen to maximize the number of participant-created preferences that would be considered a match (distance of 0 or 1) from the exemplar set. 
In addition, we include two other behavioral characteristics, one capturing whether or not the game includes a setup component, and one capturing the total number of preferences (up to 4).
In total, this set of behavioral characteristics allows for an archive size of 2000 games, of which 20 have one preference (matching one of the 9 exemplars or matching none of them, with and without a setup component), 110 have two preferences, 440 have three preferences, and 1430 have four preferences. 

    
    
    
    
    
    
    
    


In addition, we add one more ``pseudo behavioral characteristic'' that explicitly captures a few general coherence properties of games -- specifically features that we expect either \textit{all} plausibly human-generated games to either exhibit or \textit{none} of them to exhibit. While these features are also used by our learned fitness function, we use this behavioral characteristic as a sort of first-stage filter: if a game fails to meet these criteria, then it cannot reasonably be said to be ``human-quality,'' regardless of its fitness evaluation. For all reported games, we ensure that each of the criteria are satisfied. The criteria included in this behavioral characteristic include whether all all variables are defined / used in preferences, whether all preferences are used in either terminal or scoring conditions, and whether the game avoids a set of grammatical but obviously nonsensical or redundant expressions. There are a total of 21 features used in this behavioral characteristic.  
This ``pseudo behavioral characteristic'' doubles the size of the archive (from 2000 games to 4000 games), though we never evaluate any game from the half of the archive in which this feature is false.

We begin the MAP-Elites algorithm by generating 1024 random games from the PCFG. We then sort each of the games in descending order of fitness and add them to the archive until either \textbf{(a)} every possible value of each behavioral characteristic is represented by at least one game (note that this is not the same as every possible \textit{combination} of behavioral characteristic values being represented), or \textbf{(b)} at least 128 cells of the archive are occupied.

We run MAP-Elites for 8192 ``generations,'' where each generation consists of 750 potential updates in which we randomly select a parent game, sample a mutation operator to apply, and potentially add the resulting mutated game to the archive.
We outline the mutation operators we use in the \nameref{methods:map-elites} section; we use a combination of standard genetic programming operations, such as mutation, insertion, crossover, and deletion, coupled with a few custom operators designed around our domain-specific language (whose role we ablate in \cref{fig:appendix-compositionality-ablations}). 
To help our model avoid redundant samples, we include a check that the gameplay preferences in each mutated child are all unique from each other before checking if this sample should be inserted into the MAP-Elites archive.
This check helps prevent situations where performing crossover introduces multiple copies of the same preference into the game (a failure mode we occasionally observed). 

\section{DSL to natural language back-translation}
\label{sec:appendix-backtranslation}

In order to prepare games for human evaluation, we convert them from the DSL to natural language in a multi-stage process. In order to ensure consistency, we perform this back translation on both generated games and the real games (as opposed to using the original human-authored descriptions).

In the first stage of back-translation, a rule based system converts the DSL into templated language by concretely describing the definition of each predicate and grammatical rule. For instance, the expression \texttt{(once (and (agent\_holds ?d) (adjacent ?p agent)))} is converted to ``there is a state where (the agent is holding \texttt{?d}) and (\texttt{?p} is adjacent to agent).'' Each of the game's setup conditions, preferences, terminal conditions, and scoring rules are rendered in this form, which also includes the mapping from variable names (e.g. \texttt{?d}) to the types of objects that can occupy the variable (e.g. ``dodgeball''). An example of a game's preferences described in this form is presented below:

\begin{lstlisting}[language=HTML]
The preferences of the game are:

-----Preference 1-----
The variables required by this preference are:
-?p of type pyramid_block
-?d of type dodgeball
-?h of type hexagonal_bin

This preference is satisfied when:
- first, there is a state where (the agent is holding ?d) and (?p is adjacent to agent)
- next, there is a sequence of one or more states where (it's not the case that the agent is holding ?d) and (?d is in motion)
- finally, there is a state where (it's not the case that ?d is in motion) and (?d is inside of ?h)

-----Preference 2-----
The variables required by this preference are:
-?b of type building
-?l of type cube_block
-?f of type flat_block

This preference is satisfied when:
- in the final game state, (?f is used in the setup), (?f is inside of ?b), and (?l is inside of ?b)
\end{lstlisting}

Next, we use the GPT-4 large language model (LLM) \cite{OpenAI2023GPT-4} to simplify the templated description into a more naturalistic form (specifically \texttt{gpt-4-1106-preview}). The objective of this stage is to re-write any unclear formulations generated by the initial procedure and to replace abstract variable names with their actual referents. We convert each section of the game separately, using a similar prompt for each. The prompt begins with the following message:

``\textit{Your task is to convert a templated description of a game's} \texttt{<setup / rules / terminal conditions / scoring conditions>} \textit{into a natural language description. Do not change the content of the template, but you may rewrite and reorder the information in any way you think is necessary in order for a human to understand it. Use simple language and verbs that would be familiar to a human who has never played this game before.}''

We then include 10 examples of this kind of translation taken from the set of human games not used in our experiments. An example of the same preferences in this simplified form is presented below:

\begin{lstlisting}[language=HTML]
The preferences of the game are:

-----Preference 1-----
This preference is satisfied when:
-first, the agent holds a dodgeball while standing next to a pyramid block
-next, the agent throws the dodgeball
-finally, the dodgeball lands inside a hexagonal bin and stops moving

-----Preference 2-----
This preference is satisfied when:
-at the end of the game, a flat block is used in the setup of a building and both a cube block and the flat block are inside the building
\end{lstlisting}

Finally, we use the LLM again to collect the separate descriptions of each section into one a single block, further simplifying the language and expressions. The prompt is similar to that used in the previous stage, and again is followed by 10 selected examples: 

``\textit{Your task is to combine and simplify the description of a game's rules. Do not change the content of the rules by either adding or removing information, but you may rewrite and reorder the information in any way you think is necessary in order for a human to understand it. Use simple language and verbs that would be familiar to a human who has never played this game before. DO describe preferences carefully, such that a player reading the description can easily play the game. DO NOT include explicit references to a game's preferences (i.e. "Preference 1" or "Preference 2"). DO NOT include descriptions of setup or terminal conditions if they do not appear in the game.}''

Examples of complete translations are available in \autoref{fig:model-comparison} and \autoref{fig:novel-texts}.

\FloatBarrier
\section{Model sample and real game edit distance similarity}
\label{sec:appendix-model-sample-edit-distance}

We analyze our model's results through the lens of the MAP-Elites behavioral characteristics we use, as they functionally define diversity for our model (see \nameref{methods:map-elites} and \cref{sec:appendix-map-elites} for additional details). 
When we present results from the model, such as in \autoref{fig:model-comparison}, we present \texttt{matched} model samples alongside the \texttt{real} participant-created games that MAP-Elites maps to the same archive cell. 
However, there are other ways to determine similarity in a high-dimensional space, such as the one our program representations occupy.
We wish to offer additional evidence for the degree of distinctiveness of the model-generated samples.
One reasonable approach to similarity is an edit distance: for simplicity, we use the string (Levenshtein) edit distance.
Each program's syntax tree is rendered as a string. We then remove the preamble that includes the game name and combine consecutive white space tokens to a single space. 
For each model-generated game included in the human evaluation set, we compute the edit distance to all 98 real participant-created games and record the smallest such distance.
Over the entire set of \texttt{matched} games included in the human evaluation set, the mean edit distance is 134.6 characters, with a standard deviation of 80.7; for the \texttt{unmatched} games, the mean is 197.1 with a standard deviation of 106.7.
To demonstrate the variation implied by different edit distances, we present the six model-generated games showcased in\autoref{fig:model-comparison} and \autoref{fig:novel-texts} alongside the participant-created game with the smallest edit distance to each of these samples. These comparisons are presented in\autoref{fig:appendix-edit-distance}.

In one case (Matched Game \#1), this is the same participant game that occupies the same archive cell. 
In the other two matched games, the nearest game is different. 
In both cases, the nearest participant-created game retains some high-level similarity, but with different gameplay objectives than the ones our model proposed. 
We also present the nearest matches for the closest \texttt{unmatched} model-created games in figure \autoref{fig:novel-texts}.
Here we find the nearest model games further away, both in edit distance and conceptually in the goals the programs represent. 
We take this as further evidence our model generates creative samples, meaningfully different from participant-created ones. 

\section{Highest fitness games}
\label{sec:appendix-highest-fitness}
In an attempt to visualize what our learned fitness function considers to be the fittest goals in our dataset, we plot the highest participant-created and model-generated goal programs in \autoref{fig:appendix-highest-fitness}.
We observe that our fitness function appears to place the highest fitness scores on simple throwing games, particularly from the human-created set. 
Our hypothesis is that this arises from two pressures.
The first is a bias toward simplicity: while we did not encode an explicit length-based prior, generating a longer program offers more opportunities to go awry, suggesting that all else being equal, it is harder to find flaws with shorter programs.
The second is a degree of mode-seeking. 
Many of the games in our dataset involved a throwing element of some sort, and several participants reported simple throwing games.
Our fitness function seems to have picked up on this and assigns these modal games the highest scores. 

\rotatebox{90}{\begin{minipage}{\textheight}
    \includegraphics[width=\columnwidth]{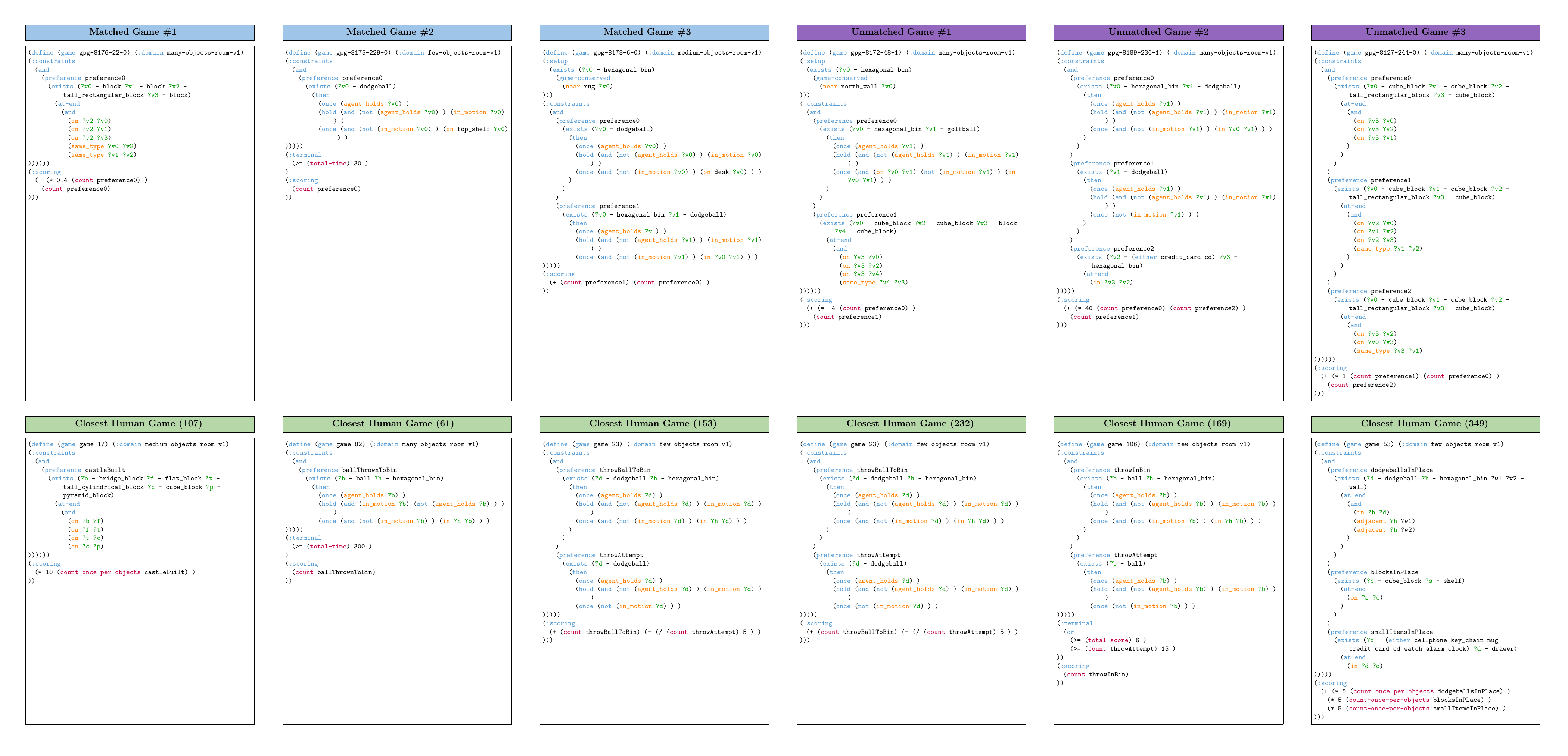}
    \captionof{figure}{\textbf{Edit distance nearest real programs to selected model samples.}
    For each model-generated sample presented in \autoref{fig:model-comparison} (first three columns with blue headers) and \autoref{fig:novel-texts} (last three columns with purple headers), we present below it the most similar real participant-created game, as measured by the Levenshtein edit distance.
    } 
    \label{fig:appendix-edit-distance}
\end{minipage}}

\rotatebox{90}{\begin{minipage}{\textheight}
    \includegraphics[width=\columnwidth]{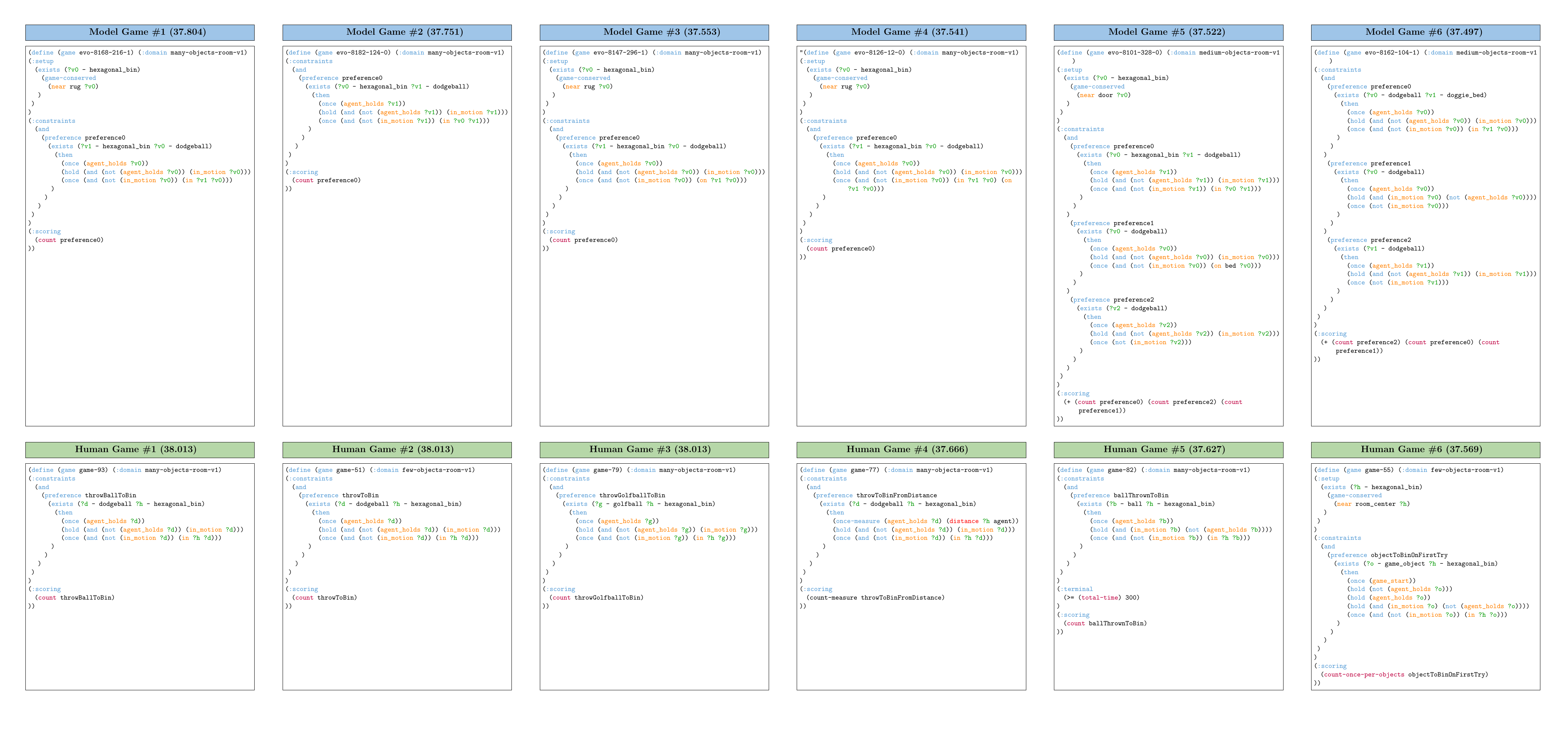}
    \captionof{figure}{\textbf{Highest fitness samples from our model and human participant dataset.}
    To visualize what our fitness model considers to be the fittest games, we print the programs of the six games with the highest fitness scores generated by our model and created by our human participants.
    The model seems to prefer simple ball-throwing games to other types of games, showing evidence of a bias toward simplicity and mode-seeking, as these are the most common types of games in our dataset. 
    } 
    \label{fig:appendix-highest-fitness}
\end{minipage}}

\FloatBarrier
\section{Human evaluations data analysis}
\label{sec:appendix-human-evals-analysis}

\subsection{Detailed human evaluation results}

We computed inter-rater reliability ratings using the Python \texttt{krippendorff} package \cite{Castro2017krippendorff}. 
We find the following results:
\begin{enumerate}
    \item \textit{Understandable}: $\alpha = 0.233$
    \item \textit{Fun to play}: $\alpha = 0.117$
    \item \textit{Fun to watch}: $\alpha = 0.077$
    \item \textit{Helpful}: $\alpha = 0.023$
    \item \textit{Difficult}: $\alpha = 0.211$
    \item \textit{Creative}: $\alpha = 0.129$
    \item \textit{Human-like}: $\alpha = 0.151$
\end{enumerate}
Given these results, while we offer results of the Mann-Whitney tests, we focus our discussion on the mixed model results.

\begin{table}[h]
\centering
\footnotesize
\begin{adjustbox}{center}
\begin{threeparttable}
\caption{ \textbf{Human evaluation result summary}  }
\label{tab:supplementary-mann-whitney}
\begin{tabular}{|p{2cm}|ccc|l|l|l|}
\toprule
 & \multicolumn{3}{|c|}{\textbf{Mean score by category}} & \textbf{\texttt{Real} vs. \texttt{Matched}} & \textbf{\texttt{Real} vs. \texttt{Unmatched}} &  \textbf{\texttt{Matched} vs. \texttt{Unmatched}} \\
  \textbf{Attribute} &   \texttt{Real} $\pm SE$ &   \texttt{Matched} $\pm SE$ &   \texttt{Unmatched} $\pm SE$ &    U-stat, P-value   &    U-stat, P-value  &    U-stat, P-value  \\
\hline
 \textit{Understandable}    & $3.943 \pm 0.068$ & $3.923 \pm 0.070$ & $3.331 \pm 0.075$ & 45088.0, $P = 0.906$ & 55921.5, $P = \num{ 1.718e-09 }^{ *** }$ & 55846.0, $P = \num{ 3.733e-09 }^{ *** }$ \\
 \textit{Fun to play}       & $2.522 \pm 0.066$ & $2.430 \pm 0.064$ & $2.068 \pm 0.062$ & 46752.5, $P = 0.352$ & 54040.5, $P = \num{ 3.235e-07 }^{ *** }$ & 52539.5, $P = \num{ 1.826e-05 }^{ *** }$ \\
 \textit{Fun to watch}   & $2.385 \pm 0.068$ & $2.313 \pm 0.066$ & $2.024 \pm 0.064$ & 46169.0, $P = 0.519$ & 51636.5, $P = \num{ 8.793e-05 }^{ *** }$ & 50515.0, $P = \num{ 1.027e-03 }^{ ** }$  \\
 \textit{Helpful}\textsuperscript{\textdagger}        & $2.997 \pm 0.068$ & $2.987 \pm 0.066$ & $2.840 \pm 0.073$ & 44802.0, $P = 0.982$ & 47372.5, $P = 0.078$                     & 47559.0, $P = 0.075$                     \\
 \textit{Difficult}       & $2.582 \pm 0.055$ & $2.660 \pm 0.056$ & $2.676 \pm 0.066$ & 42921.5, $P = 0.326$ & 42218.5, $P = 0.419$                     & 44081.0, $P = 0.947$                     \\
 \textit{Creative}                  & $2.318 \pm 0.061$ & $2.213 \pm 0.056$ & $2.143 \pm 0.064$ & 47036.0, $P = 0.282$ & 48286.0, $P = 0.025^{ * }$               & 46615.0, $P = 0.182$                     \\
 \textit{Human-like}               & $2.813 \pm 0.066$ & $2.670 \pm 0.070$ & $2.119 \pm 0.067$ & 47698.0, $P = 0.167$ & 58679.0, $P = \num{ 1.702e-13 }^{ *** }$ & 55434.5, $P = \num{ 1.333e-08 }^{ *** }$ \\
\bottomrule
\end{tabular}

\begin{tablenotes}
    \item \textbf{Evaluators don't distinguish between participant-created \texttt{real} and \texttt{matched} model games, but do distinguish \texttt{unmatched} games from both.} Participants responded to seven Likert questions on a 5-point scale, one for each attribute in the first column (see \nameref{methods:human-evals} for additional details). We first report the mean score for games in each category and the standard error of the mean. We then report the two-sided nonparametric Mann-Whitney $U$ test \cite{MannWhitney1947} for differences in outcomes, appropriate for ordinal data, between each pair of game categories. *: $P < 0.05$, **: $P < 0.01$, ***: $P< 0.001$ \\
    \textdagger: The full measure description is ``Helpful for interacting with the simulated environment.'' 
\end{tablenotes}
\end{threeparttable}
\end{adjustbox}
\end{table}

\FloatBarrier
\subsection{Fitness-less mixed models analysis}
\label{sec:appendix-no-fitness-mixed-models}
We begin our analysis focusing on the differences between the \texttt{real}, \texttt{matched}, and \texttt{unmatched} game groups.
To that end, we fit mixed-effect models that include a fixed effect for the game type (treating the \texttt{unmatched} group as our baseline), as well as random effects for the individual games and raters.
\autoref{tab:supplementary-no-fitness-mixed-models} summarizes our fitted model coefficients (akin to \autoref{tab:supplementary-mixed-models}), and \autoref{tab:supplementary-no-fitness-marginal-means} repeats the marginal means analysis (akin to \autoref{tab:supplementary-marginal-means}).
In this case, the effect of estimated marginal means is to balance out the effects of rater and individual game random effects, since there is no other fixed effect to marginalize over.
Notably, all comparisons between the \texttt{real} and \texttt{matched} groups are non-significant, and comparison between the \texttt{unmatched} and other groups are significant for the understandability, fun, and human-likeness questions.
We replicate most of these effects in the mixed effect regressions that include fitness, reported below, both in the significance of coefficients and the significance (or lack thereof) of comparisons between groups.

\begin{table}[h]
\centering
\footnotesize
\begin{adjustbox}{center}
\begin{threeparttable}
\caption{\textbf{Mixed model without fitness result summary}  }
\label{tab:supplementary-no-fitness-mixed-models}
\begin{tabular}{|p{2cm}|rrl|rrl|}
\toprule
&  \multicolumn{3}{c|}{\texttt{Matched}} & \multicolumn{3}{c|}{\texttt{Real}} \\
\textbf{Attribute} & $\beta_{\text{matched}} \pm SE$ & $Z$ & $P$-value & $\beta_{\text{real}} \pm SE$ & $Z$ & $P$-value \\ 
  \midrule
    \textit{Understandable} & $1.042 \pm 0.333$ & 3.133 & $P = \num{1.731e-03}^{ ** }$ & $1.042 \pm 0.332$ & 3.138 & $P = \num{1.699e-03}^{ ** }$ \\ 
  \textit{Fun to play} & $0.877 \pm 0.273$ & 3.210 & $P = \num{1.325e-03}^{ ** }$ & $1.020 \pm 0.274$ & 3.722 & $P = \num{1.976e-04}^{ *** }$ \\ 
   \textit{Fun to watch} & $0.757 \pm 0.257$ & 2.944 & $P = \num{3.235e-03}^{ ** }$ & $0.892 \pm 0.259$ & 3.446 & $P = \num{5.693e-04}^{ *** }$ \\ 
  \textit{Helpful}\textsuperscript{\textdagger} & $0.236 \pm 0.165$ & 1.426 & $P = 0.154$ & $0.251 \pm 0.165$ & 1.521 & $P = 0.128$ \\ 
  \textit{Difficult} & $0.006 \pm 0.361$ & 0.016 & $P = 0.987$ & $-0.194 \pm 0.361$ & -0.538 & $P = 0.591$ \\ 
   \textit{Creative} & $0.261 \pm 0.314$ & 0.832 & $P = 0.405$ & $0.489 \pm 0.316$ & 1.548 & $P = 0.122$ \\ 
   \textit{Human-like} & $1.197 \pm 0.283$ & 4.225 & $P = \num{2.388e-05}^{ *** }$ & $1.396 \pm 0.283$ & 4.927 & $P = \num{8.343e-07}^{ *** }$ \\ 
   \bottomrule
\end{tabular}
\begin{tablenotes}
    \item \textbf{Many effects of game group persist absent the fitness scores.}
    Compare this table to the full model results in \autoref{tab:supplementary-mixed-models}.
    We observe that many of the effects on membership in the \texttt{matched} or \texttt{real} are significant also without accounting for the role of fitness.
    Standard errors were estimated using the Hessian as part of model-fitting. 
    Significance tests reported using the two-sided Wald test as implemented in the \texttt{ordinal} package \cite{OrdinalPackage} in \textbf{R} \cite{RSoftware}.
    *: $P < 0.05$, **: $P < 0.01$, ***: $P< 0.001$ \\
    \textdagger: The full measure description is ``Helpful for interacting with the simulated environment.'' 
\end{tablenotes}
\end{threeparttable}
\end{adjustbox}
\end{table}

\begin{table}[htbp]
\centering
\footnotesize
\begin{adjustbox}{center}
\begin{threeparttable}
\caption{\textbf{Mixed model without fitness marginal means comparison summary}  }
\label{tab:supplementary-no-fitness-marginal-means}
\begin{tabular}{|p{2.4cm}|rrl|rrl|rrl|}
\toprule
& \multicolumn{3}{c|}{\textbf{\texttt{Real} $-$ \texttt{Matched}}} & \multicolumn{3}{c|}{\textbf{\texttt{Real} $-$ \texttt{Unmatched}}} & \multicolumn{3}{c|}{\textbf{\texttt{Matched} $-$ \texttt{Unmatched}}} \\
\textbf{Attribute} & Diff $\pm SE$ & $Z$ & $P$-value & Diff $\pm SE$ & $Z$ & $P$-value & Diff $\pm SE$ & $Z$ & $P$-value \\ 
  \midrule
\textit{Understandable} $\uparrow$ & $-0.001 \pm 0.331$ & -0.003 & $P = 1.000$ & $1.042 \pm 0.332$ & 3.138 & $P = \num{4.837e-03}^{ ** }$ & $1.042 \pm 0.333$ & 3.133 & $P = \num{4.927e-03}^{ ** }$ \\ 
  \textit{Fun to play} $\uparrow$  & $0.143 \pm 0.266$ & 0.538 & $P = 0.853$ & $1.020 \pm 0.274$ & 3.722 & $P = \num{5.791e-04}^{ *** }$ & $0.877 \pm 0.273$ & 3.210 & $P = \num{3.791e-03}^{ ** }$ \\ 
  \textit{Fun to watch} $\uparrow$ & $0.135 \pm 0.250$ & 0.542 & $P = 0.850$ & $0.892 \pm 0.259$ & 3.446 & $P = \num{1.650e-03}^{ ** }$ & $0.757 \pm 0.257$ & 2.944 & $P = \num{9.076e-03}^{ ** }$ \\ 
  \textit{Helpful}\textsuperscript{\textdagger} & $0.016 \pm 0.159$ & 0.097 & $P = 0.995$ & $0.251 \pm 0.165$ & 1.521 & $P = 0.281$ & $0.236 \pm 0.165$ & 1.426 & $P = 0.328$ \\ 
  \textit{Difficult} ${\downarrow \atop \uparrow }$  & $-0.200 \pm 0.357$ & -0.559 & $P = 0.842$ & $-0.194 \pm 0.361$ & -0.538 & $P = 0.853$ & $0.006 \pm 0.361$ & 0.016 & $P = 1.000$ \\ 
   \textit{Creative} $\uparrow$   & $0.228 \pm 0.310$ & 0.736 & $P = 0.742$ & $0.489 \pm 0.316$ & 1.548 & $P = 0.269$ & $0.261 \pm 0.314$ & 0.832 & $P = 0.683$ \\ 
  \textit{Human-like} $\uparrow$  & $0.199 \pm 0.274$ & 0.727 & $P = 0.748$ & $1.396 \pm 0.283$ & 4.927 & $P = \num{2.495e-06}^{ *** }$ & $1.197 \pm 0.283$ & 4.225 & $P = \num{7.088e-05}^{ *** }$ \\ 
   \bottomrule
\end{tabular}
\begin{tablenotes}
    \item \textbf{We recover the lack of statistically significant differences between the \texttt{real} and \texttt{matched} groups, even excluding the fitness scores from the analysis.}
    We use the method of estimated marginal (least-squares) means as implemented in the \texttt{emmeans} package \cite{EmmeansPackage} to estimate the mean score for each attribute in each category, holding fitness constant. 
    None of the comparisons between the \texttt{real} and \texttt{matched} groups are significant, and several (though not all) of the previously significant comparisons remain significant. 
    Standard errors (of the differences in estimated means) were estimated using the \texttt{emmeans} package. 
    We report two-sided significance tests adjusted using the Tukey method to control for the multiple difference tests within each attribute, as implemented in the \texttt{emmeans} package.
    *: $P < 0.05$, **: $P < 0.01$, ***: $P< 0.001$ \\
    \textdagger: The full measure description is ``Helpful for interacting with the simulated environment.'' \\
    In most measures, higher scores are better, indicated by the $\uparrow$, other than \textit{Difficult}  ${\downarrow \atop \uparrow }$, in which 3 means ``appropriately difficult'', and scores below and above indicate too easy and too hard respectively.
\end{tablenotes}
\end{threeparttable}
\end{adjustbox}
\end{table}

\FloatBarrier
\subsection{Fitness-inclusive mixed-effect model analyses}
\label{sec:appendix-mixed-effect-analyses}

In the \nameref{sec:human-evals} section, we briefly describe the mixed effect model we fit to analyze our human evaluation results and analyze the learned regression weights for the fitness function. 
Here, we build on this analysis to examine whether accounting for the mediating effect of fitness scores changes our previous observations regarding the differences between groups. 
Using the \texttt{unmatched} group as the baseline, the regression coefficients $\beta_{\text{matched}}$ and $\beta_{\text{real}}$ quantify these differences for each measure. 
We find statistically significant differences for the \texttt{matched} group (i.e. $\beta_{\text{matched}} > 0$) for ratings of fun to play, fun to watch, helpfulness, and human likeness. 
Similarly, we observe statistically significant differences ($\beta_{\text{real}} > 0$) for ratings of understandability, fun to play and watch, and human likeness. 
Finally, using the marginal (least-squares) means method\cite{EmmeansPackage}, we directly compare the \texttt{matched} and \texttt{real} categories and again find no statistically significant differences (see \nameref{methods:human-evals} for additional details and Supplementary \autoref{tab:supplementary-marginal-means} below for the full results).

\begin{table}[h]
\centering
\footnotesize
\begin{adjustbox}{center}
\begin{threeparttable}
\caption{\textbf{Mixed model result summary}  }
\label{tab:supplementary-mixed-models}
\begin{tabular}{|p{2cm}|rrl|rrl|rrl|}
\toprule
& \multicolumn{3}{c|}{\textbf{Fitness}} & \multicolumn{3}{c|}{\texttt{Matched}} & \multicolumn{3}{c|}{\texttt{Real}} \\
\textbf{Attribute} & $\beta_{\text{fitness}} \pm SE$ & $Z$ & $P$-value & $\beta_{\text{matched}} \pm SE$ & $Z$ & $P$-value & $\beta_{\text{real}} \pm SE$ & $Z$ & $P$-value \\ 
  \midrule
\textit{Understandable} & $0.846 \pm 0.150$ & 5.625 & $P = \num{1.858e-08}^{ *** }$ & $0.525 \pm 0.297$ & 1.766 & $P = 0.077$ & $1.151 \pm 0.285$ & 4.036 & $P = \num{5.431e-05}^{ *** }$ \\ 
  \textit{Fun to play}    & $0.396 \pm 0.135$ & 2.936 & $P = \num{3.322e-03}^{ ** }$ & $0.629 \pm 0.274$ & 2.298 & $P = 0.022^{ * }$ & $1.059 \pm 0.263$ & 4.021 & $P = \num{5.797e-05}^{ *** }$ \\ 
  \textit{Fun to watch} & $0.191 \pm 0.130$ & 1.469 & $P = 0.142$ & $0.641 \pm 0.266$ & 2.414 & $P = 0.016^{ * }$ & $0.912 \pm 0.257$ & 3.547 & $P = \num{3.901e-04}^{ *** }$ \\ 
  \textit{Helpful}\textsuperscript{\textdagger} & $-0.189 \pm 0.087$ & -2.163 & $P = 0.031^{ * }$ & $0.349 \pm 0.170$ & 2.048 & $P = 0.041^{ * }$ & $0.232 \pm 0.161$ & 1.441 & $P = 0.150$ \\ 
  \textit{Difficult} & $-0.588 \pm 0.171$ & -3.443 & $P = \num{5.763e-04}^{ *** }$ & $0.363 \pm 0.353$ & 1.029 & $P = 0.304$ & $-0.250 \pm 0.338$ & -0.740 & $P = 0.460$ \\  
  \textit{Creative} & $-0.486 \pm 0.152$ & -3.191 & $P = \num{1.416e-03}^{ ** }$ & $0.551 \pm 0.310$ & 1.776 & $P = 0.076$ & $0.438 \pm 0.298$ & 1.467 & $P = 0.142$ \\ 
  \textit{Human-like} & $0.570 \pm 0.132$ & 4.316 & $P = \num{1.589e-05}^{ *** }$ & $0.837 \pm 0.268$ & 3.128 & $P = \num{1.762e-03}^{ ** }$ & $1.446 \pm 0.258$ & 5.597 & $P = \num{2.179e-08}^{ *** }$ \\ 
   \bottomrule
\end{tabular}
\begin{tablenotes}
    \item \textbf{Fitness scores significantly predict several attributes, including understandability and human-likeness.}
    Fitness scores show (statistically) significant positive effects on the understandability, fun to play, and human-likeness attributes, and significant negative effects on the difficulty and creativity questions. Accounting for the role of fitness, the \texttt{matched} group membership shows a significant effect only on human likeness. The \texttt{real} group shows significant effects on understandability, fun to play to watch, and human likeness. 
    Standard errors were estimated using the Hessian as part of model-fitting. 
    Significance tests reported using the two-sided Wald test as implemented in the \texttt{ordinal} package \cite{OrdinalPackage} in \textbf{R} \cite{RSoftware}.
    *: $P < 0.05$, **: $P < 0.01$, ***: $P< 0.001$ \\
    \textdagger: The full measure description is ``Helpful for interacting with the simulated environment.'' 
\end{tablenotes}
\end{threeparttable}
\end{adjustbox}
\end{table}

\label{sec:appendix-marginal-means-analysis}

\begin{table}[htbp]
\centering
\footnotesize
\begin{adjustbox}{center}
\begin{threeparttable}
\caption{\textbf{Mixed model marginal means comparison summary}  }
\label{tab:supplementary-marginal-means}
\begin{tabular}{|p{2.4cm}|rrl|rrl|rrl|}
\toprule
& \multicolumn{3}{c|}{\textbf{\texttt{Real} $-$ \texttt{Matched}}} & \multicolumn{3}{c|}{\textbf{\texttt{Real} $-$ \texttt{Unmatched}}} & \multicolumn{3}{c|}{\textbf{\texttt{Matched} $-$ \texttt{Unmatched}}} \\
\textbf{Attribute} & Diff $\pm SE$ & $Z$ & $P$-value & Diff $\pm SE$ & $Z$ & $P$-value & Diff $\pm SE$ & $Z$ & $P$-value \\ 
  \midrule
    \textit{Understandable} $\uparrow$ & $0.626 \pm 0.305$ & 2.055 & $P = 0.099$ & $1.151 \pm 0.285$ & 4.036 & $P = \num{1.606e-04}^{ *** }$ & $0.525 \pm 0.297$ & 1.766 & $P = 0.181$ \\ 
   \textit{Fun to play} $\uparrow$ & $0.430 \pm 0.273$ & 1.577 & $P = 0.256$ & $1.059 \pm 0.263$ & 4.021 & $P = \num{1.713e-04}^{ *** }$ & $0.629 \pm 0.274$ & 2.298 & $P = 0.056$ \\  
  \textit{Fun to watch} $\uparrow$ & $0.271 \pm 0.264$ & 1.025 & $P = 0.561$ & $0.912 \pm 0.257$ & 3.547 & $P = \num{1.135e-03}^{ ** }$ & $0.641 \pm 0.266$ & 2.414 & $P = 0.042^{ * }$ \\  
  \textit{Helpful}\textsuperscript{\textdagger} $\uparrow$  & $-0.117 \pm 0.167$ & -0.701 & $P = 0.763$ & $0.232 \pm 0.161$ & 1.441 & $P = 0.320$ & $0.349 \pm 0.170$ & 2.048 & $P = 0.101$ \\ 
  \textit{Difficult} ${\downarrow \atop \uparrow }$ & $-0.613 \pm 0.355$ & -1.725 & $P = 0.196$ & $-0.250 \pm 0.338$ & -0.740 & $P = 0.740$ & $0.363 \pm 0.353$ & 1.029 & $P = 0.559$ \\ 
  \textit{Creative} $\uparrow$  & $-0.113 \pm 0.310$ & -0.364 & $P = 0.930$ & $0.438 \pm 0.298$ & 1.467 & $P = 0.307$ & $0.551 \pm 0.310$ & 1.776 & $P = 0.178$ \\ 
  \textit{Human-like} $\uparrow$ & $0.609 \pm 0.265$ & 2.299 & $P = 0.056$ & $1.446 \pm 0.258$ & 5.597 & $P = \num{6.531e-08}^{ *** }$ & $0.837 \pm 0.268$ & 3.128 & $P = \num{5.013e-03}^{ ** }$ \\ 
   \bottomrule
\end{tabular}
\begin{tablenotes}
    \item We use the method of estimated marginal (least-squares) means as implemented in the \texttt{emmeans} package \cite{EmmeansPackage} to estimate the mean score for each attribute in each category, holding fitness constant. 
    None of the comparisons between the \texttt{real} and \texttt{matched} groups are significant, and several (though not all) of the previously significant comparisons remain significant. 
    Standard errors (of the differences in estimated means) were estimated using the \texttt{emmeans} package. 
    We report two-sided significance tests adjusted using the Tukey method to control for the multiple difference tests within each attribute, as implemented in the \texttt{emmeans} package.
    *: $P < 0.05$, **: $P < 0.01$, ***: $P< 0.001$ \\
    \textdagger: The full measure description is ``Helpful for interacting with the simulated environment.'' \\
    In most measures, higher scores are better, indicated by the $\uparrow$, other than \textit{Difficult}  ${\downarrow \atop \uparrow }$, in which 3 means ``appropriately difficult'', and scores below and above indicate too easy and too hard respectively.
\end{tablenotes}
\end{threeparttable}
\end{adjustbox}
\end{table}

\FloatBarrier
\subsection{Matched-real game similarity analysis}
\label{sec:appendix-matched-real-similarity}

To functionally measure similarity, we leverage the fact that our goal programs are interpretable and automatically evaluate them on all gameplay interactions generated by participants in our first experiment. 
For each \texttt{matched} game and its corresponding \texttt{real} counterpart, we measure the number of interactions that fulfill a gameplay element in only the \texttt{matched} game, only the \texttt{real} game, or both.
While some pairs of games have their elements fulfilled by the same interactions (suggesting functional similarity), most pairs are not --- under 25\% (7/30) share more than half of their relevant interactions. 
Furthermore, the average Jaccard similarity between the sets of relevant interactions for the \texttt{matched} and \texttt{real} game is only 0.347 and the median similarity is 0.180 (identical games would score 1.0, entirely dissimilar games 0; and see summary in \autoref{fig:real-matched-comparison} and methodological details in \nameref{methods:similarity}).

\subsection{Random effects analysis}
\label{sec:appendix-random-effects}
We are interested in trying to understand which games appear to `underperform' their fitness (and category membership). 
We analyze this question through the random effects assigned by our model to each evaluated game (and participant). 
We focus on the human likeness question, and reproduce below, for each of the \texttt{real}, \texttt{matched}, and \texttt{unmatched} categories, the four games with the largest negative random effect.
These are the games whose ratings were lowest compared to what would be predicted from the coefficients our model learned for their fitness and group membership, accounting for the random effects learned for the specific participants who rated them. 
We provide both the back-translated natural language descriptions as well as the programs.

\autoref{fig:appendix-human-like-raneff-real} showcases the four participant-created games, and \autoref{fig:appendix-human-like-raneff-matched-unmatched} the eight model-generated games, four \texttt{matched} and four \texttt{unmatched}.
We examine these to try to understand the limitations of our current model and evaluation process, though we are aware that it is impossible to make far-reaching conclusions on the basis of such anecdotal evidence.
Of the \texttt{real} games, two appear to be building games using the blocks in the room, and the other two appear to be particularly simple games with short programs and descriptions.
The building games appear to highlight the limitation of our back-translation process (see \cref{sec:appendix-backtranslation} for details) --- it is certainly plausible that there are more human-like ways to describe these programs. 
The simple games appear to draw tension between the type of games participants created playing around in our environment, which account for having to learn an unfamiliar interface, and the evaluation judgments, made directly from reading descriptions (though being provided a description and images of the playroom in our virtual environment (see \autoref{fig:methods-human-evals-interface} for an example). 

The model-generated games offer fewer discernible patterns. 
The first \texttt{matched} game includes a condition that appears quite difficult to satisfy, though not impossible: placing a chair on the desk and having it move without being held. 
The fourth \texttt{matched} model game highlights another failure mode: the program, as specified, rewards throwing balls but penalizes landing them in the bin, the opposite of how human participants tended to combine these conditions. 
Otherwise, the model-generated games offer a combination of simple (1-2 gameplay preferences) and complex (4 gameplay preferences) programs. 
Some of these games appear reasonably coherent (only including block-stacking or throwing elements), while other games mix and match different gameplay components in a way only a few human participants did. 
We draw no immediate conclusions from this analysis, beyond identifying that the space of programs that are realizable in our grammar and achievable in the environment contains examples that, while reasonably coherent, are still not particularly human-like. 

\rotatebox{90}{\begin{minipage}{0.9\textheight}
    \vspace{-1in}
    \includegraphics[width=\columnwidth]{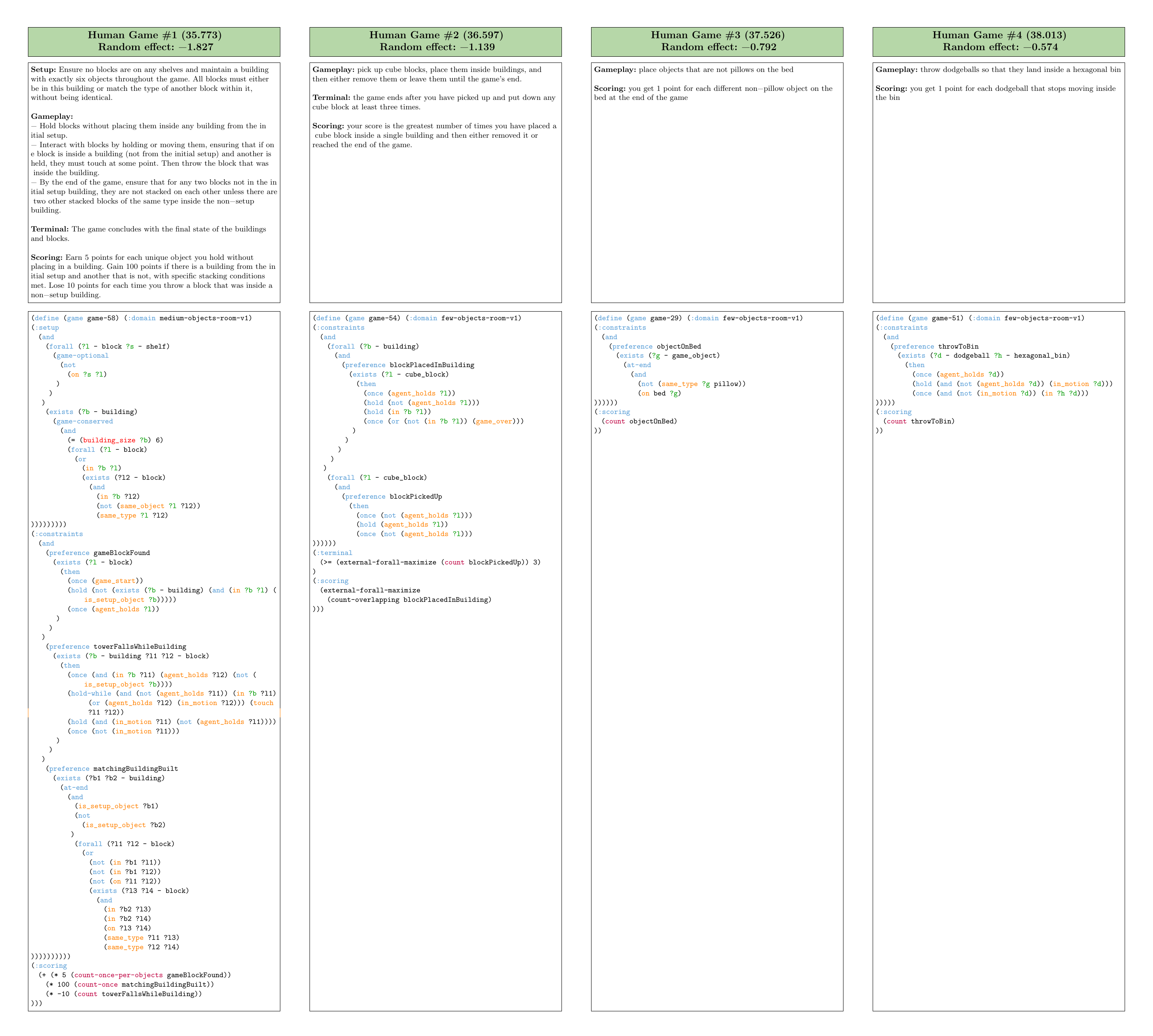}
    \captionof{figure}{\textbf{Games from the \texttt{real} category with the largest negative random effects on human likeness.}
    } 
    \label{fig:appendix-human-like-raneff-real}
\end{minipage}}

\rotatebox{90}{\begin{minipage}{0.75\textheight}
    \vspace{-1in}
    \includegraphics[width=\columnwidth]{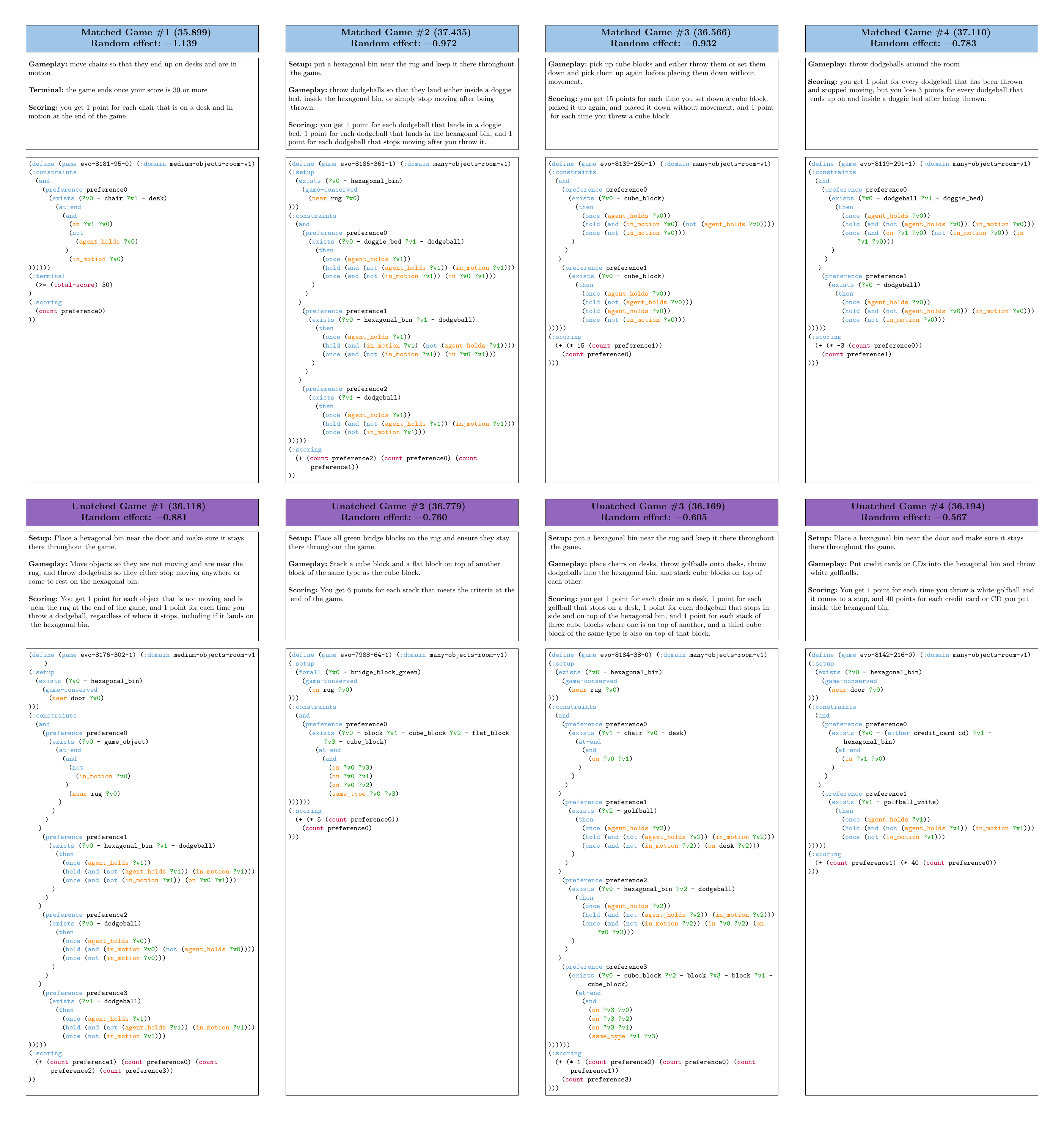}
    \captionof{figure}{\textbf{Games from the \texttt{matched}/\texttt{unmatched} categories with the largest negative random effects on human likeness.}
    } 
    \label{fig:appendix-human-like-raneff-matched-unmatched}
\end{minipage}}

\subsection{Unmatched game only mixed-model analysis}
\label{sec:appendix-evals-unamtched-only}
To evaluate how well our fitness function captures desirable properties out of distribution, we repeat the mixed model analyses of our human evaluation results, restricting only to the games in the \texttt{unmatched} category. 
We fit mixed effects models predicting each human evaluation attribute from the fitness score for only \texttt{unmatched} games, with random effects for the participant and rated game. 
If the fitness model fails entirely away from the support of the participant-created games, then for these samples, there should be no association between the fitness scores and the properties the participants rated. 

\begin{table}[!htbp]
\centering
\footnotesize
\begin{threeparttable}
\caption{\textbf{\texttt{Unmatched} games only mixed model result summary}}
\label{tab:supplementary-unmatched-only-mixed-models}
\noindent\begin{tabularx}{\linewidth}{XcX}
& 
\begin{tabular}{|p{2cm}|rrl|}
\toprule
& \multicolumn{3}{c|}{\textbf{Fitness}}  \\
\textbf{Attribute} & $\beta_{\text{fitness}} \pm SE$ & $Z$ & $P$-value  \\ 
  \midrule
    \textit{Understandable}  & $0.724 \pm 0.204$ & 3.556 & $P = \num{3.771e-04}^{ *** }$ \\ 
  \textit{Fun to play} & $0.616 \pm 0.261$ & 2.362 & $P = 0.018^{ * }$ \\ 
  \textit{Fun to watch} & $0.382 \pm 0.321$ & 1.189 & $P = 0.235$ \\ 
  \textit{Helpful}\textsuperscript{\textdagger} & $-0.233 \pm 0.179$ & -1.301 & $P = 0.193$ \\ 
  \textit{Difficult} & $-0.674 \pm 0.332$ & -2.027 & $P = 0.043^{ * }$ \\ 
  \textit{Creative} & $-0.214 \pm 0.320$ & -0.670 & $P = 0.503$ \\ 
  \textit{Human-like} & $0.798 \pm 0.239$ & 3.335 & $P = \num{8.518e-04}^{ *** }$ \\

   \bottomrule
\end{tabular} 
&
\end{tabularx}
\begin{tablenotes}
    \item \textbf{On \texttt{unmatched} games only, fitness scores still significantly predict several attributes, including understandability and human-likeness.}
    Fitness scores show (statistically) significant positive effects on the understandability, fun to play, and human-likeness attributes and a significant negative effect on the difficulty questions. 
    Compared to the full mixed model results reported in \autoref{tab:supplementary-mixed-models}, we observe a similar set of significant effects, though mostly at lower significance levels (which we ascribe in part to the smaller sample size).
    The only effect that doesn't reproduce its significance is the one on the \textit{creative} question.
    Standard errors were estimated using the Hessian as part of model-fitting. 
    Significance tests reported using the two-sided Wald test as implemented in the \texttt{ordinal} package \cite{OrdinalPackage} in \textbf{R} \cite{RSoftware}.
    *: $P < 0.05$, **: $P < 0.01$, ***: $P< 0.001$ \\
    \textdagger: The full measure description is ``Helpful for interacting with the simulated environment.'' 
\end{tablenotes}
\end{threeparttable}
\end{table}

\autoref{tab:supplementary-unmatched-only-mixed-models} summarizes our findings.
Compared to the effects of fitness in \autoref{tab:supplementary-mixed-models}, the fitness function appears to play a similar result in explaining human judgments.
The positive statistically significant effects on the understandability question, the fun to play, and human likeness all reproduce, albeit with lower significance levels (which we ascribe in part to the smaller sample size). 
Simultaneously, the statistically significant negative effect on difficulty is maintained.
We take this evidence to suggest that even among \texttt{unmatched} games only, higher fitness scores are associated with higher understandability, fun, and human-likeness ratings and, conversely, with a lower difficulty rating.
Finally, one effect from the full model does not replicate---the negative effect of fitness on creativity.
We take this to mean that within unmatched games by themselves, higher fitness is not associated with lower creativity judgments.

\FloatBarrier
\section{Model ablations}
\label{sec:appendix-ablations}

\subsection{Common sense ablation} 
\label{sec:appendix-ablations-common-sense}
The domain-specific language we use is underconstrained---many expressions that are grammatical either make no sense at all (e.g., checking a bin is in a ball, rather than a ball in a bin) or violate intuitive physical common sense (e.g., creating a game stacking balls, as opposed to stacking blocks).
We primarily operationalize the concept of physical common sense using two of our fitness features, discussed in \nameref{methods:fitness-function} and \cref{sec:appendix-features}. 
Both use a dataset of interaction traces (see \nameref{methods:dataset}) to estimate the feasibility of predicate role-filler expressions, by computing the proportion of predicate expressions (and the object types they operate over) that have appeared at least once over the set of interactions of users with the environment. 
While this condition is not necessary (as it is unlikely experiment participants explored every feasible configuration of objects in the environment), it is sufficient to determine feasibility and, therefore, serves as a good proxy for intuitive common sense.
The first feature operates over individual predicates, e.g. estimating that \lstinline{(on desk ball)} is more likely than \lstinline{(on desk bed)}.
The second feature operates on logical expressions over predicates, and might help catch contradictory predicates that are independently feasible, such as \lstinline{(and (on desk ball) (on bed ball)) }, that is, the ball might feasibly be on the desk or on the bed, but not on both. 

We know that these features are helpful for our model, as the individual predicate version of these features has the third highest weight of all features that predict real human-generated games (see \cref{sec:appendix-predictive-features} for details). 
To further evaluate the importance of these features, we fit a version of our fitness model that has no access to them, and use it as the objective for our model. 
Unsurprisingly, when we evaluate samples from this ablated model on the full fitness function (with the interaction trace features), they have statistically significantly lower fitness scores than the samples from the full model (matched-pairs $t$-test matching by archive cells, $t = -32.66, P < \num{1e-10}$). 
To offer a more fair comparison, we use the full ``reward machine'' and dataset of play traces. 
We assign a binary score to each game from the full and ablated models, 1 if each game component (gameplay preferences and the setup section (if one exists)) is satisfied at least once over the dataset, either in the same trace or in different traces.
If at least one game component is never satisfied, we assign a score of 0.
We find that 1515 (75.75\%) of the games in the full model score 1, while only 584 (29.20\%) in the ablated model do.
This difference is, as expected, also statistically significant (matched-pairs $t$-test, $t = -33.29, P < \num{1e-10}$).
We conclude that intuitive physical common sense is helpful to our model, as allowing our model to approximate the physical sensibility of predicates helps the model generate games with components that have been satisfied by our participants. 


\subsection{Compositionality ablation}
\label{sec:appendix-ablations-compositionality}

Evaluating the role of compositionality in our model is challenging as the model operates on a domain-specific language that is inherently highly compositional. 
Given the nature of program representations, it's difficult to imagine a non-compositional counterfactual DSL to compare to --- so we cannot compare to an entirely non-compositional model.
Instead, we ablate by varying how compositional we allow our MAP-Elites mutation operators to be. 
The primary operator embodying compositionality is the crossover operator, which samples two programs from the MAP-Elites archive, randomly selects exchangeable sub-trees from both programs, and creates new candidates with these trees swapped between the programs. 
We also implemented several custom operators (beyond the evolutionary programming staples of mutation, insertion, deletion, and crossover). 
Many of these implement targeted variations of crossover that we considered to be plausible higher-level changes a person might make to a game they are creating, such as sampling a preference from another game and then changing the preference's initial or terminal conditions. 
We report two ablations, one (``No Custom Ops'') where we omit the custom operators we implemented (keeping only regrowth, insertion, deletion, and crossover), and a second (``No Custom Ops, No Crossover'') where we also remove the crossover operation. 
We keep all other model details identical, crucially both the set of behavioral characteristics and fitness function, allowing us to directly compare the fitness values of games in the archives in the ablated models. 

\begin{figure}[!tbph]
    \begin{subfigure}[T]{0.48 \textwidth}
        \centering
        \includegraphics[width=\textwidth]{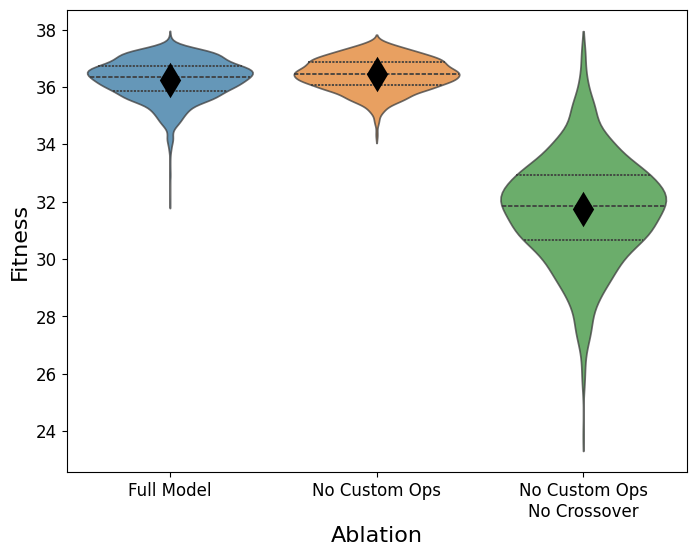}
        
        \caption{\textbf{Removing crossover drastically lowers fitness values.}
        We plot, for each game generated by a model, its fitness score under the full fitness function. Lines represent the quartiles of the data. 
        \textbf{Left:} The distribution of fitness scores in our full model.
        \textbf{Middle:} Removing the custom operators has little effect on the distribution of fitness scores.
        \textbf{Right:} Removing crossover drastically lowers the fitness scores of model samples.
        }
        \label{fig:appendix-compositionality-ablations-fitness}
    \end{subfigure}
    \hfill
    \begin{subfigure}[T]{0.48 \textwidth}
        \centering
        \includegraphics[width=\textwidth]{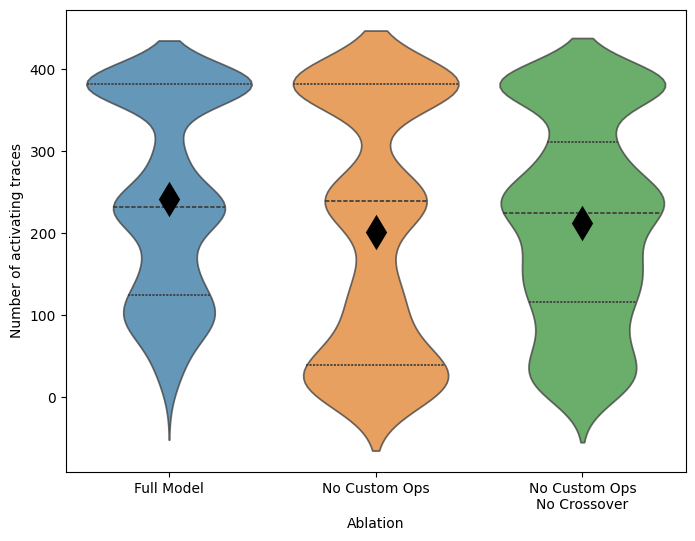}
        \caption{\textbf{Removing custom operators lowers mean trace coverage; removing crossover undoes some of the effect.}
        We measure, for each game generated by a model, how many participant interaction traces fulfill at least one gameplay element. 
        Lines represent the quartiles of the data. 
        \textbf{Left:} Of the ablations reported, our full model shows the highest number of active traces.
        \textbf{Middle:} removing our custom mutation operators lowers the mean number of active traces.
        \textbf{Right:} Removing crossover as well undoes some of the effect of removing the custom operators.
        }
        \label{fig:appendix-compositionality-ablations-diversity}
    \end{subfigure}
\caption{\textbf{The crossover operator helps generate fit goals, while the custom operators help generate solutions with higher trace coverage.}
\textbf{Left:} removing the custom operators does hurt mean fitness scores; removing the crossover operator does.
\textbf{Right:} removing the custom operators leads the model to generate samples covering fewer participant interaction traces on average. 
This could be evidence of lower feasibility (more samples in the ``no custom ops'' model are active in barely a few traces) or of mode seeking (more samples in the full model are active in a very high number of traces). 
}
\label{fig:appendix-compositionality-ablations}
\end{figure}

We visualize the results of these in \autoref{fig:appendix-compositionality-ablations}.
While removing our custom operators appears to slightly \textit{increase} the mean fitness of exemplars (\autoref{fig:appendix-compositionality-ablations-fitness}, orange), removing the crossover operation drastically decreases the fitness of games in that model's archive (\autoref{fig:appendix-compositionality-ablations-fitness}, green). 
This provides evidence that allowing our search procedure to take advantage of the compositionality in our domain is greatly beneficial in generating high-quality samples across our archive.
If our custom operators do not increase mean fitness, what impact do they have?
To quantify this question, we evaluated samples from the ablated models through the full ``reward machine.''
For each sample, we counted how many of the participant interaction traces saw the participant fulfilling one or more gameplay elements from the sample.
In other words, how many participants (unwittingly or otherwise) fulfilled at least part of the model-generated goal program?
We find that the custom operators help increase this number -- samples generated from our full model show the highest number of relevant traces (\autoref{fig:appendix-compositionality-ablations-diversity}, blue). 
This could have two interpretations: one is that the custom operators push more goals toward higher feasibility. 
Another is that this behavior is a form of mode-seeking that helps the model generate goal programs that capture more common behaviors, as opposed to more meaningful variability. 
In all, we take this as an effect that crossover is crucial to generating fit samples across our MAP-Elites archive, with some cost to diversity which our custom operators help reduce.

\subsection{Coherence ablation}
\label{sec:appendix-ablations-coherence}

As we iterated on earlier versions of our model, we discovered some `softer', higher-level issues repeatedly surfacing in model-generated goal programs. 
Even after implementing features that helped the model avoid some types of low-level mistakes (such as instantiating variables or preferences and never referencing them), and introducing approximations to intuitive physical common sense (discussed above), some aspects of the generated games remained incoherent. 
A lower-level example might be disjointedness in the arguments of temporal modals.
Consider, for instance, a preference whose modals translate to natural language as ``start with a state where the agent holds a ball, then find a collection of states where a block is on the bed, and finish with a state where the bin is upside down''. 
The awkwardness in explaining this perfectly grammatical preference (program below) is that each modal (\lstinline{(once ..), (hold ...)}) refers to a distinct set of objects, and so it feels unnatural to specify a sequential temporal preference over them. 

\begin{lstlisting}[basicstyle=\normalsize]
(preference preference0
  (exists (?v0 - hexagonal_bin ?v1 - ball ?v2 - block)
    (then
      (once (agent_holds ?v1))
      (hold (on desk ?v2))
      (once (object_orientation ?v0 upside_down))
)))
\end{lstlisting}

We observed similar, higher-level issues regarding coherence between different gameplay preferences (do they use the same objects and predicates, or distinct sets?). 
Specifically, we observed cases where game scoring conditions and ending conditions have nothing to do with each other.
A game might specify that it ends after a ball has been thrown five times, with points scored for every block placed on the desk. 
There is nothing wrong per se with this specification, but it feels unnatural---we would expect either the ball-throwing to contribute to scoring, or the block-stacking to allow the game to end, or both. 
We wrote a collection of fitness features to try to capture occurrences of such incoherence (see \texttt{game\_element\_disjointness} in \cref{sec:appendix-features}).
We have some indication that these features are important from observing that our fitness model assigns one of them the third-largest negative weight (predictive of corrupted, negative games). 
To ablate the effect of this feature group, we perform an ablation similar to the common sense ablation reported above---we fit a fitness model without these features and use it to guide our MAP-Elites search. 
As a first sanity check, we compute fitness scores under the full fitness model for games generated by the ablated model.
We find that scores in the ablated model are consistently lower (matched-pairs $t$-test, $t = -26.99, P < \num{1e-10}$), indicating that without access to these features, our model would generate programs that violate these coherence considerations. 
We also evaluate games from this ablated model using the ``reward machine'' and play traces dataset, as we did above.
As before, 1515 (75.75\%) of games in the full model have every component satisfied, while only 1224 (61.2\%) in the ablated model do. 
This difference is also statistically significant (matched-pairs $t$-test, $t = -9.73, P < \num{1e-10}$).

\FloatBarrier
\subsection{PCFG-sampling only baseline ablation}
\label{sec:appendix-prior-only-baseline}

We wish to demonstrate the quality of the samples our model begins from and illustrate the necessity of the complex mechanisms of our model.
To that effect, we provide results from repeatedly sampling from the PCFG representing our grammar, fitted to the rule and terminal counts in our small human dataset.
We generate over 6 million samples from the grammar (6144000, matching the total number of candidates our full MAP-Elites model considers over its full search process). 
We insert these into the MAP-Elites archive following an identical procedure to the full model: a new sample is inserted into the archive if it either fills an empty cell or receives a higher fitness score than the current example in its cell. 

\begin{figure}[!hbtp]
    \centering
    \includegraphics[width= \textwidth]{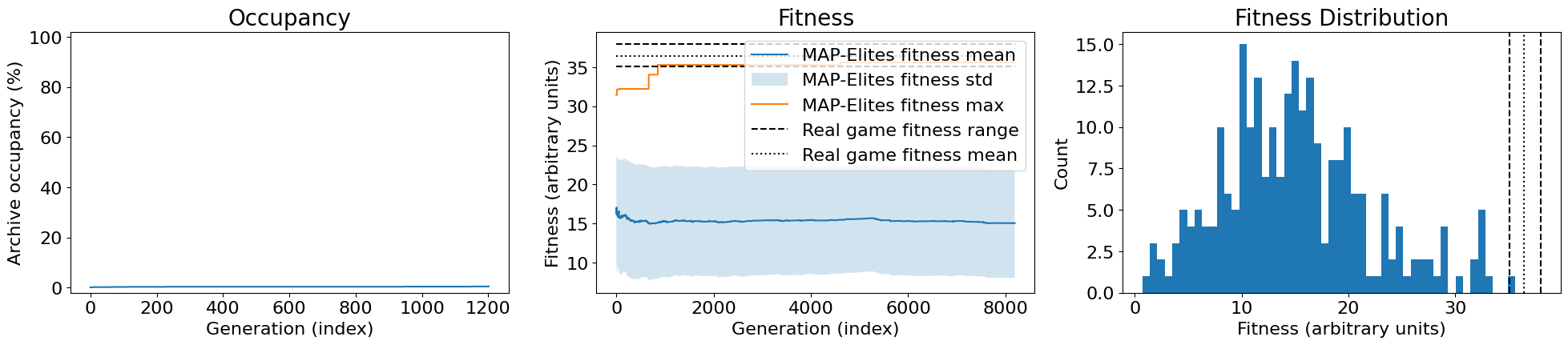}
    \caption{
        \textbf{Sampling only from the PCFG fails to match our full model .}
        Contrast this figure with \autoref{fig:results-quantitative}. 
        \textbf{Left:} PCFG samples occupy very little of the MAP-Elites archive: after generating over 6M samples, only 251/2000 (12.55\%) of the archive cells are occupied.
        \textbf{Middle:} PCFG samples are mostly drastically below the range of fitness scores assigned to human-created programs.
        \textbf{Right:} Another visualization of the distribution of fitness scores of samples directly from the PCFG. 
    }
    \label{fig:prior-only-results-quantitative}
\end{figure}

\autoref{fig:prior-only-results-quantitative} summarizes our findings. 
We find that sampling from the PCFG, by itself, neither explores the space of programs well nor generates high-quality programs. 
The 6M samples generated occupy only 251/2000 (12.55\%) of archive cells, and the generated samples vary wildly in quality, mostly falling far below the fitness scores of real games, with a mean fitness of 15.05 and a standard deviation of 6.98. 
Finally, with respect to our auxiliary check (see Auxiliary coherence check under \nameref{methods:map-elites}), only 11 of the 6144000 samples pass the heuristic criteria.
In contrast, our full model's archive is seeded from 1024 examples generated from this PCFG, and manages to fill the archive entirely with substantially fitter samples. 

\FloatBarrier
\subsection{Held-out data model ablation}
\label{sec:appendix-held-out}

To examine the robustness of our procedure to held-out data, we evaluate a version of our model using a fitness function trained on a subset of the human-created games.
We randomly held out 20 games and trained the fitness function on the remaining 78 participant-created samples. 
Otherwise, our model remains identical to the full version of our model reported in this work---we follow the same feature extraction process, using the same feature set, and search using MAP-Elites with the same behavioral features (none of the exemplar preferences used to structure our search, reported in  \autoref{tab:exemplar-preferences}, were from a held-out game). 

\begin{figure}[!htbp]
    \begin{subfigure}[T]{0.48 \textwidth}
        \centering
        \includegraphics[width=\textwidth]{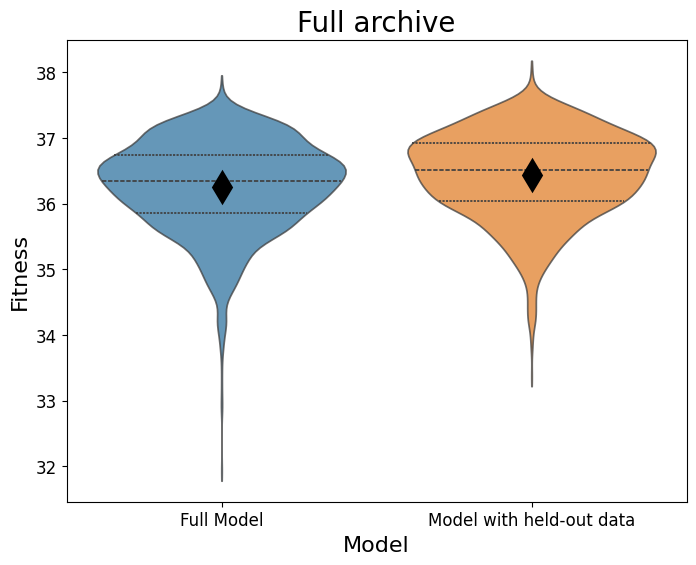}
        \caption{\textbf{Samples from the held-out model score slightly higher than samples from the full model.}
        We score samples from both models under the fitness function used for the full model.
        Lines represent the quartiles of the data. 
        \textbf{Left:} Full model.
        \textbf{Right:} Model with held-out data. 
        }
        \label{fig:appendix-held-out-full-archive}
    \end{subfigure}
    \hfill
    \begin{subfigure}[T]{0.48 \textwidth}
        \centering
        \includegraphics[width=\textwidth]{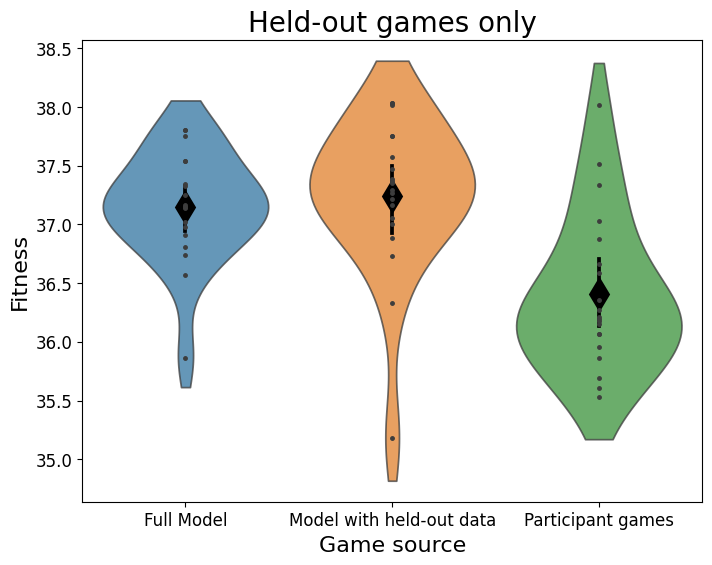}
        \caption{\textbf{In the held-out game cells, both models score higher than the participant games.}
        We score samples from both models under the fitness function used for the full model. 
        We plot the scores of the 20 games included in the held-out set and their matching games from each model.
        \textbf{Left:} Full model.
        \textbf{Middle:} Model with held-out data.
        \textbf{Right:} Participant games.
        }
        \label{fig:appendix-ablations-held-out-games-only}
    \end{subfigure}
\caption{\textbf{A version of our model optimizing a fitness function with held-out data performs \textit{slightly better} than our full model.}
Both panels report the distribution of fitness scores under the fitness function used in our full model (trained on our entire 98 participant-created game dataset). 
\textbf{Left:} Full distribution of scores in the archive.
\textbf{Right:} Scores on the games from archive cells matching the held-out games. 
}
\label{fig:appendix-held-out-ablations}
\end{figure}

We find that the fitness scores of games used to train the function (mean 36.58, standard deviation 0.71) are very similar to those of the held-out games (mean 36.55, standard deviation 0.79), and are statistically indistinguishable (two-sample $t$-test,$t = 0.13, P = 0.89, df = 26.94$). 
To compare this model to our full model, we evaluate the fitness scores of samples from the model with held-out data under the fitness function from the full model (that is, we generate samples to optimize an objective trained on partial data, and evaluate on an objective trained on the full dataset).
We summarize our findings in \autoref{fig:appendix-held-out-ablations}. 
We were surprised to find out that not only is the model optimizing the held-out data fitness function not worse than our full model, it appears to be incrementally better on average.
The difference is small (full model mean 36.26 and standard deviation 0.69; held-out model mean 36.44, standard deviation 0.66), and due to the large ($n = 2000$) sample size, statistically significant (matched-pairs $t$-test, $t = -18.50, P = \num{1.10e-70}, df = 1999$). 
We hypothesize this result may be driven by some degree of chance:
small differences in the proposals accepted during the search can cascade into substantial variations over the full procedure.
Alternatively, the smaller sample size might have allowed the fitness function to concentrate even further toward the mode of the distribution. 
We previously hypothesized that our fitness function learns to place the highest fitness scores on games that are over-represented in our dataset, primarily simple throwing games. 
By sub-sampling the data, we might have allowed the optimization process to learn weights that perform this mode seeking even more strongly, allowing higher scores to be generated at the risk of capturing less diversity.

\FloatBarrier

\section{Full domain-specific language description}
\label{sec:appendix-dsl-ref}

A game is defined by a name, and is expected to be valid in a particular domain, also referenced by a name.
A game is defined by four elements, two of them mandatory, and two optional.
The mandatory ones are the \dsl{constraints} section, which defines gameplay preferences, and the \dsl{scoring} section, which defines how gameplay preferences are counted to arrive at a score for the player in the game.
The optional ones are the \dsl{setup} section, which defines how the environment must be prepared before gameplay can begin, and the \dsl{terminal} conditions, which specify when and how the game ends.

\begin{grammar}
<game> ::= (define (game <ID>) \\
  (:domain <ID>) \\
  (:setup <setup>) \\
  (:constraints <constraints>) \\
  (:terminal <terminal>) \\
  (:scoring <scoring>) \\)

<id> ::= /[a-z0-9][a-z0-9\-]+/  "#" a letter or digit, followed by one or more letters, digits, or dashes
\end{grammar}

We will now proceed to introduce and define the syntax for each of these sections, followed by the non-grammar elements of our domain: predicates, functions, and types.
Finally, we provide a mapping between some aspects of our gameplay preference specification and linear temporal logic (LTL) operators.

\subsection{Setup} \label{sec:setup}
The setup section specifies how the environment must be transformed from its deterministic initial conditions to a state gameplay can begin at.
Currently, a particular environment room always appears in the same initial conditions, in terms of which objects exist and where they are placed.
Participants in our experiment could, but did not have to, specify how the room must be setup so that their game could be played.

The initial \dsl{setup} element can expand to conjunctions, disjunctions, negations, or quantifications of itself, and then to the \dsl{setup-statement} rule.
\dsl{setup-statement} elements specify two different types of setup conditions: either those that must be conserved through gameplay (`game-conserved'), or those that are optional through gameplay (`game-optional').
These different conditions arise as some setup elements must be maintain through gameplay (for example, a participant specified to place a bin on the bed to throw balls into, it shouldn't move unless specified otherwise), while other setup elements can or must change (if a participant specified to set the balls on the desk to throw them, an agent will have to pick them up (and off the desk) in order to throw them).

Inside the \dsl{setup-statement} tags we find \dsl{super-predicate} elements, which are logical operations and quantifications over other \dsl{super-predicate} elements,  function comparisons (\dsl{function-comparison}, which like predicates also resolve to a truth value), and predicates (\dsl{predicate}).
Function comparisons usually consist of a comparison operator and two arguments, which can either be the evaluation of a function or a number.
The one exception is the case where the comparison operator is the equality operator (=), in which case any number of arguments can be provided.
Finally, the \dsl{predicate} element expands to a predicate acting on one or more objects or variables.
For a full list of the predicates we found ourselves using so far, see \autoref{sec:predicates}.
        
\begin{grammar}
<setup> ::= (and <setup> <setup>$^+$) "#" A setup can be expanded to a conjunction, a disjunction, a quantification, or a setup statement (see below).
    \alt (or <setup> <setup>$^+$)
    \alt (not <setup>)
    \alt (exists (<variable-list>) <setup>)
    \alt (forall (<variable-list>) <setup>)
    \alt <setup-statement>

<setup-statement> ::= "#" A setup statement specifies that a predicate is either optional during gameplay or must be preserved during gameplay.
    \alt (game-conserved <super-predicate>)
    \alt (game-optional <super-predicate>)

<super-predicate> ::= "#" A super-predicate is a conjunction, disjunction, negation, or quantification over another super-predicate. It can also be directly a function comparison or a predicate.
    \alt (and <super-predicate>$^+$)
    \alt (or <super-predicate>$^+$)
    \alt (not <super-predicate>
    \alt (<variable-list>) <super-predicate>)
    \alt (<variable-list>) <super-predicate>)
    \alt <f-comp>
    \alt <predicate>

<function-comparison> ::= "#" A function comparison: either comparing two function evaluations, or checking that two ore more functions evaluate to the same result.
    \alt (<comp-op> <function-eval-or-number> <function-eval-or-number>)
    \alt (= <function-eval-or-number>$^+$)

<comp-op> ::=  \textlangle \ | \textlangle = \ | = \ | \textrangle \ | \textrangle = "#" Any of the comparison operators.

<function-eval-or-number> ::= <function-eval> | <comparison-arg-number>

<comparison-arg-number> ::= <number>

<number> ::=  /-?\textbackslash d*\textbackslash .?\textbackslash d+/  "#" A number, either an integer or a float.

<function-eval> ::= "#" See valid expansions in a separate section below

<variable-list> ::= (<variable-def>$^+$) "#" One or more variables definitions, enclosed by parentheses.

<variable-def> ::= <variable-type-def> 
    \alt <color-variable-type-def> |
    \alt <orientation-variable-type-def> 
    \alt <side-variable-type-def> "#" Colors, sides, and orientations are special types as they are not interchangable with objects.

<variable-type-def> ::= <variable>$^+$ - <type-def> "#" Each variable is defined by a variable (see next) and a type (see after).

<color-variable-type-def> ::= <color-variable>$^+$ - <color-type-def> "#" A color variable is defined by a variable (see below) and a color type.

<orientation-variable-type-def> ::= <orientation-variable>$^+$ - <orientation-type-def> "#" An orientation variable is defined by a variable (see below) and an orientation type.

<side-variable-type-def> ::= <side-variable>$^+$ - <side-type-def> "#" A side variable is defined by a variable (see below) and a side type.

<variable> ::= /\textbackslash?[a-w][a-z0-9]*/  "#" a question mark followed by a lowercase a-w, optionally followed by additional letters or numbers.

<color-variable> ::= /\textbackslash?x[0-9]*/  "#" a question mark followed by an x and an optional number.

<orientation-variable> ::= /\textbackslash?y[0-9]*/  "#" a question mark followed by an y and an optional number.

<side-variable> ::= /\textbackslash?z[0-9]*/  "#" a question mark followed by an z and an optional number.

<type-def> ::= <object-type> | <either-types> "#" A veriable type can either be a single name, or a list of type names, as specified below

<color-type-def> ::= <color-type> | <either-color-types> "#" A color variable type can either be a single color name, or a list of color names, as specified below

<orientation-type-def> ::= <orientation-type> | <either-orientation-types> "#" An orientation variable type can either be a single orientation name, or a list of orientation names, as specified below

<side-type-def> ::= <side-type> | <either-side-types> "#" A side variable type can either be a single side name, or a list of side names, as specified below

<either-types> ::= (either <object-type>$^+$)

<either-color-types> ::= (either <color>$^+$)

<either-orientation-types> ::= (either <orientation>$^+$)

<either-side-types> ::= (either <side>$^+$)

<object-type> ::= <name>

<name> ::= /[A-Za-z][A-za-z0-9\_]+/  "#" a letter, followed by one or more letters, digits, or underscores

<color-type> ::= 'color'

<color> ::= 'blue' | 'brown' | 'gray' | 'green' | 'orange' | 'pink' | 'purple' | 'red' | 'tan' | 'white' | 'yellow'

<orientation-type> ::= 'orientation'

<orientation> ::= 'diagonal' | 'sideways' | 'upright' | 'upside\_down'

<side-type> ::= 'side'

<side> ::= 'back' | 'front' | 'left' | 'right'

<predicate> ::= "#" See valid expansions in a separate section below

<predicate-or-function-term> ::= <object-name> | <variable> "#" A predicate or function term can either be an object name (from a small list allowed to be directly referred to) or a variable.

<predicate-or-function-color-term> ::= <color> | <color-variable>

<predicate-or-function-orientation-term> ::= <orientation> | <orientation-variable>

<predicate-or-function-side-term> ::= <side> | <side-variable>

<predicate-or-function-type-term> ::= <object-type> | <variable>

<object-name> ::= 'agent' | 'bed' | 'desk' | 'door' | 'floor' | 'main\_light\_switch' | 'mirror' | 'room\_center' | 'rug' | 'side\_table' | 'bottom\_drawer' | 'bottom\_shelf' | 'east\_sliding\_door' | 'east\_wall' | 'north\_wall' | 'south\_wall' | 'top\_drawer' | 'top\_shelf' | 'west\_sliding\_door' | 'west\_wall'

\end{grammar}

\subsection{Gameplay Preferences} \label{sec:constraints}
The gameplay preferences specify the core of a game's semantics, capturing how a game should be played by specifying temporal constraints over predicates.
The name for the overall element, \dsl{constraints}, is inherited from the PDDL element with the same name.

The \dsl{constraints} elements expands into one or more preference definitions, which are defined using the \dsl{pref-def} element.
A \dsl{pref-def} either expands to a single preference (\dsl{preference}), or to a \dsl{pref-forall} element, which specifies variants of the same preference for different objects, which can be treated differently in the scoring section.
A \dsl{preference} is defined by a name and a \dsl{preference-quantifier}, which expands to an optional quantification (exists, forall, or neither), inside of which we find the \dsl{preference-body}.

A \dsl{preference-body} expands into one of two options:
The first is a set of conditions that should be true at the end of gameplay, using the \dsl{at-end} operator.
Inside an \dsl{at-end} we find a \dsl{super-predicate}, which like in the setup section, expands to logical operations or quantifications over other \dsl{super-predicate} elements, function comparisons, or predicates.

The second option is specified using the \dsl{then} syntax, which defines a series of temporal conditions that should hold over a sequence of states.
Under a \dsl{then} operator, we find two or more sequence functions (\dsl{seq-func}), which define the specific conditions that must hold and how many states we expect them to hold for.
We assume that there are no unaccounted states between the states accounted for by the different operators -- in other words, the \dsl{then} operators expects to find a sequence of contiguous states that satisfy the different sequence functions.
The operators under a \dsl{then} operator map onto linear temporal logic (LTL) operators, see \autoref{sec:LTL} for the mapping and examples.

The \dsl{once} operator specifies a predicate that must hold for a single world state.
If a \dsl{once} operators appears as the first operator of a \dsl{then} definition, and a sequence of states $S_a, S_{a+1}, \cdots, S_b$ satisfy the \dsl{then} operator, it could be the case that the predicate is satisfied before this sequence of states (e.g. by $S_{a-1}, S_{a-2}$, and so forth).
However, only the final such state, $S_a$, is required for the preference to be satisfied.
The same could be true at the end of the sequence: if a \dsl{then} operator ends with a \dsl{once} term, there could be other states after the final state ($S_{b+1}, S_{b+2}$, etc.) that satisfy the predicate in the \dsl{once} operator, but only one is required.
The \dsl{once-measure} operator is a slight variation of the \dsl{once} operator, which in addition to a predicate, takes in a function evaluation, and measures the value of the function evaluated at the state that satisfies the preference.
This function value can then be used in the scoring definition, see \autoref{sec:scoring}.

A second type of operator that exists is the \dsl{hold} operator.
It specifies that a predicate must hold true in every state between the one in which the previous operator is satisfied, and until one in which the next operator is satisfied.
If a \dsl{hold} operator appears at the beginning or an end of a \dsl{then} sequence, it can be satisfied by a single state,
Otherwise, it must be satisfied until the next operator is satisfied.
For example, in the minimal definition below:
\begin{lstlisting}
(then
    (once (pred_a))
    (hold (pred_b))
    (once (pred_c))
)
\end{lstlisting}
To find a sequence of states $S_a, S_{a+1}, \cdots, S_b$ that satisfy this \dsl{then} operator, the following conditions must hold true: (1) pred\_a is true at state $S_a$, (2) pred\_b is true in all states $S_{a+1}, S_{a+2}, \cdots, S_{b-2}, S_{b-1}$, and (3) pred\_c is true in state $S_b$.
There is no minimal number of states that the hold predicate must hold for.

The last operator is \dsl{hold-while}, which offers a variation of the \dsl{hold} operator.
A \dsl{hold-while} receives at least two predicates.
The first acts the same as predicate in a \dsl{hold} operator.
The second (and third, and any subsequent ones), must hold true for at least state while the first predicate holds, and must occur in the order specified.
In the example above, if we substitute \lstinline{(hold (pred_b))} for \lstinline{(hold-while (pred_b) (pred_d) (pred_e))}, we now expect that in addition to pred\_b being true in all states $S_{a+1}, S_{a+2}, \cdots, S_{b-2}, S_{b-1}$, that there is some state $S_d, d \in [a+1, b-1]$ where pred\_d holds, and another state, $S_e, e \in [d+1, b-1]$ where pred\_e holds.
        
\begin{grammar}
<constraints> ::= <pref-def> | (and <pref-def>$^+$)  "#" One or more preferences.

<pref-def> ::= <pref-forall> | <preference> "#" A preference definitions expands to either a forall quantification (see below) or to a preference.

<pref-forall> ::= (forall <variable-list> <preference>) "#" this syntax is used to specify variants of the same preference for different objects, which differ in their scoring. These are specified using the <pref-name-and-types> syntax element's optional types, see scoring below.

<preference> ::= (preference <name> <preference-quantifier>) "#" A preference is defined by a name and a quantifer that includes the preference body.

<preference-quantifier> ::= "#" A preference can quantify exsistentially or universally over one or more variables, or none.
\alt (exists (<variable-list>) <preference-body>
\alt  (forall (<variable-list>) <preference-body>)
\alt <preference-body>)

<preference-body> ::=  <then> | <at-end>

<at-end> ::= (at-end <super-predicate>) "#" Specifies a prediicate that should hold in the terminal state.

<then> ::= (then <seq-func> <seq-func>$^+$) "#" Specifies a series of conditions that should hold over a sequence of states -- see below for the specific operators (<seq-func>s), and Section 2 for translation of these definitions to linear temporal logicl (LTL).

<seq-func> ::= <once> | <once-measure> | <hold> | <hold-while> "#" Four of thse temporal sequence functions currently exist: 

<once> ::= (once <super-predicate>) "#" The predicate specified must hold for a single world state.

<once-measure> ::= (once <super-predicate> <function-eval>) "#" The predicate specified must hold for a single world state, and record the value of the function evaluation, to be used in scoring.

<hold> ::= (hold <super-predicate>) "#" The predicate specified must hold for every state between the previous temporal operator and the next one.

<hold-while> ::= (hold-while <super-predicate> <super-predicate>$^+$) "#" The first predicate specified must hold for every state between the previous temporal operator and the next one. While it does, at least one state must satisfy each of the predicates specified in the second argument onward

\end{grammar}
For the full specification of the \dsl{super-predicate} element, see \autoref{sec:setup} above.

\subsection{Terminal Conditions} \label{sec:terminal}
Specifying explicit terminal conditions is optional, and while some of our participants chose to do so, many did not.
Conditions explicitly specified in this section terminate the game.
If none are specified, a game is assumed to terminate whenever the player chooses to end the game.

The terminal conditions expand from the \dsl{terminal} element, which can expand to logical conditions on nested \dsl{terminal} elements, or to a terminal comparison.
The terminal comparison (\dsl{terminal-comp}) expands to one of three different types of copmarisons: \dsl{terminal-time-comp}, a comparison between the total time spent in the game (\lstinline{(total-time)}) and a time number token, \dsl{terminal-score-comp}, a comparison between the total score (\lstinline{(total-score)}) and a score number token, or \dsl{terminal-pref-count-comp}, a comparison between a scoring expression (\dsl{scoring-expr}, see below) and a preference count number token.
In most cases, the scoring expression is a preference counting operation.
        
\begin{grammar}
<terminal> ::= "#" The terminal condition is specified by a conjunction, disjunction, negation, or comparson (see below).
        \alt (and <terminal>$^+$)
        \alt (or <terminal>$+$)
        \alt (not <terminal>)
        \alt <terminal-comp>

<terminal-comp> ::= "#" We support three ttypes of terminal comparisons:
        \alt <terminal-time-comp>
        \alt <terminal-score-comp>
        \alt <terminal-pref-count-comp>

    <terminal-time-comp> ::= (<comp-op> (total-time) <time-number>) "#" The total time of the game must satisfy the comparison.

    <terminal-score-comp> ::= (<comp-op> (total-score) <score-number>) "#" The total score of the game must satisfy the comparison.

    <terminal-pref-count-comp> ::= (<comp-op> <scoring-expr> <preference-count-number>) "#" The number of times the preference specified by the name and types must satisfy the comparison.

    <time-number> ::= <number>  "#" Separate type so the we can learn a separate distribution over times than, say, scores.

    <score-number> ::= <number>

    <preference-count-number> ::= <number>

    <comp-op> ::=  \textlangle \ | \textlangle = \ | = \ | \textrangle \ | \textrangle =

\end{grammar}
For the full specification of the \dsl{scoring-expr} element, see \autoref{sec:scoring} below.

\subsection{Scoring} \label{sec:scoring}
Scoring rules specify how to count preferences (count once, once for each unique objects that fulfill the preference, each time a preference is satisfied, etc.), and the arithmetic to combine preference counts to a final score in the game.

A \dsl{scoring-expr} can be defined by arithmetic operations on other scoring expressions, references to the total time or total score (for instance, to provide a bonus if a certain score is reached), comparisons between scoring expressions (\dsl{scoring-comp}), or by preference evaluation rules.
Various preference evaluation modes can expand the \dsl{preference-eval} rule, see the full list and descriptions below.
        
\begin{grammar}
<scoring> ::= <scoring-expr> "#" The scoring conditions maximize a scoring expression. 

<scoring-expr> ::= "#" A scoring expression can be an arithmetic operation over other scoring expressions, a reference to the total time or score, a comparison, or a preference scoring evaluation.
        \alt <scoring-external-maximize>
        \alt <scoring-external-minimize>
        \alt (<multi-op> <scoring-expr>$^+$) "#" Either addition or multiplication.
        \alt (<binary-op> <scoring-expr> <scoring-expr>) "#" Either division or subtraction.
        \alt (- <scoring-expr>)
        \alt (total-time)
        \alt (total-score)
        \alt <scoring-comp>
        \alt <preference-eval>
        \alt <scoring-number-value>

<scoring-external-maximize> ::= (external-forall-maximize <scoring-expr>) "#" For any preferences under this expression inside a (forall ...), score only for the single externally-quantified object that maximizes this scoring expression.

<scoring-external-minimize> ::= (external-forall-minimize <scoring-expr>) "#" For any preferences under this expression inside a (forall ...), score only for the single externally-quantified object that minimizes this scoring expression.

<scoring-comp> ::=  "#" A scoring comparison: either comparing two expressions, or checking that two ore more expressions are equal.
        \alt (<comp-op> <scoring-expr> <scoring-expr>)
        \alt (= <scoring-expr>$^+$)

<preference-eval> ::= "#" A preference evaluation applies one of the scoring operators (see below) to a particular preference referenced by name (with optional types).
        \alt <count>
        \alt <count-overlapping>
        \alt <count-once>
        \alt <count-once-per-objects>
        \alt <count-measure>
        \alt <count-unique-positions>
        \alt <count-same-positions>
        \alt <count-once-per-external-objects>

<count> ::= (count <pref-name-and-types>) "#" Count how many times the preference is satisfied by non-overlapping sequences of states.

<count-overlapping> ::= (count-overlapping <pref-name-and-types>) "#" Count how many times the preference is satisfied by overlapping sequences of states.

<count-once> ::= (count-once <pref-name-and-types>) "#" Count whether or not this preference was satisfied at all.

<count-once-per-objects> ::= (count-once-per-objects <pref-name-and-types>) "#" Count once for each unique combination of objects quantified in the preference that satisfy it.

<count-measure> ::= (count-measure <pref-name-and-types>) "#" Can only be used in preferences including a <once-measure> modal, maps each preference satistifaction to the value of the function evaluation in the <once-measure>.

<count-unique-positions> ::= (count-unique-positions <pref-name-and-types>) "#" Count how many times the preference was satisfied with quantified objects that remain stationary within each preference satisfcation, and have different positions between different satisfactions.

<count-same-positions> ::= (count-same-positions <pref-name-and-types>) "#" Count how many times the preference was satisfied with quantified objects that remain stationary within each preference satisfcation, and have (approximately) the same position between different satisfactions.

<count-once-per-external-objects> ::=  (count-once-per-external-objects <pref-name-and-types>) "#" Similarly to count-once-per-objects, but counting only for each unique object or combination of objects quantified in the (forall ...) block including this preference.

<pref-name-and-types> ::= <name> <pref-object-type>$^*$ "#" The optional <pref-object-type>s are used to specify a particular instance of the preference for a given object, see the <pref-forall> syntax above.

    <pref-object-type> ::= : <type-name>  "#" The optional type name specification for the above syntax. For example, pref-name:dodgeball would refer to the preference where the first quantified object is a dodgeball.

    <scoring-number-value> ::= <number>

\end{grammar}

\section{Non-Grammar Definitions}

\subsection{Predicates} \label{sec:predicates}
The following section describes the predicates we define.
        Predicates operate over a specified number of arguments, which can be variables or object names, and return a boolean value (true/false).

\begin{lstlisting}
(above <arg1> <arg2>) [5 references] ; Is the first object above the second object?
(adjacent <arg1> <arg2>) [84 references] ; Are the two objects adjacent? [will probably be implemented as distance below some threshold]
(adjacent_side <3 or 4 arguments>) [15 references] ; Are the two objects adjacent on the sides specified? Specifying a side for the second object is optional, allowing to specify <obj1> <side1> <obj2> or <obj1> <side1> <obj2> <side2>
(agent_crouches ) [2 references] ; Is the agent crouching?
(agent_holds <arg1>) [327 references] ; Is the agent holding the object?
(between <arg1> <arg2> <arg3>) [7 references] ; Is the second object between the first object and the third object?
(broken <arg1>) [2 references] ; Is the object broken?
(equal_x_position <arg1> <arg2>) [2 references] ; Are these two objects (approximately) in the same x position? (in our environment, x, z are spatial coordinates, y is the height)
(equal_z_position <arg1> <arg2>) [5 references] ; Are these two objects (approximately) in the same z position? (in our environment, x, z are spatial coordinates, y is the height)
(faces <arg1> <arg2>) [6 references] ; Is the front of the first object facing the front of the second object?
(game_over ) [4 references] ; Is this the last state of gameplay?
(game_start ) [3 references] ; Is this the first state of gameplay?
(in <arg1> <arg2>) [121 references] ; Is the second argument inside the first argument? [a containment check of some sort, for balls in bins, for example]
(in_motion <arg1>) [312 references] ; Is the object in motion?
(is_setup_object <arg1>) [13 references] ; Is this the object of the same type referenced in the setup?
(near <arg1> <arg2>) [63 references] ; Is the second object near the first object? [implemented as distance below some threshold]
(object_orientation <arg1> <arg2>) [14 references] ; Is the first argument, an object, in the orientation specified by the second argument? Used to check if an object is upright or upside down
(on <arg1> <arg2>) [165 references] ; Is the second object on the first one?
(open <arg1>) [3 references] ; Is the object open? Only valid for objects that can be opened, such as drawers.
(opposite <arg1> <arg2>) [4 references] ; So far used only with walls, or sides of the room, to specify two walls opposite each other in conjunction with other predicates involving these walls
(rug_color_under <arg1> <arg2>) [11 references] ; Is the color of the rug under the object (first argument) the color specified by the second argument?
(same_color <arg1> <arg2>) [23 references] ; If two objects, do they have the same color? If one is a color, does the object have that color?
(same_object <arg1> <arg2>) [7 references] ; Are these two variables bound to the same object?
(same_type <arg1> <arg2>) [14 references] ; Are these two objects of the same type? Or if one is a direct reference to a type, is this object of that type?
(toggled_on <arg1>) [4 references] ; Is this object toggled on?
(touch <arg1> <arg2>) [48 references] ; Are these two objects touching?
\end{lstlisting}

\subsection{Functions} \label{sec:functions}
he following section describes the functions we define.
        Functions operate over a specified number of arguments, which can be variables or object names, and return a number.
\begin{lstlisting}
(building_size <arg1>) [2 references] ; Takes in an argument of type building, and returns how many objects comprise the building (as an integer).
(distance <arg1> <arg2>) [50 references] ; Takes in two arguments of type object, and returns the distance between the two objects (as a floating point number).
(distance_side <arg1> <arg2> <arg3>) [6 references] ; Similarly to the adjacent_side predicate, but applied to distance. Takes in three or four arguments, either <obj1> <side1> <obj2> or <obj1> <side1> <obj2> <side2>, and returns the distance between the first object on the side specified to the second object (optionally to its specified side).
(x_position <arg1>) [4 references] ; Takes in an argument of type object, and returns the x position of the object (as a floating point number).
\end{lstlisting}

\subsection{Types} \label{sec:types}
The types are currently not defined as part of the grammar, other than the small list of \dsl{object-name} tokens that can be directly referred to, and are marked with an asterisk below, and the sides, colors, and orientations, which are separated from object types. 
        The following enumerates all expansions of the various \dsl{type} rules:
        
\begin{lstlisting}
game_object [33 references] ; Parent type of all objects
agent* [100 references] ; The agent
building [20 references] ; Not a real game object, but rather, a way to refer to structures the agent builds
---------- (* \textbf{Blocks} *) ----------
block [28 references] ; Parent type of all block types:
bridge_block [11 references] 
bridge_block_green [0 references] 
bridge_block_pink [0 references] 
bridge_block_tan [0 references] 
cube_block [38 references] 
cube_block_blue [8 references] 
cube_block_tan [1 reference] 
cube_block_yellow [8 references] 
cylindrical_block [11 references] 
cylindrical_block_blue [0 references] 
cylindrical_block_green [0 references] 
cylindrical_block_tan [0 references] 
flat_block [5 references] 
flat_block_gray [0 references] 
flat_block_tan [0 references] 
flat_block_yellow [0 references] 
pyramid_block [13 references] 
pyramid_block_blue [3 references] 
pyramid_block_red [2 references] 
pyramid_block_yellow [2 references] 
tall_cylindrical_block [7 references] 
tall_cylindrical_block_green [0 references] 
tall_cylindrical_block_tan [0 references] 
tall_cylindrical_block_yellow [0 references] 
tall_rectangular_block [0 references] 
tall_rectangular_block_blue [0 references] 
tall_rectangular_block_green [0 references] 
tall_rectangular_block_tan [0 references] 
triangle_block [3 references] 
triangle_block_blue [0 references] 
triangle_block_green [0 references] 
triangle_block_tan [0 references] 
---------- (* \textbf{Balls} *) ----------
ball [40 references] ; Parent type of all ball types:
beachball [23 references] 
basketball [18 references] 
dodgeball [108 references] 
dodgeball_blue [6 references] 
dodgeball_red [4 references] 
dodgeball_pink [8 references] 
golfball [25 references] 
golfball_green [3 references] 
golfball_white [0 references] 
---------- (* \textbf{Colors} *) ----------
color [6 references] ; Likewise, not a real game object, mostly used to refer to the color of the rug under an object
blue [6 references] 
brown [5 references] 
gray [0 references] 
green [8 references] 
orange [3 references] 
pink [19 references] 
purple [4 references] 
red [8 references] 
tan [2 references] 
white [1 reference] 
yellow [14 references] 
---------- (* \textbf{Furniture} *) ----------
bed* [51 references] 
blinds [2 references] ; The blinds on the windows
desk* [45 references] 
desktop [6 references] 
main_light_switch* [3 references] ; The main light switch on the wall
side_table* [6 references] ; The side table/nightstand next to the bed
shelf_desk [2 references] ; The shelves under the desk
---------- (* \textbf{Large moveable/interactable objects} *) ----------
book [11 references] 
chair [18 references] 
laptop [7 references] 
pillow [14 references] 
teddy_bear [14 references] 
---------- (* \textbf{Orientations} *) ----------
diagonal [1 reference] 
sideways [2 references] 
upright [10 references] 
upside_down [1 reference] 
---------- (* \textbf{Ramps} *) ----------
ramp [0 references] ; Parent type of all ramp types:
curved_wooden_ramp [17 references] 
triangular_ramp [11 references] 
triangular_ramp_green [1 reference] 
triangular_ramp_tan [0 references] 
---------- (* \textbf{Receptacles} *) ----------
doggie_bed [27 references] 
hexagonal_bin [122 references] 
drawer [5 references] ; Either drawer in the side table
bottom_drawer* [0 references] ; The bottom of the two drawers in the nightstand near the bed.
top_drawer* [6 references] ; The top of the two drawers in the nightstand near the bed.
---------- (* \textbf{Room features} *) ----------
door* [15 references] ; The door out of the room
floor* [26 references] 
mirror* [0 references] 
poster* [0 references] 
room_center* [33 references] 
rug* [37 references] 
shelf [10 references] 
bottom_shelf* [1 reference] 
top_shelf* [5 references] 
sliding_door [2 references] ; The sliding doors on the south wall (big windows)
east_sliding_door* [1 reference] ; The eastern of the two sliding doors (the one closer to the desk)
west_sliding_door* [0 references] ; The western of the two sliding doors (the one closer to the bed)
wall [17 references] ; Any of the walls in the room
east_wall* [0 references] ; The wall behind the desk
north_wall* [1 reference] ; The wall with the door to the room
south_wall* [2 references] ; The wall with the sliding doors
west_wall* [3 references] ; The wall the bed is aligned to
---------- (* \textbf{Small objects} *) ----------
alarm_clock [8 references] 
cellphone [6 references] 
cd [6 references] 
credit_card [1 reference] 
key_chain [5 references] 
lamp [2 references] 
mug [3 references] 
pen [2 references] 
pencil [2 references] 
watch [2 references] 
---------- (* \textbf{Sides} *) ----------
back [3 references] 
front [9 references] 
left [3 references] 
right [2 references] 
\end{lstlisting}

\section{Modal Definitions in Linear Temporal Logic}
\label{sec:LTL}
\subsection{Linear Temporal Logic definitions}
We offer a mapping between the temporal sequence functions defined in \autoref{sec:constraints} and linear temporal logic (LTL) operators.
As we were creating this DSL, we found that the syntax of the \dsl{then} operator felt more convenient than directly writing down LTL, but we hope the mapping helps reason about how we see our temporal operators functioning.
LTL offers the following operators, using $\varphi$ and $\psi$ as the symbols (in our case, predicates).
Assume the following formulas operate sequence of states $S_0, S_1, \cdots, S_n$:
\begin{itemize}
    \item \textbf{Next}, $X \psi$: at the next timestep, $\psi$ will be true. If we are at timestep $i$, then $S_{i+1} \vdash \psi$

    \item \textbf{Finally}, $F \psi$: at some future timestep, $\psi$ will be true. If we are at timestep $i$, then $\exists j > i:  S_{j} \vdash \psi$

    \item \textbf{Globally}, $G \psi$: from this timestep on, $\psi$ will be true. If we are at timestep $i$, then $\forall j: j \geq i: S_{j} \vdash \psi$

    \item \textbf{Until}, $\psi U \varphi$: $\psi$ will be true from the current timestep until a timestep at which $\varphi$ is true. If we are at timestep $i$, then $\exists j > i: \forall k: i \leq k < j: S_k \vdash \psi$, and $S_j \vdash \varphi$.
    \item \textbf{Strong release}, $\psi M \varphi$: the same as until, but demanding that both $\psi$ and $\varphi$ are true simultaneously: If we are at timestep $i$, then $\exists j > i: \forall k: i \leq k \leq j: S_k \vdash \psi$, and $S_j \vdash \varphi$.

    \textit{Aside:} there's also a \textbf{weak until}, $\psi W \varphi$, which allows for the case where the second is never true, in which case the first must hold for the rest of the sequence. Formally, if we are at timestep $i$, \textit{if} $\exists j > i: \forall k: i \leq k < j: S_k \vdash \psi$, and $S_j \vdash \varphi$, and otherwise, $\forall k \geq i: S_k \vdash \psi$. Similarly there's \textbf{release}, which is the similar variant of strong release. We're leaving those two as an aside since we don't know we'll need them.

\end{itemize}

\subsection{Satisfying a \dsl{then} operator}
Formally, to satisfy a preference using a \dsl{then} operator, we're looking to find a sub-sequence of $S_0, S_1, \cdots, S_n$ that satisfies the formula we translate to.
We translate a \dsl{then} operator by translating the constituent sequence-functions (\dsl{once}, \dsl{hold}, \dsl{while-hold})\footnote{These are the ones we've used so far in the interactive experiment dataset, even if we previously defined other ones, too.} to LTL.
Since the translation of each individual sequence function leaves the last operand empty, we append a `true' ($\top$) as the final operand, since we don't care what happens in the state after the sequence is complete.

(once $\psi$) := $\psi X \cdots$

(hold $\psi$) := $\psi U \cdots$

(hold-while $\psi$ $\alpha$ $\beta$ $\cdots \nu$) := ($\psi M \alpha) X (\psi M \beta) X \cdots X (\psi M \nu) X \psi U \cdots$ where the last $\psi U \cdots$ allows for additional states satisfying $\psi$ until the next modal is satisfied.

For example, a sequence such as the following, which signifies a throw attempt:
\begin{lstlisting}
(then
    (once (agent_holds ?b))
    (hold (and (not (agent_holds ?b)) (in_motion ?b)))
    (once (not (in_motion ?b)))
)
\end{lstlisting}
Can be translated to LTL using $\psi:=$ (agent\_holds ?b), $\varphi:=$ (in\_motion ?b) as:

$\psi X (\neg \psi \wedge \varphi) U (\neg \varphi) X \top $

Here's another example:
\begin{lstlisting}
(then
    (once (agent_holds ?b))  (* \color{blue} $\alpha$*)
    (hold-while
        (and (not (agent_holds ?b)) (in_motion ?b)) (* \color{blue} $\beta$ *)
        (touch ?b ?r) (* \color{blue} $\gamma$*)
    )
    (once  (and (in ?h ?b) (not (in_motion ?b)))) (* \color{blue} $\delta$*)
)
\end{lstlisting}
If we translate each predicate to the letter appearing in blue at the end of the line, this translates to:

$\alpha X (\beta M \gamma) X \beta U \delta X \top$

\end{document}